\documentclass[9pt,a4paper,twocolumn,twoside,lineno]{article}

\usepackage[pages=all, color=black, position={current page.south}, placement=bottom, scale=1, opacity=1, vshift=5mm]{background}

\usepackage[margin=2.0cm]{geometry} % full-width

\usepackage{amsthm}
\usepackage{amsmath,amsfonts,amssymb}

\usepackage[utf8]{inputenc}
\usepackage{graphicx}
\usepackage{color}
\usepackage{multirow}
\usepackage{multicol}
\usepackage{rotating}
\usepackage{bm}
\usepackage{url}
\usepackage{caption}
\usepackage{booktabs}

\usepackage{array}
\usepackage{afterpage}
\usepackage{subfigure}
\usepackage[biblabel]{cite}
\usepackage{float}
\usepackage{placeins}
\usepackage{mathrsfs}

\usepackage{tikz}
\usetikzlibrary{positioning,fit}
\usetikzlibrary{matrix}
\usetikzlibrary{calc}
\usepackage{ifthen}
\usepackage{pgf}
\usepackage{pgffor}
\usepgfmodule{shapes}
\usepgfmodule{plot}
\usetikzlibrary{decorations}
\usetikzlibrary{arrows}
\usetikzlibrary{snakes}
\usepackage{pgfplots}

\usepackage{hyperref}
\usepackage{orcidlink}
\usepackage{etoolbox} % Patch the standard article title spacing

\usepackage{algorithm}
\usepackage{algorithmicx}
\usepackage{algpseudocode}
\algnewcommand{\LineComment}[1]{\State \(\triangleright\) #1}

\definecolor{darkgreen}{rgb}{0.0, 0.2, 0.13}
\definecolor{darkolivegreen}{rgb}{0.33, 0.42, 0.18}

\newcommand{\eg}{\textit{e.g.,}\ }
\newcommand{\ie}{\textit{i.e.,}\ }

\newcommand{\firstfigure}{%
  \par\vspace{1em}
  \begin{minipage}{\textwidth}
    \centering
    \includegraphics[width=\linewidth]{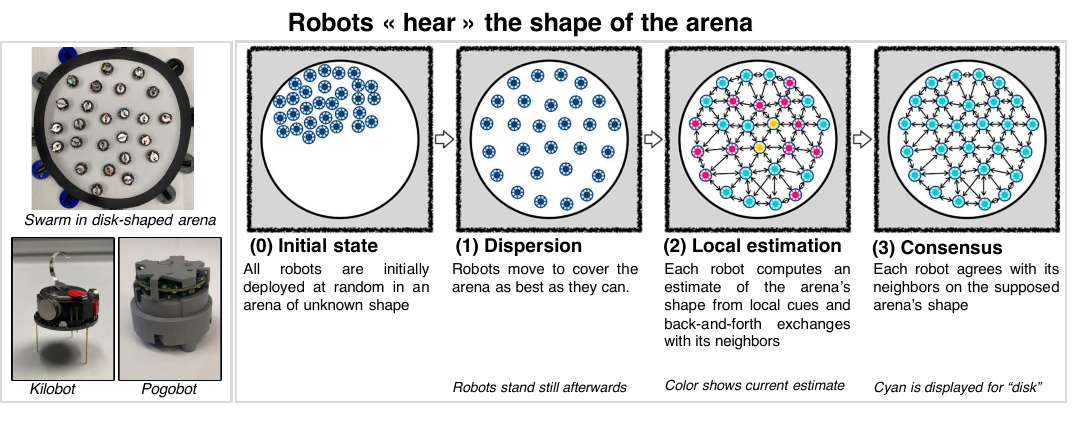}

    \captionof{figure}{\textbf{Framework of spectral swarm robotics.}
    {\small
    Description of our contribution: a swarm of robots computes the spectral
    fingerprint of its arena in a distributed way, so that each individual
    robot has its own fingerprint estimate; the robots reach a consensus on
    this value across the entire swarm; finally, this value is used to classify
    the shape in which the swarm is located. Our approach was validated using
    swarms of Kilobots~\cite{rubenstein2012kilobot,rubenstein2014programmable}
    and Pogobots~\cite{loi2025pogobot} robots.
    }}

    \label{fig:overview}
  \end{minipage}
  \par\vspace{1em}
}

\newcommand{\firstpageabstract}{%
  \par\vspace{0.5em}

  {\noindent\Large\bfseries Abstract\par}
  \vspace{0.4em}

  \begin{multicols}{2}
  \noindent
  Swarm robotics relies on robots with minimal sensing and computation that
  achieve self-organization at the level of the collective to address a
  variety of tasks. Yet swarm robotics still struggles with global tasks that
  require understanding the broader spatial context in which robots operate,
  such as identifying the shape of the arena in which they are embedded.

  Here, we introduce spectral swarm robotics, a framework in which
  robots recover global spectral descriptors of the swarm’s geometric
  structure through strictly local interactions. To do so, robots emulate
  diffusion to locally estimate Laplacian spectral invariants, quantities
  normally obtained only through centralized computation.

  These spectral values serve as compact descriptors of the environment. In
  the present proof of concept, robots estimated the second Laplacian
  eigenvalue, $\lambda_2$, and used it to classify arenas from a predefined set
  of shapes with distinct spectral signatures.

  We show that estimating a single eigenvalue is sufficient for one-shot
  classification among selected confined environments with distinct spectral
  signatures. We uncovered a scaling law linking swarm size (a global
  parameter) and interaction radius (a local parameter) that ensures robust
  estimation. We validated the approach experimentally on two distinct
  hardware platforms, Kilobots and Pogobots, achieving reliable classification
  of arena shapes under realistic noise and sparse-connectivity conditions.
  Last, we discuss how locally recovering Laplacian spectral invariants opens
  avenues for swarm robotics in settings where producing efficient collective
  behaviors strongly depends on the global structure of the swarm.
  \end{multicols}

  \par\vspace{0.5em}
}

\makeatletter
\def\ps@myheadings{%
    \let\@oddfoot\@empty\let\@evenfoot\@empty
    \def\@evenhead{\thepage\hfil\slshape\leftmark}%
    \def\@oddhead{{\slshape\rightmark}\hfil\thepage}%
    \let\@mkboth\@gobbletwo
    \let\sectionmark\@gobble
    \let\subsectionmark\@gobble
    }
  \if@titlepage
  \renewcommand\maketitle{\begin{titlepage}%
  \let\footnotesize\small
  \let\footnoterule\relax
  \let \footnote \thanks
  \null\vfil
  \vskip 60\p@
  \begin{center}%
    {\LARGE \@title \par}%
    \vskip 3em%
    {\large
     \lineskip .75em%
      \begin{tabular}[t]{c}%
        \@author
      \end{tabular}\par}%
      \vskip 1.5em%
    {\large \@date \par}%       % Set date in \large size.
  \end{center}\par
  \@thanks
  \vfil\null
  \end{titlepage}%
  \setcounter{footnote}{0}%
}
\else
\renewcommand\maketitle{\par
  \begingroup
    \renewcommand\thefootnote{\@fnsymbol\c@footnote}%
    \def\@makefnmark{%
      \rlap{\@textsuperscript{\normalfont\@thefnmark}}%
    }%
    \long\def\@makefntext##1{%
      \parindent 1em\noindent
      \hb@xt@1.8em{%
        \hss\@textsuperscript{\normalfont\@thefnmark}%
      }##1%
    }%
    \if@twocolumn
      \ifnum \col@number=\@ne
        \@maketitle
        \firstpageabstract
        \firstfigure
      \else
        \twocolumn[{%
          \@maketitle
          \firstpageabstract
          \firstfigure
        }]%
      \fi
    \thispagestyle{plain}\@thanks
  \endgroup
  \setcounter{footnote}{0}%
}
\makeatother

\makeatletter
\patchcmd{\@maketitle}{\null}{\null\vspace*{-1.5cm}}{}{%
  \PackageWarning{ssr-arxiv}{Could not adjust the title top spacing}%
}
\makeatother

\makeatletter

\newcommand{\makesupplementarytitle}{%
  \begingroup
    \centering

    {\LARGE
      \@title\par
    }

    \vspace{0.6em}

    {\Large\bfseries
      Supplementary Materials\par
    }

    \vspace{2em}

    {\large
      \lineskip .75em
      \begin{tabular}[t]{c}
        \@author
      \end{tabular}
      \par
    }

    \vspace{1.5em}

    {\large \@date\par}

  \endgroup

  \vspace{2em}
}

\makeatother

\title{Hearing the shape of an arena with spectral swarm robotics}

\author
{Leo Cazenille,$^{1,4\ast}$
Nicolas Lobato-Dauzier,$^{2,3\ast}$
Alessia Loi,$^4$\\
Loona Macabre,$^{4,5}$
Mika Ito,$^{5}$
Olivier Marchal,$^6$
Soo-Hyeon Kim,$^2$\\
Nathanael Aubert-Kato,$^5$
Nicolas Bredeche,$^{4,3\ast}$
Anthony J. Genot$^{2\dagger}$%\\
}

\date{
\footnotesize
$^{1}$Universit\'{e} Paris Cit\'{e}, CNRS, LIED UMR 8236, F-75006 Paris, France\\%
$^{2}$LIMMS (IRL2820)/CNRS-IIS, University of Tokyo, Tokyo, Japan\\%
$^{3}$Sorbonne Universit\'{e}, CNRS, Institut de Biologie Paris-Seine, Laboratoire Jean Perrin, F-75005, Paris, France\\%
$^{4}$Sorbonne Universit\'{e}, CNRS, Institut des Syst\`{e}mes Intelligents et de Robotique, ISIR, F-75005 Paris, France\\%
$^{5}$Ochanomizu University, Department of Information Sciences, Tokyo, Japan\\%
$^{6}$Universit\'{e} Jean Monnet Saint-\'{E}tienne, CNRS UMR 5208, Institut Camille Jordan, Institut Universitaire de France, F-42000 Saint-Etienne, France\\%
$^\ast$To whom correspondence should be addressed; E-mails:  leo.cazenille@gmail.com, nicolas.bredeche@sorbonne-universite.fr, nicolas.lobato-dauzier@sorbonne-universite.fr\\%
$^\dagger$In memory of Anthony J. Genot, who suddenly passed away on April 8th 2025. His guidance and enthusiasm for science remain with us, and we will miss him deeply.%
\\%
\small
This is the author's version of the work. It is posted here by permission of the AAAS for personal use, not for redistribution. The definitive version was published in \textit{Science Robotics} \textbf{11}(119), eadr9578 on 21 October 2026,\\
DOI: \url{https://doi.org/10.1126/scirobotics.adr9578}.
}

\begin{document} 

\maketitle

\section*{Introduction}

In swarm robotics~\cite{kube1993collective,beni1993swarm,dudek1993taxonomy,Brambilla2013,hamann2018swarm,dorigo2020reflections  ,floreano2021individual}, robots collaborate to solve a given problem. Each robot has a limited capacity for sensing, processing, and actuation, but together, the robots as a swarm benefit from massive parallelism, redundancy, and adaptability. Proof-of-concepts in swarm robotics have showcased the construction of nests~\cite{werfel2014designing}, the formation of shapes~\cite{rubenstein2014programmable,slavkov2018morphogenesis,sun2023mean}, swarms of  UAVs flying in formation~\cite{viragh2014flocking,coppola2020survey}, swarms of micro-robots for transport and manipulation\cite{heuthe2024counterfactual}, achieving collective aggregation~\cite{saintyves2024self,aubert2017evolutionary,bray2023recent}, performing social learning~\cite{bredeche2022rspt} and behaving as smart materials~\cite{devlin2025material}. 

Yet, it remains unclear how a swarm of robots with limited sensing and computation can solve global problems that requires each robot to obtain precise knowledge of the state of the whole swarm.
This is the topic of this paper, for which we tackle a first instance of distributed \textbf{shape identification}: the ability of a robot swarm to estimate a spectral descriptor of its interaction graph and use it to classify the arena among predefined candidate shapes in a fully decentralized way.
The typical scenario is pictured in Figure~\ref{fig:overview}, showing a robot swarm being deployed in a bounded environment. The swarm first spreads to cover the area, and robots work together through pairwise interactions to estimate the geometric shape of the swarm, which is then ultimately known to each robot.

Swarm robotics poses a unique problem compared to more classic collective robotic setups, as robots are rudimentary and often lack basic equipment to measure their position or detect their environment (\textit{e.g.,} limited availability or reliability of inertial measurement units, wheel encoders, camera, GPS, etc.) and to perform advanced computations (limited memory and low computing power). Under such conditions, powerful methods such as Simultaneous Localization and Mapping (SLAM) are not an option~\cite{wang2020shape}, and we are forced to rely on robot-to-robot signaling~\cite{cazenille2025signalling} to estimate the global shape of the swarm.
The expected outcome is that all robots of a swarm ultimately reach a consensus on a spectral descriptor, here $\lambda_2$, which can then be mapped to a shape label when a predefined library of candidate arenas is available.

In nature, biological swarms are known to solve various types of geometric problems relying solely on limited perception and mobility in noisy environments~\cite{reina2021collective,reynolds1987flocks,vicsek1995novel,vicsek2012collective}. Ants find the shortest path between their nest and a food source~\cite{detrain2008collective,bradbury1998principles}
and termites build 3D networks of conduits to ventilate their nest~\cite{wilson2000sociobiology}. Schools of fish or flocks of birds move in a synchronized pattern~\cite{attanasi2014information},
and a colony of bees can split into two colonies to grow~\cite{camazine2020self}.
Cells depend on the spatial gradients of chemicals
to establish a universal positioning system~\cite{turing1990chemical}. Although details vary, these global patterns emerge from the local diffusion of information between swarm agents. This suggests an underlying principle of information diffusion, whereby complex spatial problems can be solved without explicit mathematical computation.

While distributed shape identification has not yet been solved in swarm robotics, several important works explored how local signaling can be used to reveal the global geometry through purely decentralized interactions. These methods broadly fall into two categories: building an emergent coordinate system and using pairwise diffusion methods. Firstly, several works rely on building a coordinate system from a few well-positioned robots to perform shape generation~\cite{rubenstein2014programmable,gauci2018programmable}, to incrementally build a mesh of robots to explore an unknown environment~\cite{lee2014exploration,lee2016structured} or to perform boundary detection based on local topological properties~\cite{mclurkin2009distributed,kuang2019boundary}. A mandatory technical requirement is that each robot must be able to deduce its position relative to the others, which means that (at the very least) distance to at least 3 neighbors must be accessible for unambiguous localization, as long as they are not collinear. While this is generally (though not always) the case, the resulting localization process can be subject to noise and may accumulate errors as the swarm grows.

Secondly, several authors explored diffusion models to extract macroscopic information at the local level. Diffusion, in this case, relies on pairwise exchange of information among close enough robots. This has been used for various purposes, including forming Turing patterns~\cite{slavkov2018morphogenesis}, consensus formation~\cite{valentini2015self,talamali2021less}. A foundational contribution from~\cite{olfati2007consensus} established a link between consensus formation and spectral graph theory, using the so-called Laplacian operator~\cite{evans2022partial}, a ubiquitous measure in physics~\cite{arendt2009weyl}, as a means to monitor the speed of convergence of the whole swarm towards consensus.

Going a step further, later works explored how to use the Laplacian to achieve distributed formation control in a robot swarm~\cite{lin2014distributed,demarina2021distributed,origane2022wave}. However, these works concentrate on achieving a specific shape formation, and each robot is provided with both the target and its final position, which makes it possible to compute locally an error between the robot's current and final locations. Work in swarm robotics is still far from addressing geometric problems as solved by biological swarms (Supplementary Sec.~1).

In order to perform shape analysis in the context of a collective of robots, we draw our inspiration from the work of Mark Kac's 1966 paper~\cite{kac1966can} on ``hearing the shape of a drum'' using spectral shape analysis~\cite{spielman2012spectral,chung1997spectral}. In this seminal paper and the studies that followed, it is revealed that the natural vibration frequencies of a drumhead (\textit{i.e.,} the spectrum of the Laplacian operator for a particular shape) encodes much of the geometric information~\cite{reuter2006laplace,belkin2006manifold,rustamov2007laplace}. 
However, swarm robotics faces a unique challenge: shape estimation must be achieved in a fully distributed way, since each robot interacts only with its immediate neighbors. Without access to a global view of the swarm, every robot must locally estimate the eigenvalues of the Laplacian—a task that has so far remained elusive due to the complex mathematical nature of the Laplacian operator. This will be the main contribution of this paper: \textbf{we provide a robust method based on spectral shape analysis for distributed estimation of spectral properties}, which can be used to reveal the swarm shape, a global structural property, to each robot of the swarm. 

In this sense, ``hearing the shape'' refers to the spectral viewpoint inherited from Kac’s question: geometric information is accessed through Laplacian spectral quantities. Our contribution is to translate this viewpoint into swarm robotics. The swarm does not reconstruct an arbitrary boundary from a global map; instead, it physically estimates spectral information generated by its own local interaction graph. 

In the method we propose, the robot swarm collectively infers the geometric shape of its environment through the distributed computation of Laplacian eigenvalues. Each robot locally exchanges information with its neighbors to estimate a component of the arena’s spectral fingerprint. 
In the present work, we focus on the first non-trivial and experimentally accessible descriptor, $\lambda_2$, as a minimal proof of concept, as it reflects the global geometry of the space. 
As information diffuses through the swarm, higher-frequency modes quickly decay, allowing each robot to locally isolate the dominant component associated with the spatial structure of the arena. As the rate at which this component stabilizes depends directly on the swarm’s global geometry, each robot is able to extract a unique numerical signature to the swarm’s spatial configuration. Provided the swarm is well distributed in space, each individual can thus autonomously extract the geometric signature of the arena.

In the following, we present our spectral method for distributed shape identification, and derive a universal scaling law from dimensional analysis. We confirm our theoretical result with numerical simulations and provide an experimental validation with two swarm robotic platforms with different characteristics: Kilobots~\cite{rubenstein2012kilobot} and Pogobots~\cite{loi2025pogobot}.

\begin{figure*}[p!]
\begin{center}
\includegraphics[width=1.00\textwidth]{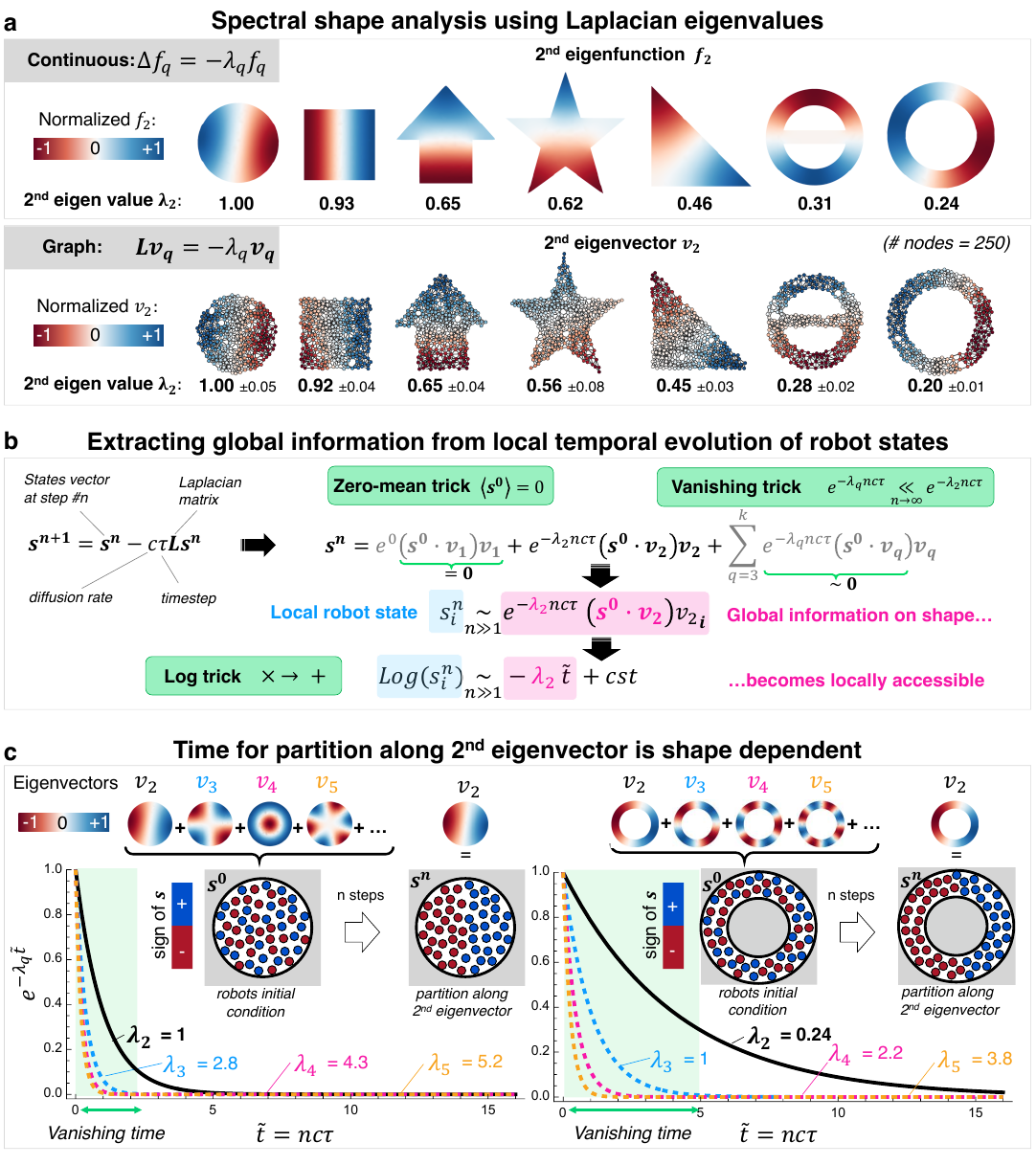}%
\caption{
\textbf{Framework of spectral swarm robotics.} {\small
\textbf{a:}~Comparison of results from the continuous and graph Laplace operators over 7 geometric shapes. Color codes correspond to the local components of the second eigenfunction (in the continuous case) and eigenvector (in the graph case). The graph cases have 250 nodes randomly distributed in the shapes. $\lambda_2$ is averaged over 64 runs with different seeds and shown with its standard deviation. To easily compare the continuous and graph cases, values of the second eigenvalue $\lambda_2$ are normalized to have a value of 1 for disk arenas. Images are re-scaled for better visualization, computations are performed with arenas of the same surface.
\textbf{b:}~Main mathematical tricks allowing each robot to extract $\lambda_2$ from diffusion of information.
\textbf{c:}~Exponential decays associated to first five eigenvalues. After a vanishing time, only $e^{-\lambda_2 \tilde{t}}$ prevails. Schematics show how a random initial condition, consisting of a superposition of multiple modes converges towards a partition along the 2nd eigenvector. 
}}
\label{fig:workflow}
\end{center}
\end{figure*}

\section*{Methods}

We introduce \textbf{spectral swarm robotics} - adapting spectral approaches (details are provided in Supplementary Sec.~1) to be computed in a distributed way on robot swarms. 
The diffusion of information within a certain shape is strongly influenced by its geometry. Specifically, the spectral signature of the Laplacian (\textit{i.e.,} its eigenvalues) suggests a way for a robotic swarm to sense the geometry of its arena with diffusion (Fig.~\ref{fig:workflow}a, Supplementary Sec.~2). 
Therefore, robots can learn about the geometry of their arena by passing information to each other to emulate the Laplacian operator: the robot swarm simulates heat diffusion as each robot adjusts its internal state to the average temperature (internal state) of its neighbors. By looking at \textbf{how quickly} the swarm gets to a stable state (\textit{i.e.,} by locally measuring the relaxation time to the equilibrium), we can reliably estimate the eigenvalues locally. As a result, each robot can determine the spatial configuration of the entire swarm, exploiting the fact that some shapes are more conducive to diffusion than others (\textit{e.g.,} heat can diffuse uniformly across two dimensions in a disk, but it is predominantly confined to one dimension in an elongated ellipse).
Robots then move to change their neighbors and repeat these measurements several times to reduce variance and eventually converge to an estimated eigenvalue. While, for a finite swarm of robots, this average value would not necessarily converge to the true value of the continuous case, it can be used as a fingerprint that helps the robots to sense the surroundings of the whole swarm.

In particular, the second smallest eigenvalue ($\lambda_2$) of the Laplacian, also known as ``algebraic connectivity'' or ``Fiedler value''~\cite{fiedler1989laplacian} influences the swarm's connectivity and synchronization speed, and is often used as a fingerprint by the spectral shape analysis community~\cite{reuter2006laplace,wang2019intrinsic}\footnote{The first smallest eigenvalue $\lambda_1$ is always equal to 0.} (Supplementary Sec.~2.3). 

Figure~\ref{fig:workflow}a shows, for seven different arenas, the second smallest eigenfunction $f_2$ or eigenvector $v_2$ of the continuous and graph Laplace operators as well as the corresponding eigenvalue $\lambda_2$. The selected shapes are sufficiently different to have distinct $\lambda_2$ values. The graph Laplacian can be created by placing nodes at random positions within the shape and connecting neighboring nodes. The graph eigenvector $v_2$ approximates well its continuous counterpart $f_2$ and, once normalized, the values of $\lambda_2$ are very similar in both cases. $\lambda_2$ is higher in shapes with higher connectivity, like the disk, and decreases with the presence of nodes with lower connectivity acting as information bottlenecks, such as in the case of the annulus. 

\subsection*{Mathematical foundation}

Our method decentralizes the classification of the shape of an arena.
We now provide a theoretical description of our proposed spectral swarm robotics algorithm for shape analysis, as illustrated in Fig.~\ref{fig:overview}.

The process begins with an optional step where $N$ robots randomly walk in the arena until they are evenly distributed. They then identify their neighbors, defined as robots located within their range of communication $\sigma$.
For typical robotic experiments, $\sigma$ is several times the size of a single robot, meaning that a robot only sees a fraction of the swarm at any time (which may consist of tens or hundreds of robots).
The set of neighbors forms a graph $\mathcal{G}$ that captures the swarm's connectivity and the arena's geometry (by convention a robot is not its own neighbor).

The robots then engage in rounds of diffusion of information, as follows.
Robots initially set their internal state $s_i^{0}$ to $-1$ or $1$ (a state chosen to have a zero mean, which is conserved during diffusion) and, at each round of diffusion, they update their internal state according to the law of diffusion on a graph~\cite{baxter1984equivalence,freidlin1993diffusion} (Supplementary Sec.~2.1):
\begin{equation} \label{eq:diff}
s_i^{n+1}=s_i^{n}+c\tau\sum_{j=1}^{N} A_{ij} (s_j^{n}-s_i^{n})
\end{equation}
with $n$ the current time step.

This equation can be rewritten in a matrix form (cf Fig.~\ref{fig:workflow}b):
\begin{equation}
\textbf{s}^{n+1}=\textbf{s}^{n} - c \tau \textbf{L} \textbf{s}^{n}
\end{equation}

where $\textbf{L}$ is the Laplacian matrix of the graph, $c$ is the diffusion rate across edges of the graph and $\tau$ is time between two steps of diffusion. Note that $c$ includes a notion of distances between nodes; here we assume that all edges have the same weight (distance is 1, \textit{i.e.,} $\mathcal{G}$ is unweighted).
On the diagonal, $L_{i,i}$ is the number of neighbors detected by the robot $i$, and off the diagonal, $L_{i,j}=-1$ if $i$ and $j$ are neighbors, and $L_{i,j}=0$ otherwise.

Assuming $c\tau$ is small, this time evolution can be approximated analytically by taking the exponential of the Laplacian matrix. 
\begin{equation}
\textbf{s}^{n}=e^{-n c\tau \textbf{L}} \textbf{s}^{0}
\end{equation}

Since the Laplacian matrix $\textbf{L}$ is symmetric and positive semidefinite, its eigenvalues $\lambda_i$ are real and positive, with $\lambda_1=0\leq\lambda_2\leq\lambda_3\leq\cdots$. By the spectral theorem, the time evolution of $\textbf{s}^{n}$ is given by its projection on the eigenvectors $\textbf{v}_i$ of $\textbf{L}$:
\begin{equation}
\textbf{s}^{n}=\sum_{q=1}^{N} e^{-\lambda_q n c\tau} (\textbf{s}^{0}\cdot \textbf{v}_q) \textbf{v}_q
\end{equation}

where $q$ indexes the $N$ eigenmodes of $\textbf{L}$, with eigenvalue $\lambda_q$ and associated eigenvector $\textbf{v}_q$, and $\textbf{s}^{0}\cdot \textbf{v}_q$ is the projection of the initial state $\textbf{s}^{0}$ onto $\textbf{v}_q$. In order for robots to obtain $\lambda_2$ locally from this global equation, we use 3 mathematical tricks: the ``zero-mean trick'', the "vanishing trick" and the ``log trick'' (Fig.~\ref{fig:workflow}b). The first two tricks remove terms related to other eigenvalues than $\lambda_2$ while the log trick allows each robot to extract $\lambda_2$ locally.

The ``zero-mean trick'' removes the first term of the sum as follows. The internal state $s^n_i$ of each robot converges to the mean of the initial $\textbf{s}^{0}$, which corresponds to the state where the information is fully mixed and the robots share the same internal state. Since by design the mean $\langle\textbf{s}^{0}\rangle$ was set to be $0$, the first term of the sum disappears. Indeed, the first eigenmode is $\textbf{v}_1=1/\sqrt{N} (1,1,\ldots,1)$, and thus the projection of the initial state $(\textbf{s}^{0}\cdot \textbf{v}_1)=\sqrt{N}\langle\textbf{s}^{0}\rangle=0$. 

The ``vanishing trick'' allows us to remove all terms of the sum for $q\ge3$. In fact, for a large enough n,  $e^{-\lambda_q n c\tau}\ll e^{-\lambda_2 n c\tau}$, such that after some time, $\textbf{s}^{n}$ decays exponentially along $\textbf{v}_2$:
\begin{equation} \label{eq:dominant_eigenmode}
\textbf{s}^{n}=e^{-\lambda_2 n c\tau} (\textbf{s}^{0}\cdot \textbf{v}_2) \textbf{v}_2
\end{equation}
This is illustrated in Fig.~\ref{fig:workflow}c: the initial state corresponds to a sum of eigenvectors, that each decays to 0 at different speed (eigenvalues as exponents). We refer to the ``vanishing time'' as the time required for all eigenvectors $q\ge3$ to disappear, leaving only $\textbf{v}_2$.

The ``log trick'' allows us to separate the $e^{-\lambda_2 n c\tau}$ part that can be measured locally by robots from the $(\textbf{s}^{0}\cdot \textbf{v}_2) \textbf{v}_2$ part that requires global knowledge. We then apply the logarithm to that formula to reformulate the time-evolution of $s$ as \textbf{a simple locally-computable linear equation:}

\begin{equation} \label{eq:eqST}
\log(|s_i^{n}|)={-\lambda_2 n c\tau}+C  = {-\lambda_2 \tilde{t}} + C
\end{equation}
with $\tilde{t} = n c \tau$. This yields $-\lambda_2$ the slope of $\log(|s_i^{n}|)$ and $C$ its intercept. 

\textbf{This equation is the crux of our strategy to identify the shape of the robot arena.} The shape controls the rate at which information diffuses in the robot graph through $\lambda_2$, which is known in the literature as the algebraic connectivity of the neighborhood graph~\cite{fiedler1989laplacian,molitierno2016applications}.
The magnitude of $\lambda_2$ reflects the connectivity of the robot graph: the more paths there are between any two robots, the larger $\lambda_2$ tends to be, and the faster the robots synchronize their internal state and converge to their collective mean. Thus, a swarm of robots dispersed in a disk-shaped arena is expected to have a larger connectivity than a swarm dispersed in an annulus with the same surface, and information is expected to spread faster in the former case.
Of note, if the robot graph is not connected (there exists at least one pair of robots without a path between them in the graph), then $\lambda_2=0$ and the robotic swarm is partitioned into distinct clusters (``nodal domains'') that synchronize internally, but not globally. This can happen even for connected shapes when the field of perception or the number of robots is too small, which opens holes in the neighborhood graph and prevents information from fully flowing in the swarm. The associated eigenvector $\textbf{v}_2$ also plays an important role: it is known as the Fiedler vector and the sign of its coordinates define a partition of the graph, as seen in Fig.~\ref{fig:workflow}a where color codes correspond to local components of the second eigenvector. 

From Eq.~\ref{eq:eqST}, each robot individually estimates $\lambda_2$ by linearly fitting the exponential decay rate -- after a suitable time -- of its internal variable $s^n_i$ to $0$. Each robot thus gets a slightly different estimate of $\lambda_2$, which is then averaged across the swarm after each iteration to obtain a more accurate value.
Still, this estimate remains noisy since the number of robots is finite.
To refine this estimate, after each iteration, the robots perform a random walk within the arena, stop, and start again the process of estimating $\lambda_2$ from a new communication graph. This repeated sampling reduces the influence of irregular local configurations: the method does not require a perfectly uniform distribution at each iteration, but cumulatively averages the spectral estimates obtained from multiple spatial configurations.
For sets of shapes that are not pathological (\textit{i.e.,} with distinct $\lambda_2$), it is, in principle, possible to discriminate them provided enough iterations, since our Monte Carlo estimate converges to $\lambda_2$ as $1/\sqrt{N}$.

\begin{figure*}[h]
\begin{center}
\includegraphics[width=0.99\textwidth]{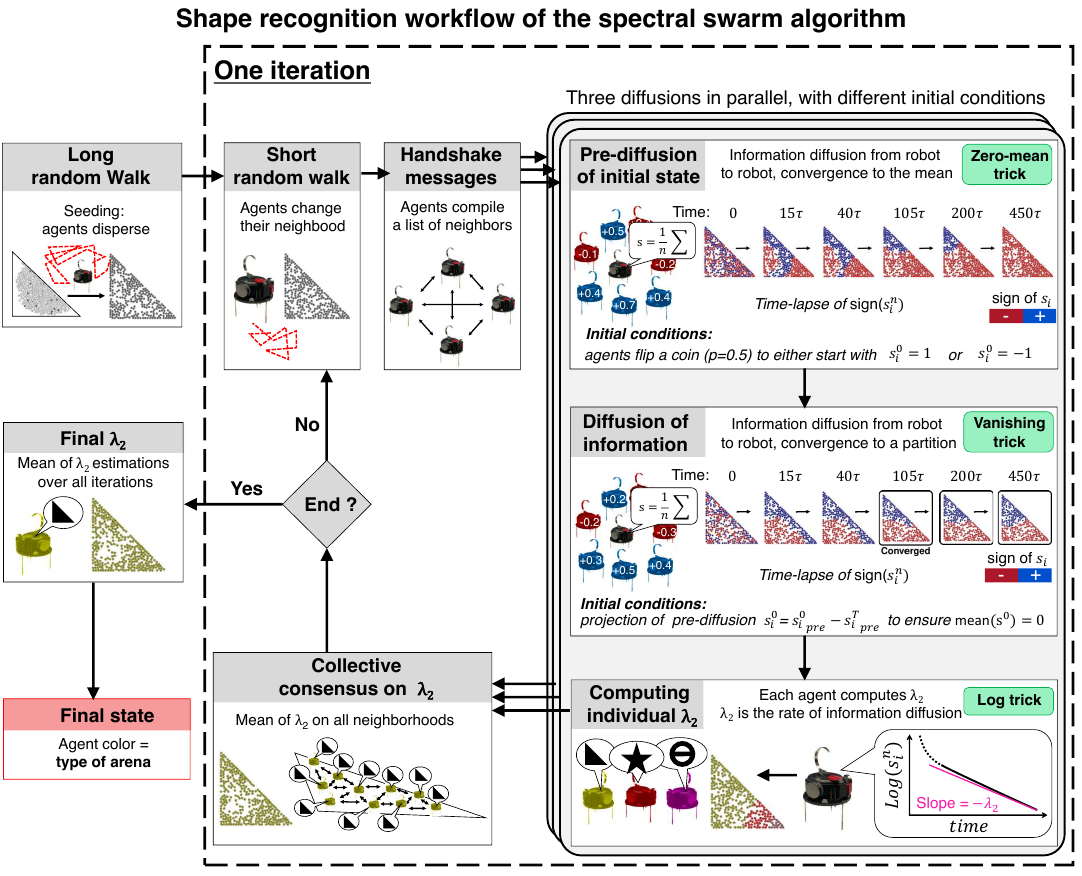}%
\caption{Workflow of the spectral swarm algorithm for decentralized shape classification. Random walks and handshake messages allow for robots to change and identify their neighbors. Pre-diffusion ensures the applicability of the "zero-mean trick": from a random initial condition Kilobots will diffuse their internal state, resulting in an homogeneous state, the mean. This value will be subtracted to the initial condition of the diffusion step. The Diffusion step ensures the applicability of the "vanishing trick": as depicted in Fig2c, robots converge to a partition along the 2nd eigenvector. The "log trick" allows then to estimate $\lambda_2$ directly from the convergence rate of the diffusion process . The robots reach a consensus on this value across the entire swarm and show a different LED color depending on its value, resulting in a classification decision.}
\label{fig:workflowAlgo}
\end{center}
\end{figure*}

\begin{figure*}[p!]
\begin{center}
\includegraphics[width=1.00\textwidth]{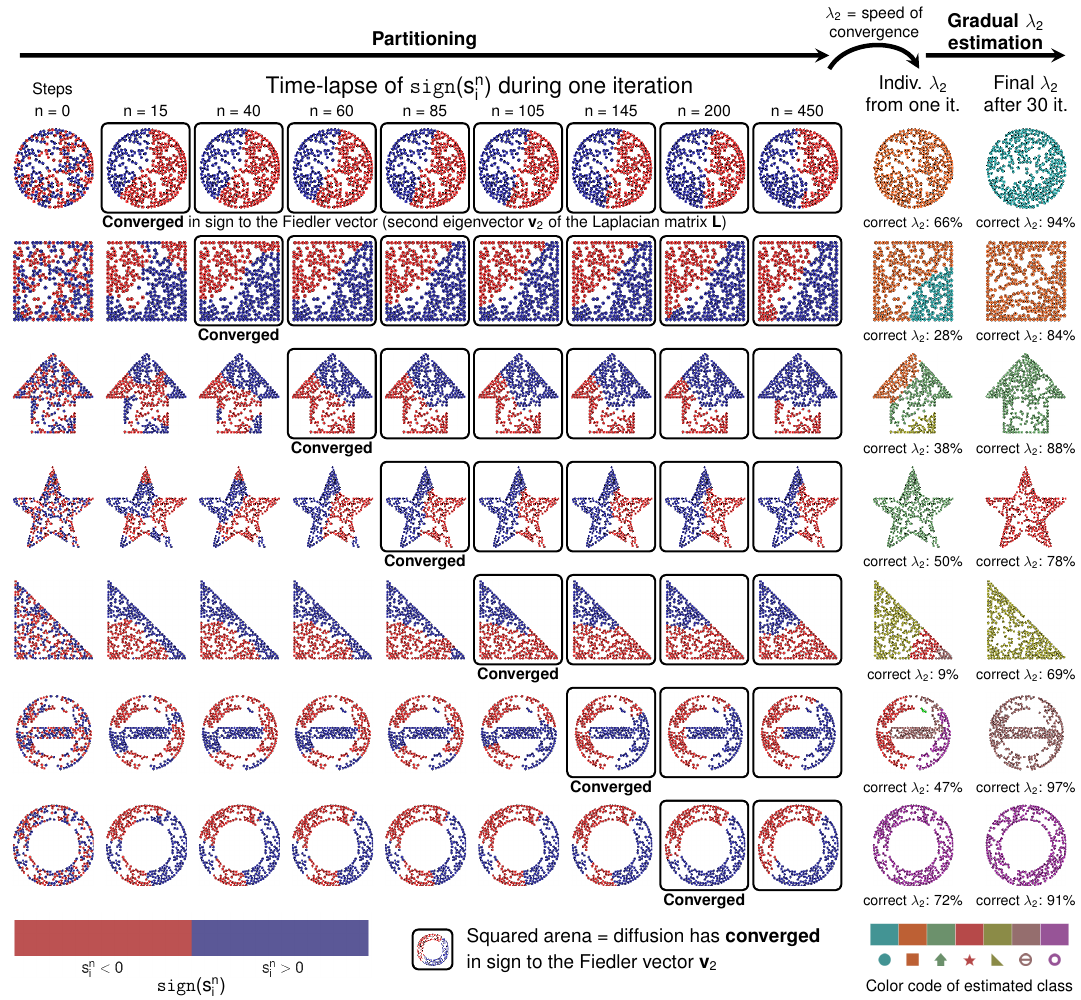}%
\caption{\textbf{Representative time-lapses of the behavior of the proposed algorithm applied to 7 arenas in simulations.} {\small
\textbf{Left:}~diffusion stage, resulting in a partitioning of the shapes (color code: sign of internal state $s_i^n$). The second eigenvalue $\lambda_2$ of the Laplacian matrix of the communication graph between robots corresponds to the convergence rate of diffusion: the information diffusion converges at different rates depending on the topology of the arena.
\textbf{Right:}~in turn, robots use the observed convergence rate to compute a local estimate of this value iteratively refined at each iteration of the algorithm. After 30 iterations, the estimations have converged and formed a consensus. Based on the value of $\lambda_2$, the robots display a color code corresponding to the detected arena shape. Images are re-scaled for visualization; during simulations, arenas surface are normalized to be $500000\texttt{mm}^2$, with $\tau = 1/15$ seconds corresponding to the amount of time between two steps of diffusion. We use the same parameters as regime r3 of~Fig.~\ref{fig:phase_diagrams}, with $N = 300$ robots and $\sigma = 85 \texttt{mm}$. The percentages of correct $\lambda_2$ are computed over 64 runs.
}}
\label{fig:timelapses}
\end{center}
\end{figure*}

\subsection*{Algorithmic description}
Here we propose to translate the theoretical approach presented earlier into a multi-agent algorithm that can be implemented as a swarm robotic controller in simulations or experiments.
The robot swarm collectively classifies the shape of their arena based on distributed spectral shape analysis.
The spectral swarm robotics algorithm is summarized in Fig.~\ref{fig:workflowAlgo}. It is also formally described in Alg.~\ref{alg:main}% and Alg.~\ref{alg:diffusion} 
\footnote{The C source code used for simulations and experiments can be found in a supplementary information archive and in the repository:~\url{https://github.com/leo-cazenille/spectral-swarm-robotics}.}. 
This algorithm is designed to be implementable on real robots with limited sensing and computation (no localization, mapping, wall sensing, global communication, or centralized computation). It makes the following practical assumptions: 1) robots are roughly temporally synchronized, or can locally resynchronize their internal clocks through exchanged messages; 2) robots can communicate with neighbors across a given field of perception; 3) robots can display ongoing results, such as their estimated value of $\lambda_2$, through a visual output such as an on-board LED; and 4) robots can perform simple scalar computations, including absolute values and logarithms, either directly or through lightweight approximations such as lookup tables.

Robots are assumed to be initially placed packed at the center of the arena. Our approach requires robots to be distributed in all parts of the arenas, so the spectral swarm robotics algorithm starts by making the robots disperse and move to random positions in the arena (``initial seeding") by using the ``run-and-tumble" algorithm.

After the seeding stage, the algorithm starts its main loop. In each iteration $a$ of the main loop, the robots will compute their own estimation of $\lambda_2$, termed here as $\texttt{Indiv}\lambda_{2_i}(a)$ for each robot $i$.
During one iteration, the robots will sequentially perform the following operations:

\begin{enumerate}
    \item Robots will move with a short budget of time, in order to change their neighbors. This allows the algorithm to change its initial conditions and to start with a different communication graph from iteration to iteration. After the allotted time has elapsed, robots will stop moving and stay immobile for the rest of the iteration.
    \item Optionally, robots will use a handshaking system to determine what are their ``stable" neighbors, \textit{i.e.,} the list of neighbors $k$ of a focal robot $i$ such as, most of the time, $i$ can broadcast messages to $k$ and $k$ can broadcast messages to $i$. That process is useful if there is variability in the field of perception $\sigma$ of the robots, due to small physical differences in IR captors or environmental conditions. This can result in cases where robot $i$ is able to reach $k$ while $k$ cannot reach $i$. However, our algorithm relies on the Laplacian of the communication graph which is expected to be symmetrical. 
    \item Robots compute $P = 3$ diffusion sessions in parallel, each with different initial conditions. This process allows each robot to compute $P$ different local estimates of $\texttt{Indiv}\lambda_{2_i}(a,d)$ with $d$ the index of the diffusion session performed in parallel. Each robot then averages its estimates to obtain a single local value $\texttt{Indiv}\lambda_{2_i}(a)$.
    \item Local estimates are broadcast on all neighborhoods so that robots can reach a consensus towards a global estimate $\texttt{Consensus}\lambda_{2}$. This is achieved through a collective averaging process: robots iteratively average their local estimate with those of their neighbors~\cite{kempe2003gossip}.
    \item Robots check whether the total allotted time has been exceeded. If not, they proceed to the next iteration.
\end{enumerate}

The algorithm will sequentially compute $I$ iterations. The $\texttt{Consensus}\lambda_{2_i}$ values obtained at each iteration are averaged to compute the overall $\lambda_{2_i}$ estimation, termed $\texttt{Final}\lambda_{2_i}$. The robots will set the color of their LED depending on the value of $\texttt{Final}\lambda_{2_i}$: for a given set of arenas, we define a corresponding set of LED colors, respectively associated in turn to a set of centroid values of $\lambda_2$ values.
Then, the predicted class of agent $i$ is decided using a nearest-centroid classifier based on its current value of $\texttt{Final}\lambda_{2_i}$. This final labeling step is supervised: the robots do not discover arbitrary shapes from scratch, but classify the arena among a predefined library of candidate shapes whose reference $\lambda_2$ centroids are known beforehand.

\algblockdefx[ParallelFor]{ParallelFor}{EndParallelFor}%
    [1]{\textbf{parallel-for} #1 \textbf{do}}%
    {\textbf{end parallel-for}}

\begin{algorithm*}[h!]
\scriptsize
\begin{algorithmic}
\State \texttt{run\_and\_tumble(seeding\_duration)} \Comment{Seeding, long random walk}
\item[]
\For {$a$ in $1..I$} \Comment{Perform $I$ iterations}
    \State \texttt{run\_and\_tumble(short\_rw\_duration)} \Comment{Short random walk at the beginning of each iteration}
    \item[]
    \State $\texttt{neighbors} \gets \texttt{handshake()}$ \Comment{Identify stable neighbors}
    \item[]
    
    \LineComment{Start diffusion of information with neighboring robots}
    \ParallelFor {$d$ in $1..P$} \Comment{Perform $P$ diffusion sessions in parallel}
        \State $s_i^0(a,d) \gets \texttt{random\_choice(-1, 1)}$      \Comment{Initialization, with random values: either -1 or 1}
        \State $s_i(a,d), \texttt{Indiv}\lambda_{2_i}(a,d) \gets \texttt{diffusion(neighbors}, s_i^0(a,d)\texttt{)}$   \Comment{Pre-diffusion}
        \State $s_i^0(a,d) \gets s_i^0(a,d) - s_i^T(a,d)$      \Comment{Initialization from pre-diffusion values to ensure $\sum_i s_i^0(a,d) = 0$}
        \State $s_i(a,d), \texttt{Indiv}\lambda_{2_i}(a,d) \gets \texttt{diffusion(neighbors}, s_i^0(a,d)\texttt{)}$   \Comment{Diffusion}
    \EndParallelFor
    \State $\texttt{Indiv}\lambda_{2_i}(a) \gets \frac{\sum_{d}^P \texttt{Indiv}\lambda_{2_i}(a,d)}{P}$  \Comment{Average $\lambda_{2_i}$ values estimated through the $P$ diffusion sessions}

    \item[]
    \LineComment{Reach consensus across the swarm, through collective averaging of $\lambda$}
    \State $\texttt{Consensus}\lambda_{2_i}(a) \gets \texttt{coll\_averaging(neighbors,} \texttt{Indiv}\lambda_{2_i}(a)\texttt{)}$
\EndFor

\item[]

\State $\texttt{Final}\lambda_{2_i} \gets \frac{\sum_{a}^{I} \texttt{Consensus}\lambda_{2_i}(a)}{I}$ \Comment{Average $\texttt{Consensus}\lambda_{2_i}$ across all iterations}

\State $\texttt{set\_robot\_LED\_color(Final}\lambda_{2_i}\texttt{)}$

\item[]
\item[]

\Function{diffusion}{neighbors, $s_i^0$}
    \LineComment{Start diffusion of information with neighboring robots}
    \State $t_i^0 \gets 0$
    \For {$j$ in $1..T$} \Comment{Perform $T$ diffusion steps, each with a duration of $\texttt{timeout\_diff}$}
        \State $\texttt{initial\_time\_diff}\gets \texttt{current\_time\_ticks()}$ 
        \State \texttt{broadcast($s_i^{j-1}$)} \Comment{Allow neighboring robots to retrieve it}
        \LineComment{Retrieve all $s_k$ from neighbors, until timeout}
        \State $c_j \gets 0$
        \While {$\texttt{current\_time\_ticks()} - \texttt{initial\_time\_diff} < \texttt{timeout\_diff}$}
            \State $s_k^j \gets \texttt{retrieve\_value\_from\_next\_neighbor(neighbors)}$
            \State $c_j \gets c_j + (s_k^j - s_i^j)$
        \EndWhile
        \State $s_i^j \gets s_i^{j-1} + \tau c_j$ \Comment{Update $s_i$ according to the law of diffusion}
        \State $t_i^j \gets \tau j$
    \EndFor
    \item[]
    \LineComment{Compute local approximation of $\lambda_2$ by using ordinary least square (OLS)}
    \LineComment{$B = $ Number of diffusion steps to ignore}
    \State $\mathtt{Indiv}\,\lambda_{2_i} \gets
-\frac{(T-B)\displaystyle\sum_{j=B}^{T}\left(t_i^{\,j}\,\log|s_i^{\,j}|\right)
-\left(\displaystyle\sum_{j=B}^{T} t_i^{\,j}\right)\left(\displaystyle\sum_{j=B}^{T}\log|s_i^{\,j}|\right)}
{(T-B)\displaystyle\sum_{j=B}^{T}(t_i^{\,j})^{2}
-\left(\displaystyle\sum_{j=B}^{T} t_i^{\,j}\right)^{2}}$
    \State \textbf{return} $s_i, \texttt{Indiv}\lambda_{2_i}$
\EndFunction
\end{algorithmic}
\caption{Main algorithm, executed on each robot $i$.}
\label{alg:main}
\end{algorithm*}

\section*{Results}

\subsection*{Shape classification}
We tested our method using physical solid-body simulations with up to 550 robots and classified seven geometric shapes. The simulations were computed using the Kilombo~\cite{jansson2015kilombo} simulator (more details in the experimental validation subsection below).

Figure~\ref{fig:timelapses} shows the behavior of the algorithm in simulation. The diffusion processes relaxed at different speeds depending on the shape, as in the theoretical results of Fig.~\ref{fig:workflow}c, with speed corresponding to $\lambda_2$. Thus, $\lambda_2$ was a nearly unique fingerprint for classifying the selected shapes. It could be estimated locally, but individual estimates were noisy. Averaging the estimates over several iterations of the algorithm increased their accuracy and allowed the robots to classify shapes from averaged $\lambda_2$ estimates.
The time-lapses of the diffusion process in Fig.~\ref{fig:timelapses} confirm the evolution of the sign and magnitude of state $\textbf{s}^{n}$. The swarm started from a highly heterogeneous $\textbf{s}^{n}$, where typically half of the neighbors of each robot $i$ had an internal state with the opposite sign of $s_i^0$. With time, diffusion smoothed local disparities in sign: robots homogenized their individual states with their neighbors, and the swarm eventually partitioned into two stable nodal domains of opposite signs. The partition was reproducible between runs for shapes without symmetries (such as the arrow), because the eigenvalues were not expected to be degenerate ($\lambda_2<\lambda_3$, with a one-dimensional eigenspace associated with $\lambda_2$). The absolute value of each $s_i$ decayed exponentially after sufficient time, although the time needed to reach this regime depended on the spatial position of the robot. The signs of the states of robots near the interface were expected to resolve later than those of robots farther from the interface.
After diffusion, the robots computed local estimates of $\lambda_2$ and predicted the arena shape from those estimates. The local estimates varied with initial conditions and differed between iterations because robots moved and changed neighbors at the beginning of each iteration. After 30 iterations, a consensus emerged, and the probability of correctly identifying the shape increased substantially (for example, from $66\%$ to $94\%$ for the disk). The algorithm's accuracy measured how often the robots correctly classified the arena in a distributed manner. Supplementary Fig. S2 shows the improvement in accuracy from one to 30 iterations.

\begin{figure*}[p!]
\begin{center}

\includegraphics[width=0.80\textwidth]{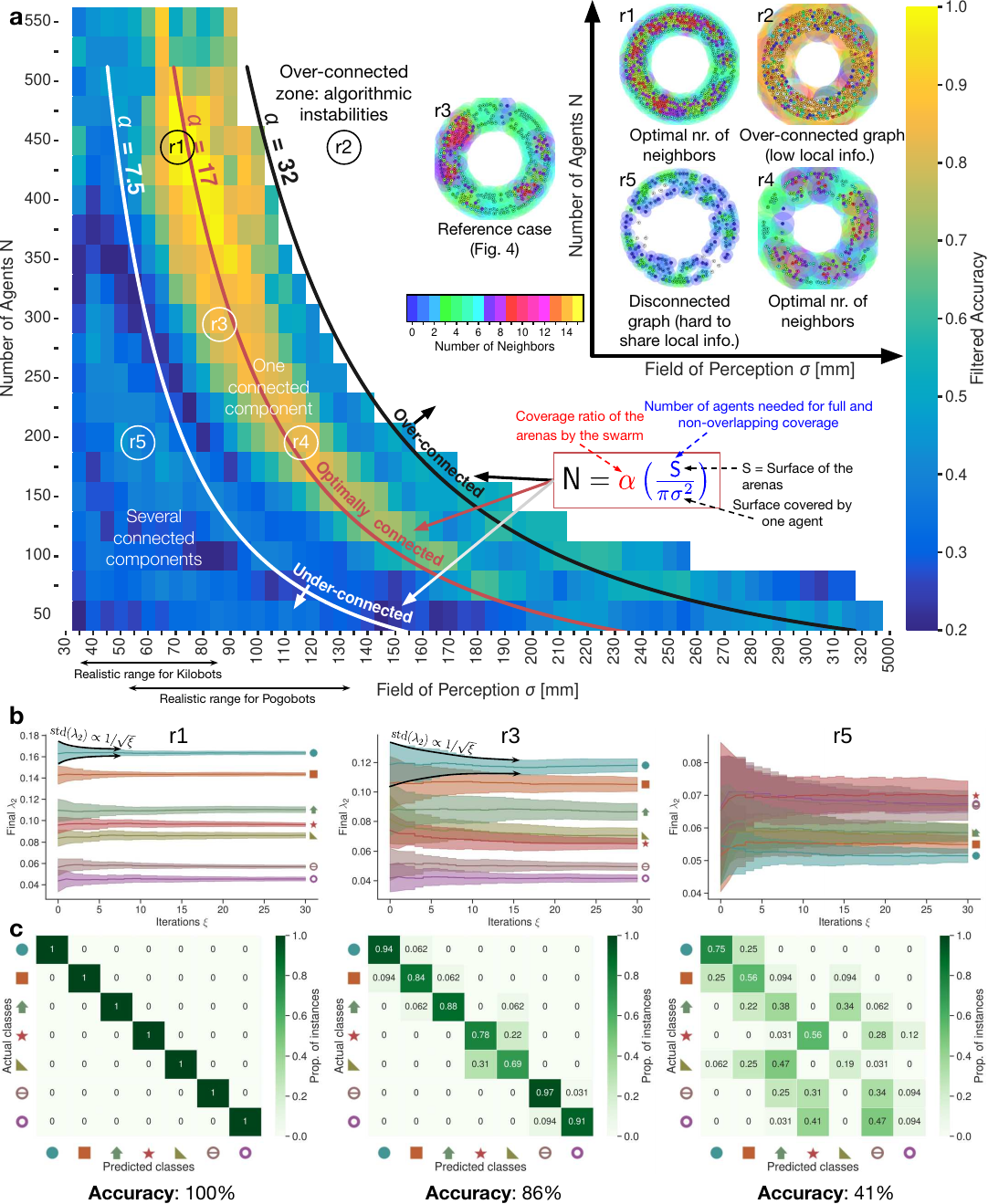}%
\caption{\textbf{Classifying regimes of the system computed in simulations over 64 runs.} \small{
\textbf{a:}~Influence of the number of robots $N$ and of the field of perception $\sigma$ on the accuracy. We filter the accuracy score to only show results where the simulated diffusion sessions converge: empty bins correspond to results with more than $50\%$ of cases with algorithmic instabilities from overconnected graphs and numerical errors (\textit{i.e.,} without exponential decay of $s_i^{n}$). We identify five representative regimes: r1 to r5. Examples of regime r3 behavior are shown in Fig.~\ref{fig:timelapses}. The best-performing results are found on the hyperbola $N = 17 S / (\pi \sigma^2)$. Top-right corner: examples of robot distribution in the annulus arena; colored disks around each robot represent their field of perception $\sigma$.
\textbf{b:}~Evolution of estimated $\lambda_2$ values over 30 iterations of the algorithm, for each arena. Line colors correspond to the color code in Fig.~\ref{fig:timelapses}. Only r1, r3, r5 are shown (r4 is similar to r3, and r2 is an overconnected case). The order of $\lambda_2$ values for each arena is generally conserved across regimes, except in r5, where the order is inverted compared to r1, r3, r4.
\textbf{c:}~Confusion matrices and accuracy scores for regimes r1, r3, r5, showing the performance of the system to accurately classify the shapes. Each row represents the instances of an actual class (\textit{i.e.,} the shapes the robots are situated in) while each column represents the instances of a predicted class (\textit{i.e.,} the shapes the robots report using LED color). The diagonal elements of the matrix (from top-left to bottom-right) represent the number of points for which the predicted class is equal to the true class, \textit{i.e.,} the correct predictions; perfect accuracy corresponds to the identity matrix. Off-diagonal elements in the matrix describe the misclassifications, \textit{i.e.,} the instances where the model predicted a class that differs from the true class. 
}}
\label{fig:phase_diagrams}
\end{center}
\end{figure*}

\subsection*{Universal scaling law}

We next derived a scaling law for the swarm through dimensional analysis. Here, the field of perception $\sigma$ controls the granularity and ``resolution'' at which the swarm perceives its surrounding geometry. $\sigma$ is a distance, and ideally it should match the typical size of geometric features (for instance the radius of curvature of the area if it is a disk). Yet there is a trade-off between the extent of the field of perception and the number of robots \textit{N}. If the \textit{N} and $\sigma$ are too small, the coverage of the area by the robots is sparse and robots cannot find their neighbors (regime \textbf{r5} in Fig.~\ref{fig:phase_diagrams}a). The connection graph is disconnected and composed of islands of independent clusters that cannot communicate to reach a consensus on the geometry. On the other hand if \textit{N} and $\sigma$ are too large, the neighborhood of each robot is saturated with other robots and the adjacency matrix locally resembles that of a clique (full of 1, indicating that neighboring robots are all connected to each other). This configuration carries little information about local geometric details (regime \textbf{r2} in Fig.~\ref{fig:phase_diagrams}a). 

We can analyze that behavior quantitatively.
The local area covered by a robot through its field of perception is $\pi\sigma^2$, and the total area covered by a swarm of $N$ robots (possibly with overlaps) is therefore $N\pi\sigma^2$. We define the coverage ratio as $\alpha(N,\sigma)=N\pi\sigma^2/S$. Small values of $\alpha$ indicate that large portions of the arena are not covered by the swarm, whereas large values indicate that each patch is, on average, covered by multiple robots. We therefore hypothesized that the optimal configurations would correspond to a constant $\alpha$: in other words, the optimal $N$ and $\sigma$ would satisfy $N=\alpha S/(\pi\sigma^2)$, where $\alpha$ would be independent of $N$, $\sigma$, and arena shape. 

We performed simulations for 7 shapes and for a number of robots \textit{N} varying from ~50 to ~500, and $\sigma$ ranging from 30 to 320 mm (which are typical values for a swarm). We found that the optimal 
pairs of $(N,\sigma)$ clustered along the hyperbola $\alpha=17$ (\textbf{r1}, \textbf{r3}, \textbf{r4}) shown in red in Fig.~\ref{fig:phase_diagrams}a: \textbf{r1} has a relatively high number of robots, with low $\sigma$, \textbf{r4} has a relatively low number of robots but with high $\sigma$, and \textbf{r3} is a middle-ground between \textbf{r1} and \textbf{r4}. Those results supported the coverage-ratio hypothesis. As suggested by~\cite{talamali2021less}, the highest values of the field of perception were not necessarily correlated with higher accuracy.

The lower boundary of the optimal conditions region is the hyperbola $\alpha=7.5$ (white in Fig.~\ref{fig:phase_diagrams}a) that marks the transition between graphs with several connected components (below the curve) to only one component (above the curve, normal behavior of the algorithm). The mean number of connected graph components is presented in Supplementary Fig. S3 (bottom). 
Below the curve, robots can still propagate local information from components to components when they move at the beginning of each iteration and change their neighbors. The swarm will converge to inexact $\lambda_2$ estimates but that may still be different depending on the shape of the arena, as seen in regime \textbf{r5} where the algorithm can distinguish a disk from an annulus, but is not accurate enough to identify 7 different shapes.
Results directly on the curve correspond to a ``low-accuracy valley" at the bifurcation between the two dynamics.

The upper boundary of the optimal conditions region is the hyperbola $\alpha=32$ (black in Fig.~\ref{fig:phase_diagrams}a) that separates convergent results (below the curve) from overconnected configurations leading to algorithmic instabilities (regime \textbf{r2}). The former are cases where the diffusion process behaves as expected, with an exponential decay of $|s_i^n|$ on all robots. The latter cases had overconnected graphs that rendered $\lambda_2$ estimation more difficult for two reasons: 1) all arenas will correspond to relatively similar communication graph topologies and 2) too many neighbors broadcast information, resulting in numerical instabilities (as we rely on the explicit Euler method for ODE resolution to simplify implementation on real robots). Further analyses demonstrate that the degree distribution of the communication networks across all arenas affects the classification performance of the algorithm: having more than 25 mean neighbors does not allow the algorithm to converge (Supplementary Fig.~S3 Top). Best-performing configurations tend to have a skewness of the degree distribution that varies depending on the arena (Supplementary Fig.~S4) with a lower mean number of neighbors close to a wall.

We estimated how many runs were necessary for a swarm to discriminate the shapes (Fig.~\ref{fig:phase_diagrams}b). For regimes \textbf{r1} and \textbf{r3}, most shapes are readily separated after a few iterations (\textit{e.g.,} the square and the disk are easily separated from the rest). Following the central limit theorem, the standard deviation of $\lambda_2$ estimates across runs decreased as $\xi^{-k}$ with $\xi$ the number of iterations, and $0 < k \leq 1$ a variable that depends on the case. It takes about 20 iterations to separate shapes with similar $\lambda_2$, such as the square and the disk. The confusion matrices in Fig.~\ref{fig:phase_diagrams}c confirm the quality of classification: \textbf{r1} and \textbf{r3} have respectively $100\%$ (perfect score) and $86\%$ of accurately classified shapes. Interestingly, the magnitude of $\lambda_2$ is roughly in line with the expected connectivity of the shape. A convex shape like a disk diffuses information approximately ten times more quickly than a non-convex shape like the annulus. 

The way the robots move during their random walk phase influences the diffusion dynamics (Supplementary Sec.~3.2). If the robots are evenly distributed within their arena and only form one connected graph component, they will be able to capture all local features of the shape and communicate it to the entire swarm -- in this case, staying immobile and avoiding the random walk phase would not have a detrimental effect on the accuracy of the algorithm. Inversely, if the robots are aggregated only in some parts of the arena, and form several disconnected graph components (\textit{e.g.,} regime \textbf{r5}), it is necessary for them to disperse at each iteration to change their neighbors and broadcast their states from graph component to graph component.

To assess whether $\lambda_2$ provided information beyond simpler distributed statistics, we compared the spectral descriptor with several baselines in simulation: mean degree, statistics of the degree distribution, and basic consensus convergence time (Supplementary Sec. 3.5). These baselines captured local density, graph heterogeneity, or generic mixing speed, but they did not reproduce the same robust shape-dependent separation across regimes as the $\lambda_2$ estimate.

\begin{figure*}[p!]
\begin{center}
\includegraphics[width=1.00\textwidth]{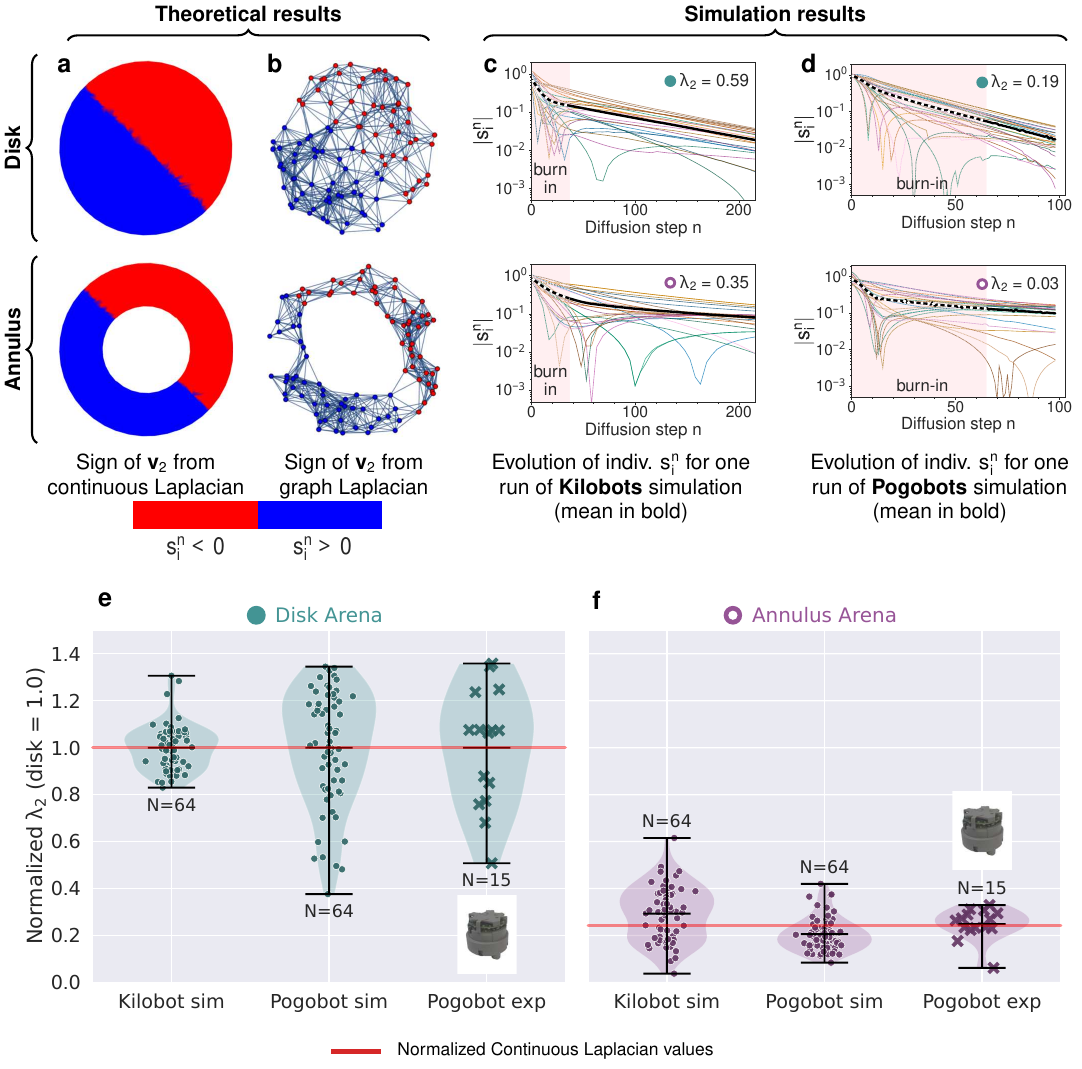}%
\caption{\textbf{Comparison between theoretical, simulated, and experimental results on a two-class classification setting with two shapes (disk and annulus) using one iteration of the algorithm.} {\small
\textbf{(A) and (B)} Examples of arena partition in the continuous and discrete (graph) cases. Colored zones and nodes correspond to the sign of the local component of the Fiedler vector $\textbf{v}_2$ of the Laplacian (red, negative; blue, positive). In both cases, global information (access to the full Laplacian matrix) is used to compute $\textbf{v}_2$.
\textbf{(C) and (D)} Evolution of the internal state $|s_i^n|$ of each robot $i$ (Kilobots in C, Pogobots in D) during a diffusion session in simulations, using only local information. Thin curves show individual robot trajectories; the bold black curve shows their mean in that run. The slope of each individual curve is used to estimate a local value of $\lambda_2$.
\textbf{(E) and (F)} Distributions of estimated $\lambda_2$ values for disk (E) and annulus (F) arenas in Kilobot simulations, Pogobot simulations, and Pogobot experiments. Symbols represent run- or experiment-level mean estimates. Horizontal central marks indicate the mean, and vertical bars span the minimum and maximum. Each condition comprised $N=64$ simulation runs or $n=15$ experimental runs per arena, with 25 robots per run; each point is the mean estimate across the 25 robots in one run. Values were normalized such that the mean of the corresponding disk condition is 1.0; red horizontal lines indicate the normalized continuous-Laplacian reference values. Disk and annulus distributions were compared separately for Kilobot simulations, Pogobot simulations, and Pogobot experiments using Mann--Whitney--Wilcoxon tests with Bonferroni correction for multiple comparisons (adjusted $P<10^{-4}$ for each comparison).
}}
\label{fig:statsSimuVsExpes}
\end{center}
\end{figure*}

\begin{figure*}[p!]
\begin{center}
\includegraphics[height=0.82\textheight]{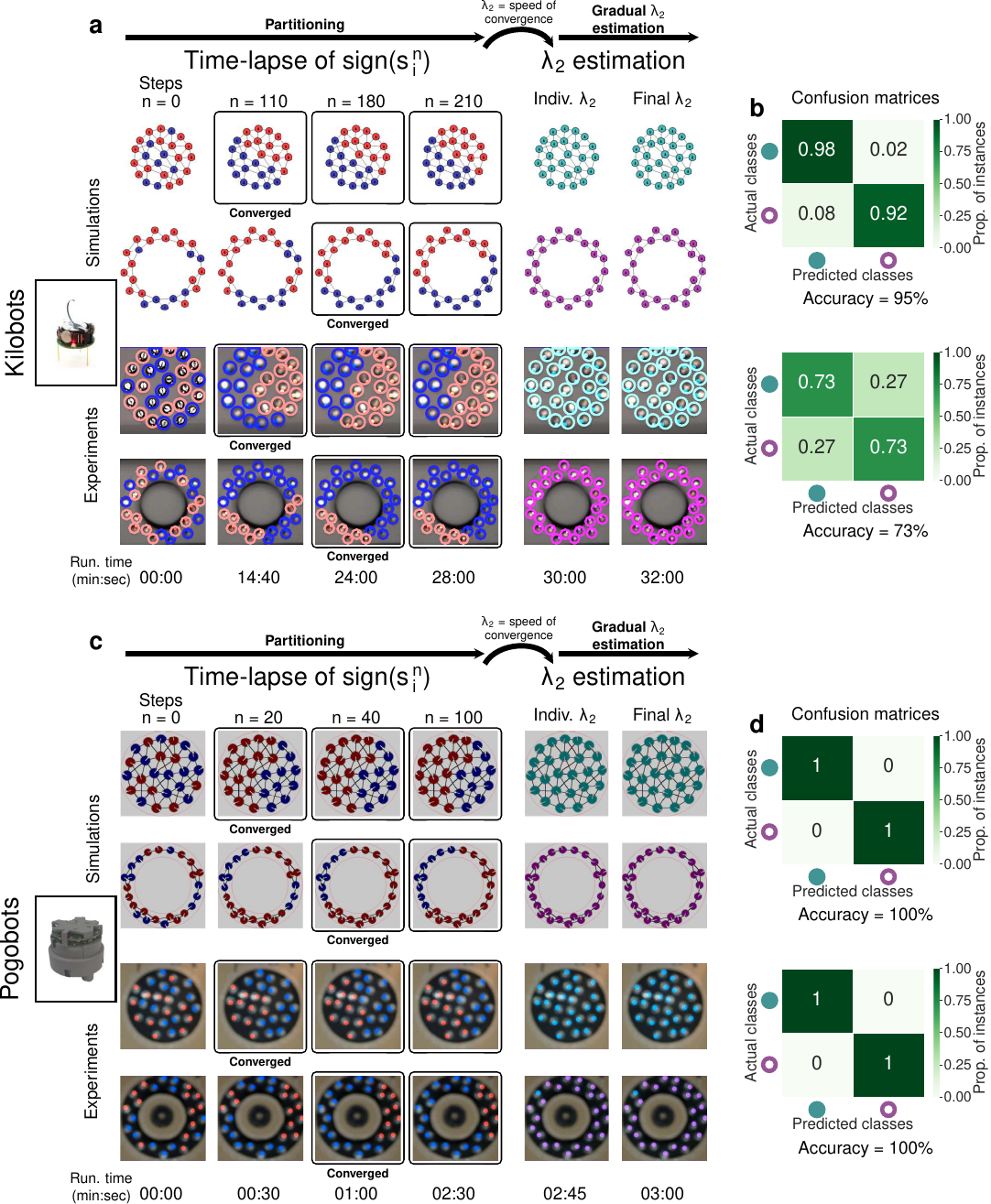}%
\caption{\textbf{Comparison between simulated and experimental results on a two classes classification setting with two shapes (disk and annulus) using one iteration of the algorithm.} \small{
\textbf{a \& c:}Time-lapses of the algorithm, in simulations and in experiments, with 25 robots (Kilobots in \textbf{a}, Pogobots in \textbf{c}). The times exclude the prediffusion stage for simplicity, so actual durations are nearly double. Experimental photos are blurred to ease the visualization of LED colors. Left: diffusion, resulting in partitioning (the color of each robot corresponds to a local component of the Fiedler vector $\textbf{v}_2$). Right: individual $\lambda_2$ estimation on each robot from the convergence rate of diffusion, and consensus (collective averaging) over the entire swarm. Robots show a LED color-code according to the value of $\lambda_2$: cyan when a disk is detected, violet for an annulus. Images are re-scaled for better visualization; both simulations and experiments have arenas surface approximately equal to $70000\texttt{mm}^2$ for Kilobots, and $150000\texttt{mm}^2$ for Pogobots (both setups have roughly the same density). 
\textbf{b \& d:}~Confusion matrices and accuracy scores of simulations (Kilobots in \textbf{b}, Pogobots in \textbf{d}) over 64 runs per arena, and experiments (over 15 runs per arena).
}}
\label{fig:simuVsExpes}
\end{center}
\end{figure*}

\begin{figure*}[p!]
\begin{center}
\includegraphics[width=1.00\textwidth]{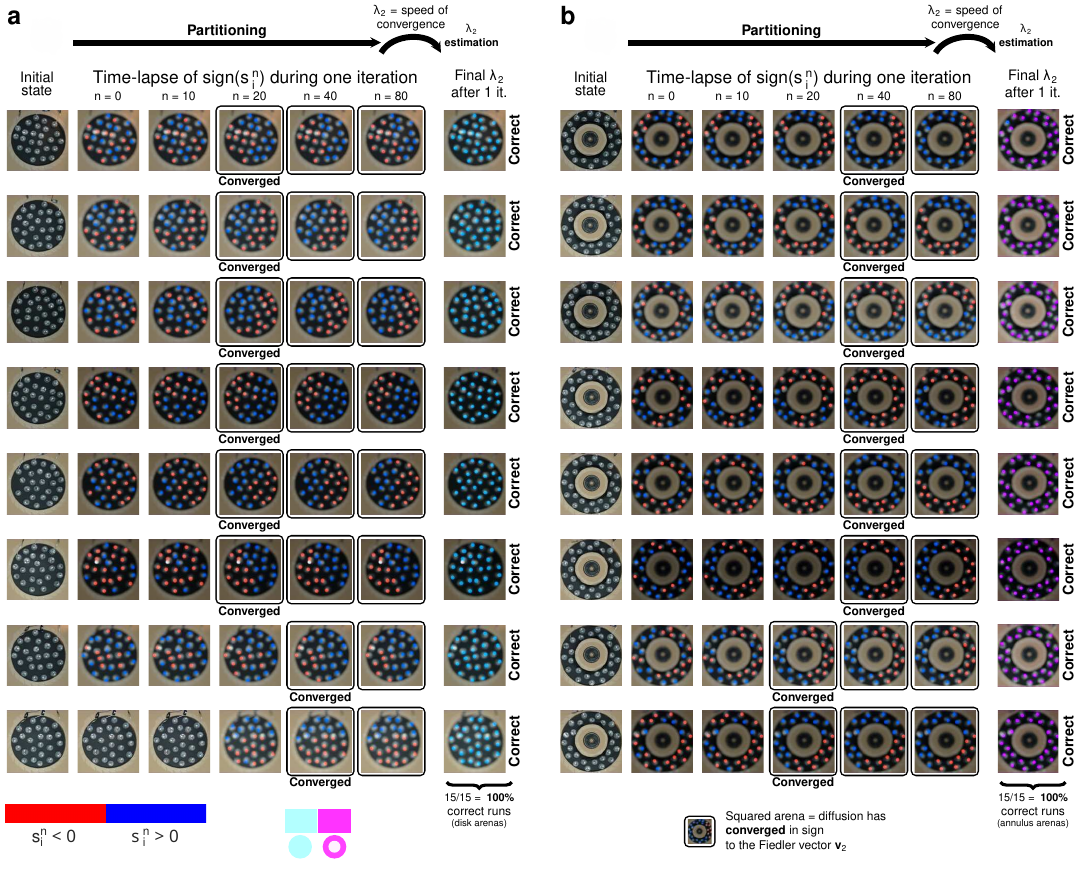}%
\caption{\textbf{Time-lapses of the experiments with Pogobots}, with 25 Pogobot robots, respectively for the disk (\textbf{a}) and annulus (\textbf{b}) arenas. \small{ Experiments use the same parameters as simulations and experiments of Fig.~\ref{fig:simuVsExpes}. Only the results of 16 runs are shown (over 30 considered runs). Time-lapses of all runs are presented in Supplementary movie 1. Experimental photos are blurred to ease the visualization of LED colors (except left column). Images are re-scaled for better visualization; experiments have arenas surface approximately equal to $150000\texttt{mm}^2$.
\textbf{Left}:~Initial state of the robots before the start of experiments.
\textbf{Middle}:~diffusion, resulting in partitioning (the color of each robot corresponds to a local component of the Fiedler vector $\textbf{v}_2$). In the studied configuration, robots typically converge in the domain $n\in [10, 20]$ for the disk arenas and in the domain $n\in [40, 80]$ for the annulus arenas. In rare cases, the robots may still change state after convergence, due to computation errors, or message transmission errors -- in such case, the partitioning the robots converged into may be broken (\textit{e.g.,} last experiment on the annulus arena). The spectral swarm robotics algorithm will detect and ignore divergent cases after convergence (cf details in Supplementary Sec.~2).
\textbf{Right}:~individual $\lambda_2$ estimation on each robot from the convergence rate of diffusion, and consensus (collective averaging) over the entire swarm. Robots show a LED color-code according to the value of $\lambda_2$: cyan when a disk is detected, violet for an annulus.
The robots detected the correct shape on 30 runs over 30 runs (15 runs on each arena), which translates into an accuracy score of $30/30 = 100\%$. }}
\label{fig:expes_pogobots}
\end{center}
\end{figure*}

\subsection*{Experimental validation with swarms of robots}

We validated our approach on two robotic platforms, with different properties (Fig.~\ref{fig:overview}, Supplementary Fig. S1): \textbf{Kilobots} and \textbf{Pogobots}.

Over the past ten years, Kilobots~\cite{rubenstein2012kilobot,rubenstein2014programmable} have become a standard to prototype robot swarms. They have been used for numerous tasks including self-assembling into predetermined shapes~\cite{rubenstein2014programmable,slavkov2018morphogenesis}, phototaxis~\cite{divband2019adaptive}, creating moving and deforming soft-bodied kilobots aggregates~\cite{pratissoli2023coherent}, morphological computation and decentralized learning~\cite{ben2023morphological}, contour detection~\cite{raoufi2023estimation}.

On its own, a Kilobot is a centimeter-sized robot that (rudimentarily) moves and (imperfectly) communicates with its neighbors up to $85$mm and at very low speeds: up to 2 messages of 9 bytes per second, but only $\approx 5 \texttt{bytes}/\texttt{s}$ when taking into account message collisions. They lack sensors for position or velocity and haphazardly navigate their environments, similarly to a random walk. Kilobots are also ``blind" to their environment. They only sense ambient light and lack the panoply of proximity sensors (camera, LIDAR, radar, ultrasound, capacitive or force sensors) embarked on larger robots required to detect obstacles, follow walls, or orient themselves.

Pogobots~\cite{loi2025pogobot} are a recent robotic framework designed by the ISIR lab at Sorbonne Universit\'{e}. They feature vibration-based locomotion, infrared communication, and an array of sensors in a cost-effective package. Pogobots offer an accessible alternative to Kilobot platforms while providing advanced capabilities including high-throughput directional communication between units: they can send up to 120 messages of 20-100 bytes per second up to $135$mm (center to center), and are $\approx 30-500$ times faster than Kilobots. In turn, Pogobots lack the ability to estimate communication distances between neighbors, a feature implemented by the Kilobots -- it is thus not possible to control the field of perception $\sigma$ with Pogobots.

We implemented the SSR algorithm in Kilobot and Pogobot simulations using, respectively, two simulators: Kilombo~\cite{jansson2015kilombo} and Pogosim~\cite{cazenille2025pogosim}. They both allowed the same respective C code to be used both in simulations and as controllers of actual robots, without any modification. Both simulators modeled collisions through a physics engine and reproduced communication loss and sensor and motor biases.

We successfully validated experimentally our algorithm on two swarms of 25 Kilobot and Pogobot robots (Fig.~\ref{fig:statsSimuVsExpes},~\ref{fig:simuVsExpes} and Fig.~\ref{fig:expes_pogobots}). The experimental setup is described in Supplementary Sec.~3, and videos listed are in Supplementary Sec.~3.6.
We focused on the main experimental figures on two arenas: the disk and the annulus. These shapes were selected because they correspond to well-separated $\lambda_2$ values (Fig.~\ref{fig:phase_diagrams}b), making them suitable for validating the full distributed spectral-estimation pipeline on physical robots under realistic communication and hardware constraints. To test classification beyond this binary contrast, we also performed additional experiments with a third arena shape (triangle), reported in Supplementary Sec. 3.4.
To make the experimental validation more challenging, and showcase the robustness of our algorithm, we selected particularly harsh conditions in terms of number of robots and number of iterations. Simulation results (Supplementary Fig.~S11) showed that on two-arena classification tasks only one iteration of the algorithm was sufficient, even for a relatively small number of robots.
In the experimental setup, robots started well distributed in the arena to study the performance of the diffusion process (Supplementary Sec.~2.1). 

The cases that used only local information (simulations and experiments in Fig.~\ref{fig:simuVsExpes}a,c and Fig.~\ref{fig:statsSimuVsExpes}c,d,e,f) showed behavior similar to the theoretical results computed using global information (with access to the full Laplacian matrix, either in the continuous or discrete case: Fig.~\ref{fig:statsSimuVsExpes}a,b). Notably, the diffusion processes in all four cases resulted in a graph partition into two zones, suggesting that the swarm had correctly converged along the second eigenvector.
Figures~\ref{fig:statsSimuVsExpes}c,d show instances of time-evolution of the internal state of 25 robots during a diffusion process in simulations, respectively for Kilobots and Pogobots: both cases showed an exponential decay whose slope (convergence rate) approximated $\lambda_2$. In both simulations and experiments, the diffusion processes tended to converge faster in disk arenas than in annulus arenas (Figs.~\ref{fig:statsSimuVsExpes}c,d and ~\ref{fig:simuVsExpes}a,c and ~\ref{fig:expes_pogobots}a,b), as expected from theory.

We considered the diffusion process to have converged when two distinct nodal zones emerged. In most cases, no subsequent evolution occurred. However, diffusion could occasionally diverge (the partition no longer consisted of only two zones) after reaching convergence because computation errors propagated when the local states were close to zero. After diffusion converged, the algorithm ignored any possible subsequent divergence in its estimation of $\lambda_2$.
Communication errors and local asymmetries in the Laplacian matrix (one-way communication between a robot and its neighbor) could lead to uneven partitions (Fig.~\ref{fig:expes_pogobots}), despite algorithmic methods to reduce communication errors (Supplementary Secs.~2.3 and~3.3).

Our approach was robust to these issues for shape classification purposes (Fig.~\ref{fig:statsSimuVsExpes}e,f), both in experiments (accuracies of Kilobot: 73\%, Pogobots: 100\%) and simulations (accuracies of Kilobots: 95\%, Pogobots: 100\%). Moreover, the mean normalized $\lambda_2$ values obtained in simulations and experiments matched the theoretical results based on global information, \textit{i.e.}, by computing $\lambda_2$ directly from the entire Laplacian of the continuous shape (Fig.~\ref{fig:statsSimuVsExpes}a) or on an embedded graph (Fig.~\ref{fig:statsSimuVsExpes}b).

The experimental results with Kilobots showed that the algorithm could operate on robots with very slow communication channels and low computing capabilities. Pogobots were able to solve the classification task much faster: $\approx$ 5 min per iteration including pre-diffusion, instead of $\approx$ 79 minutes per iteration for Kilobots~\footnote{Note that the times in Fig.~\ref{fig:simuVsExpes} do not include the pre-diffusion part for simplicity, such that actual times are nearly double.}, \textit{i.e.,} a 15 times speed-up. Because Pogobots were unable to estimate neighbor distances, their field of perception $\sigma$ was fixed, as opposed to Kilobots that could change it by a cutoff threshold. However, Pogobots were still able to fully solve the classification problem with higher experimental accuracy than Kilobots. This suggested that communication bandwidth and speed were the dominant limiting factors in experimental settings.

Finally, we assessed the impact of 1) communication noise, \textit{i.e.,} loss of messages and 2) temporal noise, \textit{i.e.,} drift of internal clocks on the algorithm.
They were not idealized away in the physical implementation but were part of the reality gap addressed by the Kilobot and Pogobot experiments. The robustness analyses in Supplementary Sec. 3.3 quantified how the algorithm behaved under these imperfections.
In the first case, a single parameter needed to be adjusted to be fully robust to message loss: the amount of time for which neighboring messages were retained by each focal robot. Incorrect settings could result in large accuracy losses (for example, from 100\% to 31\%). In the second case, the algorithm could synchronize internal clocks within a neighborhood by including the current time in exchanged messages. However, if neighboring robots' clocks differed by several times $\tau$, the algorithm could exhibit strong numerical instabilities, similar to regime \textbf{r2} in Fig.~\ref{fig:phase_diagrams}, because robots could skip several diffusion time steps while synchronizing their clocks.

\section*{Discussion}

We showed that a swarm of simple robots can ``hear'' the shape of its environment by collectively estimating a Laplacian eigenvalue from purely local interactions. The robots repeatedly exchange scalar states with their neighbors to emulate diffusion, then infer a global spectral quantity from the decay rate of this process. This mechanism requires no localization, mapping, or global coordination. It is validated both numerically at large scale and experimentally on two different robotic platforms, Kilobots and Pogobots, which impose severe constraints on communication bandwidth, sensing, and computation.

A central result of our study is that the spectral estimate is robust and scalable. Unlike gradient-based methods, where a single faulty update can trigger cascades of errors that propagate through the swarm~\cite{gauci2017error,wang2020fast}, diffusion naturally regularizes transient local errors: perturbations are progressively smoothed as information spreads~\cite{slavkov2018morphogenesis}, and catastrophic failures do not occur except in deliberately overconnected configurations. 
Moreover, robots move between successive diffusion runs, enabling iterative refinement of the classification results. This motion introduces small shape-dependent perturbations in the communication graph across runs that reduce the risk of confusion between non-isometric but isospectral arenas, a known theoretical limitation of Laplacian-based shape analysis~\cite{gordon1992one}.

The present implementation should therefore be interpreted as a first demonstration of spectral swarm robotics rather than a complete solution to arbitrary shape recognition. It estimates one descriptor, $\lambda_2$, which is primarily sensitive to global connectivity and bottlenecks. Richer spectral signatures, involving several eigenvalues and possibly eigenvectors, will be required to approach more general shape recognition or reconstruction.

The classification step used here is therefore library-based. The distributed part of the algorithm is the estimation of the spectral descriptor; the mapping from this descriptor to a semantic label is task-dependent and requires predefined candidate classes. Extending the method toward more open-ended shape recognition would require estimating richer spectral signatures and developing unsupervised or adaptive interpretation layers.

A first extension of this work would be to further exploit spectral inference. Here, we estimate the second eigenvalue to provide a robust fingerprint of the swarm’s shape, but extracting higher modes would provide richer geometric information and could improve discrimination between shapes (families of arenas in 2D or even 3D environments). Moving from a single scalar to a higher-dimensional spectral signature would nevertheless require more accurate numerical schemes and careful embedded implementation, which is challenging given the hardware constraints of onboard execution. 

A second extension is to couple spectral inference with motility, moving from static interaction graphs to fully dynamic ones. Allowing robots to move continuously during spectral estimation could rely on time-averaged or temporally filtered Laplacians~\cite{holme2012temporal,masuda2013temporal}, asynchronous consensus and gossip algorithms~\cite{boyd2006randomized,kempe2003gossip}, or signal-processing tools for dynamic graphs such as graph Fourier methods~\cite{shuman2013emerging}. Oscillator-based strategies, inspired by natural synchronization phenomena, offer another way to encode spectral information in the phase relations of moving agents. In this regime, the spectrum would no longer describe only the geometry of a static swarm, but the joint structure of geometry and motion. Naturally, this raises the important question of how to ensure synchronization across multiple timescales: the timescale of information diffusion and that of physical diffusion.

Finally, the same spectral approach could be applied to other tasks beyond arena recognition, such as enabling self-counting~\cite{wang2020fast}, boundary and interface detection~\cite{mclurkin2009distributed}, identification of internal cavities~\cite{hidalgo2024frugal}, or monitoring density and stress patterns within the swarm~\cite{boudet2021collections}. Coupled with locomotion strategies, it may become possible to perform tasks such as morphing between target shapes or redistributing density in response to external stimuli. In all these cases, spectral indices serve as global order parameters guiding spatial self-organization of the swarm.

We believe that our main contribution is to position spectral swarm robotics not merely as a single algorithm, but as a broader research direction in which swarms exploit the spectral properties of their interaction graphs as a substrate for perception and control. By linking diffusion, topology, and geometry to practical robotic behaviors, our framework offers a spectral perspective on embodied collective intelligence: swarms that compute using their own connectivity pattern, extracting global invariants from local exchanges, and using these invariants to adapt their spatial structure and function to their environments.

\bibliographystyle{unsrt}
\bibliography{biblio}

\section*{Acknowledgments}
We thank Yoones Mirhosseini and Jérémy Fersula for their advice on using Kilobots. We thank Laurent Fabre for his help with the Pogobot experimental setup.

\textbf{Funding:} This work was supported by the MSR and SSR projects funded by the Agence Nationale pour la Recherche under Grant No ANR-18-CE33-0006 and ANR-24-CE33-7791 and JSPS Fostering Joint International Research (B) Grant No JP19KK0261. NLD acknowledges support from European Research Council (ERC) under the European's Union Horizon 2020 program (grant No 770940).

\textbf{Authors contributions:} A.J.G., L.C., and N.B. conceptualized, planned, and supervised the project. A.J.G., N.L-D., N.A-K., O.M. provided the theoretical background and design, L.C. and A.J.G. designed the algorithm. L.C. wrote most of the code for robotic controllers and simulation, and most of analysis and plotting scripts. L.M. and A.L. made some code additions. A.J.G. and N.L-D. performed the theoretical analysis. L.C. computed and analyzed all simulations. L.C., N.B, A.L., and M.I. conducted all experiments with robots, with additional help from L.M. for Pogobot experiments. N.B. and S-H.K. provided experimental resources. L.C., N.L-D., and N.B. assembled the figures. L.C., A.J.G., N.B., N.L-D., and N.A-K. wrote the manuscript. L.C., A.G., N.A-K., and N.B. wrote the supplementary information and designed the supplementary videos. N.B., N.A-K., A.J.G., L.C., and S-H.K. acquired the funding.

\textbf{Competing interests} The authors declare no competing interests.

\textbf{Data, code, and materials availability:} Data supporting the conclusions are included in the main text and Supplementary Materials. Tabulated numerical data underlying the figures, including statistical tests and $P$ values, are provided in data file S1 accompanying the published article (\url{https://doi.org/10.1126/scirobotics.adr9578}). Source code is archived on Figshare at \url{https://doi.org/10.6084/m9.figshare.33963613} and is also available on GitHub at \url{https://github.com/leo-cazenille/spectral-swarm-robotics}.

%%%%%%%%%%%%%%%%%%%%%%%%%%%%%%%%%%%%%%%%%%%%%%%%%%%%%%%%%%%%%%
%%%%%%%%%%%%%%%%%%%%%%%%%%%%%%%%%%%%%%%%%%%%%%%%%%%%%%%%%%%%%%
%%%%%%%%%%%%%%%%%%%%%%%%%%%%%%%%%%%%%%%%%%%%%%%%%%%%%%%%%%%%%%
%%%%%%%%%%%%%%%%%%%%%%%%%%%%%%%%%%%%%%%%%%%%%%%%%%%%%%%%%%%%%%
%%%%%%%%%%%%%%%%%%%%%%%%%%%%% SI %%%%%%%%%%%%%%%%%%%%%%%%%%%%%

%\clearpage
\onecolumn

%\title{Hearing the shape of an arena with spectral swarm robotics\\Supplementary Materials}

\makesupplementarytitle

\setcounter{section}{0}

% Redefine the numbering for supplementary figures
\renewcommand{\thefigure}{S\arabic{figure}}
\setcounter{figure}{0}
% Redefine the numbering for supplementary tables
\renewcommand{\thetable}{S\arabic{table}}
\setcounter{table}{0}
% Redefine the numbering for equations
\renewcommand{\theequation}{S\arabic{equation}}
\setcounter{equation}{0}

\section{Theoretical background}

\subsection{Spectral shape analysis}
Mark Kac's famous 1966 paper, ``Can One Hear the Shape of a Drum?''~\cite{kac1966can} presented a novel and fascinating question rooted in the field of mathematical physics: if you only heard the sound a drum makes, could you figure out what shape the drum is?
In other terms, is it possible to identify the shape of a drumhead merely from the list of its vibrational frequencies (or eigenvalues), essentially 'hearing' its shape?
This concept is an important question in the realm of spectral geometry. It challenges us to consider whether different drumheads (mathematically modeled as bounded planar domains) could produce identical sounds, thereby sharing the same eigenvalue spectrum, or if the sound of a drum is as unique as its shape. While it has been established that instances of differing shapes producing identical spectra are possible, such cases are remarkably rare, and tend to be anomalies with non-smooth, non-convex boundaries~\cite{gordon1992one}. Such cases arise only through carefully engineered constructions and are considered non-generic; randomly chosen domains almost surely do not share identical spectra. This emphasizes the intricate interplay between the auditory perception of an object and its geometric structure.

In network theory, related questions have been investigated in the context of temporal networks~\cite{holme2012temporal}. These networks describe whether and when interactions occur between node pairs, emphasizing the timing and flow of information rather than its semantic content.
Related ideas have also been developed in graph signal processing, which studies the harmonics of a graph through so-called Graph Fourier Transforms~\cite{shuman2013emerging}.
A similar exploration was made in spectral graph theory~\cite{spielman2012spectral,chung1997spectral}, but here the aim is to pull out global graph traits under the assumption of a static graph and a central observer overseeing all nodes. Specifically, methods of \textbf{spectral shape analysis}~\cite{kac1966can} extract global details about a shape (\ie the macroscopic organization of a swarm) by examining the local diffusion of information within the graph of linked nodes (\ie the graph of communication channels between robots).

At the core of these spectral approaches lies the Laplacian operator.
Its eigenvalues characterize the natural diffusion modes of the system and thus provide a compact spectral signature of the network’s global structure.
More generally, diffusion processes, whether in continuous space or on discrete graphs, are governed by a Laplacian operator $\Delta$~\cite{evans2022partial} - a second-order linear operator that corresponds to the divergence ($\nabla \cdot$) of the gradient ($\nabla$) and quantifies the local curvature of a function $\Delta f = \nabla \cdot (\nabla f)$ by comparing its value to the average of its neighborhood.
The Laplacian is ubiquitous in physics~\cite{arendt2009weyl} and describes the diffusion of heat, matter, or momentum (through the heat equation) or the propagation of mechanical or electromagnetic waves (through the wave equation). The Laplacian also controls the propagation of the wavefunction in a quantum system  and its energy levels (through the Schr\"{o}dinger equation). The Laplacian is an intrinsic differential operator: it is based on the internal geometry of a space and does not need to refer to any external coordinate system (as opposed to the gradient for instance).
This property makes it invariant by rotation and translation - which is essential for formulating physical laws that are independent of the frame of reference.

For more complex spaces, the Laplace–Beltrami operator (LBO) is a generalization of the Laplace operator to functions defined on submanifolds in Euclidean space and more generally on Riemannian manifolds.
The recent surge of interest in the spectral analysis of the LBO has resulted in a considerable number of spectral shape signatures extraction methods that have been successfully applied to a broad range of areas, including manifold learning~\cite{belkin2006manifold}, object recognition and deformable shape analysis~\cite{rustamov2007laplace,levy2006laplace}, medical imaging~\cite{chaudhari2014global}, shape classification~\cite{reuter2006laplace} and distributed sensor networks (e.g., 5G antenna networks).
The LBO is ideal for shape analysis for several reasons:
(1) it is intrinsic (invariant to isometric deformations~\cite{zhang2010spectral});
(2) it is not tied to a specific shape representation and can be discretized on 2D manifolds, meshes, point clouds, and so on~\cite{melzi2018localized};
(3) its eigenbasis is optimal for approximating smooth functions on the manifold, and a small number of low-frequency eigenfunctions capture the dominant large-scale geometric structure~\cite{aflalo2015optimality}.
This property underlies spectral shape descriptors such as Shape-DNA~\cite{reuter2006laplace}, which characterize a shape by computing its first Laplace–Beltrami eigenvalues.

However, a critical limitation for swarm robotics is that, while diffusion can be enacted at a microscopic scale, graph traits are derived by centrally combining information. This implies that individual nodes do \textit{not} have access to macroscopic data. By using linear and nonlinear operators for a distributed application of diffusion, we propose that the diffusion of information can serve as a bridge between the local scale of an individual robotic agent and the global scale of the swarm (Main Text Fig.~1).

To access global structural information at the level of individual robots, diffusion processes can be instantiated over the communication graph. Although the Laplacian is a local operator—defined through interactions between neighboring nodes—its eigenvalues are determined by the global geometry and boundary conditions of the environment. Consequently, spectral quantities such as the second eigenvalue $\lambda_2$ (the algebraic connectivity) can serve as fingerprints of the swarm’s spatial configuration. For example, the shapes shown in Main Text Fig.~2a yield distinct values of $\lambda_2$, making it a discriminative global descriptor.

Spectral analysis is typically only computed using global information, by having the entire Laplacian matrix available~\cite{melzi2018localized,choukroun2018hamiltonian}. Robots only have access to local information (their own internal state and information communicated by their neighbors). As such, computing global statistics on individual robots is challenging. Due to very slow communications (a few bytes per second) and very limited memory capabilities, it may be impossible, or at least extremely difficult, to regroup the entire Laplacian matrix on all robots of the swarm.
Instead, we propose to estimate spectral statistics in a distributed way, and embodied in a swarm of robots, allowing the swarm to collectively ``hear'' the shape of their arena.

A recent study~\cite{hidalgo2024frugal} demonstrated shape recognition using probabilistic geometry methods on a single robot (but without spectral analysis). However, that approach relies on centralized information and does not readily extend to a distributed swarm setting, where communication bandwidth and memory are severely constrained. In contrast, our method is inherently distributed and allows the swarm to collectively infer global structure without aggregating the full Laplacian matrix on any single agent.

\subsection{Nomenclature}
\begin{table*}[h]
\begin{center}
\resizebox{1.00\textwidth}{!}{%
\renewcommand{\arraystretch}{1.2}
\begin{tabular}{|l |l |p{15cm} |}
\hline
Variable name & Unit & Description \\
\hline
$N$ & - & Number of agents. \\
$H$ & - & Mean number of neighbors of agents in a given configuration. \\
$\sigma$ & $[m]$ & Radius of the field of perception of the agents. \\
$S$ & $[m^2]$ & Arena surface. \\
$\mathcal{G}$ & - & Graph of communication flows between agents \\
$\textbf{A}$ & - & Adjacency matrix of $\mathcal{G}$. $A_{i,j}=1$ if the agents $i$ and $j$ are neighbor, and 0 otherwise (by convention a robot is not its own neighbor). \\
$\textbf{L}$ & - & Laplacian matrix of $\mathcal{G}$. On the diagonal, $L_{i,i}$ is the number of neighbors detected by the agent $i$ (max. value: $H$), and off the diagonal, $L_{i,j}=-1$ if i and j are neighbors, and $L_{i,j}=0$ otherwise. \\
$s_i^{n}$ & -& Internal state of agent $i$ at diffusion step $n$ (\ie time $t = n \times \tau$). Used during the diffusion stages of the algorithm. Intended to have a zero mean at each step $n$. \\
$\kappa$ & $[m^2.s^{-1}]$ & Coefficient of information diffusion.\\
$c = 1 s^{-1}$ & $[s^{-1}]$ & Information diffusion rate across edges of $\mathcal{G}$. Note that c includes a notion of distances between nodes; here we assume that all edges have the same weight (distance is 1, i.e., $\mathcal{G}$ is unweighted).\\
$\tau$ & $[s]$ & Amount of time between two steps of diffusion. \\
$T$ & - & Number of diffusion steps. \\
$B$ & - & Number of diffusion steps to ignore at the beginning of a diffusion session when estimating $\lambda_{2}$. \\
$C$ & - & Number of collective averaging rounds. \\
$\lambda_i$ & - & i-th eigenvalue of the Laplacian $\textbf{L}$. \\
$\texttt{Indiv}{\lambda_i}_j$ & - & Local estimation of the i-th eigenvalue of the Laplacian $\textbf{L}$ by agent j during one iteration. \\
$\texttt{Consensus}{\lambda_i}_j$ & - & Consensus estimation of the i-th eigenvalue of the Laplacian $\textbf{L}$ by agent j during one iteration. \\
$\texttt{Final}{\lambda_i}_j$ & - & Final estimation of the i-th eigenvalue of the Laplacian $\textbf{L}$ by agent j. \\
$\textbf{v}_i$ & - & i-th eigenvector of the Laplacian $\textbf{L}$. \\
$l = \sqrt{S/N}$ & $[m]$ & Typical lengthscale of the distance between agents. \\
$P$ & - & Number of information diffusion sessions computed in parallel. \\
$I$ & - & Number of iterations. \\
\hline
\end{tabular}
}
\caption{Nomenclature of all considered variables}
\label{tab:nomenclature}
\end{center}
\end{table*}

\subsection{Information diffusion model}
The diffusion model refers to a mathematical representation of the physical process of diffusion, in which a large number of small elements spread in the environment through random movement and collisions. For the two-dimensional case considered in this article, assuming the number of elements is large enough, one can define their distribution at a given spatial position $(x,y)$ as a local concentration $s(x,y)$, with changes over time according to the following partial differential equation:

\begin{equation}\label{eq:diffCont}
    \frac{\partial s}{\partial t} = \kappa \nabla^2 s
\end{equation}
where $\kappa$ is the coefficient of diffusion, a macroscopic value specific to the elements and $\nabla^2$ is a 2D Laplacian operator defined as:
\begin{equation}\label{eq:2dlaplacian}
    \nabla^2 = \frac{\partial^2}{\partial x^2} + \frac{\partial^2}{\partial y^2}
\end{equation}

That notion of diffusion can be extended to work with graphs, where the concentrations of interest are stored in the nodes and diffuse through the edges. In this case, we need to define a diffusion rate $c$ that works similarly to $\kappa$ and describes the rate at which the elements will spread through the edges. While $c$ could be edge specific (the same way that $\kappa$ can be position specific in a non-homogeneous environment), for the sake of simplicity, we set $c$ to be identical for all edges. 

Then, based on the formulation of diffusion, the flow going through an edge connecting two nodes $i$ and $j$, with respective concentrations $s_i$ and $s_j$, in the $j$ to $i$ direction over a period $dt$, will be $c(s_j-s_i) dt$. Thus, the contribution of that edge to the concentrations $s_i$ and $s_j$ will be:

\begin{align}
\left.\frac{d s_i}{d t}\right\vert_{(i,j)} & = c (s_j - s_i) &
\left.\frac{d s_j}{d t}\right\vert_{(i,j)} & = c (s_i - s_j)
\end{align}

We can then extend that calculation to the whole graph:

\begin{equation}\label{eq:diffGraph}
    \frac{d s_i}{d t} = c \sum_{j=1}^N A_{i,j} (s_j - s_i)
\end{equation}

where $A$ is the adjacency matrix, defining $A_{i,j} = 1$ if nodes $i$ and $j$ are connected by an edge, and $0$ otherwise. Note that, since we are using undirected graphs, $A$ is necessarily symmetric.

When considering discrete time steps of duration $\tau$,  equation (4) becomes:

\begin{equation} \label{eq:si_diff}
s_i^{n+1}=s_i^{n}+c\tau\sum_{j=1}^N A_{ij} (s_j^{n}-s_i^{n})
\end{equation}
which is exactly the equation provided in the main text to describe how robots update their internal state.

\subsection{Convergence}
In this section, we discuss the convergence rate of the discrete graph Laplacian toward the continuous Laplacian. Here, we consider the more general context of~\cite{singer2006graph}. Following its terminology, we consider a Riemannian manifold $M$ of dimension $d=2$, and $N$ points $x_i$ which are independently and uniformly distributed over $M$. We consider a weight matrix $W$ defined as:

\begin{equation}
 W_{ij}=k \left( \frac{||\textbf{x}_i-\textbf{x}_j||^2}{2\epsilon}\right)
\end{equation}

where $k$ is the kernel function, and $\epsilon$ is the kernel bandwidth. In machine learning, a typical choice for the kernel function is $k(x)=e^{-x}$ (radial basis function kernel). However, in our experiments and simulations we use a step kernel ($k_\theta(x)=0$ if $x>\theta$, and $k_\theta(x)=1$ if $x \leq \theta$ for some constant $\theta$). The parameter $\sqrt{\epsilon}$ is homogeneous to a distance and plays the role of the field of perception in our setting. It is the distance over which the points can communicate with their neighbors. Thanks to that definition, we set $\theta = 1$ in the step kernel function. As such, we obtain in our case $W = A$, where $A$ is the adjacency matrix defined earlier.

To build the graph Laplacian, we then consider the diagonal matrix $D$ given by:
\begin{equation}
D_{ii}=\sum_{j=1}^{N} W_{ij}
\end{equation}

For a step kernel, $D$ is simply the degree matrix counting the number of neighbors of each point $\textbf{x}_i$ within a distance $\theta$.

We now define the matrix of the negative defined left-normalized discrete Laplacian:
\begin{equation}
\tilde{L}=D^{-1}W-I
\end{equation}

(Note that in our simulation and experiments, we consider the non-normalized discrete Laplacian $L = D - A$).

We consider a smooth test function $f$. Singer established the following uniform estimate of the error of the left-normalized discrete Laplacian compared to the continuous Laplacian~\cite{singer2006graph}.

\begin{equation}
\frac{1}{\epsilon}\sum_{j=1}^{N} \tilde{L}_{ij}f(\textbf{x}_i)=\frac{1}{2}\nabla f(\textbf{x}_i)
+O\left(\frac{1}{N^{1/2}\epsilon},\epsilon\right)
\end{equation}

This error estimate shows that for a fixed $\epsilon$, the left-normalized discrete Laplacian converges point-wise to the continuous Laplacian as $1/\sqrt{N}$, which is the classical scaling expected from the central law. Similarly to our numerical simulations on robots, it also predicts that for a fixed $N$, there is an optimal bandwidth $\epsilon$ that minimizes error because the error terms contains two sub terms that behaves oppositely with $\epsilon$. If $\epsilon$ is too large, all points are connected and the Laplacian matrix is an almost constant matrix (a clique connectivity) that carries little information over the local geometry. Conversely, if $\epsilon$ is too small, points are disconnected and the left-normalized Laplacian matrix does not encode  information about the geometry either. Balancing the two errors terms, Singer shows that the optimal $\epsilon$ is:

\begin{equation}
\epsilon = \frac{C(M)}{N^{1/4}}
\end{equation}

where $C(M)$ is a function that depends on the manifold $M$.
This result predicts an optimal scaling for the field of perception of the form $\sqrt{\epsilon}\sim N^{-1/8}$, which is sensibly different from the scaling $\sim N^{-1/2}$ that we find in numerical simulations for robots.
The discrepancies in the exponents may be explained by various factors: Singer considers Riemannian manifold without boundaries and point, while we consider shapes with boundaries with non-penetrating hard spheres. In addition we consider slightly different problems: Singer estimates the optimal field of perception for approximating the continuous Laplacian~\cite{singer2006graph}, while we estimate the optimal field of perception for classifying distinct shapes.

\subsection{Spectral gap}

In this section, we discuss the physical interpretation of $\lambda_2$ in light of spectral graph theory~\cite{fiedler1989laplacian}. In the general case, the spectral gap of a self-adjoint linear operator is the modulus of the smallest non-zero eigenvalue. The spectral gap plays a central role in graph theory and quantum mechanics, as it controls the dynamical properties of the system on which the operator is defined. In our case, if we assume that the graph is connected, and given that the first eigenvalue of the graph Laplacian $\lambda_1$ is always 0 (because the constant vector is an eigenvector with a null eigenvalue), the spectral gap is simply $\lambda_2$. As such, $\lambda_2$ is expected to provide information about the shape of the graph, which allows us to distinguish between the different arenas presented in the main text.

\subsubsection{Isoperimetric inequality and Cheeger constant}
The eigenvalue $\lambda_2$ of the graph Laplacian measures how easily information diffuses on a graph. A large $\lambda_2$ implies that information flows easily and smoothly between all parts of the graph, while a small $\lambda_2$ indicates the presence of bottlenecks in the flow. In the worst case, $\lambda_2=0$ implies that the graph is not connected, with at least two independent parts.

To see this quantitatively, let us first define the boundary $\partial(\mathscr{S})$ of a subset $\mathscr{S}$ of vertices as the set of edges that connect a vertex inside S to a vertex outside A.

\begin{equation}
\partial(\mathscr{S})=\{(u,v)\in \mathscr{E} \, | \, u\in \mathscr{S}, v\notin \mathscr{S}\}
\end{equation}

The isoperimetric ratio of $\mathscr{S}$ measures the size of the boundary $\partial(\mathscr{S})$ with respect to the size of $\mathscr{S}$:

\begin{equation}
h(\mathscr{S})=\frac{|\partial(\mathscr{S})|}{|\mathscr{S}|}
\end{equation}
where $|\,.\,|$ is the sums of degrees $d$ of vertices in a set:
\begin{equation}
|\mathscr{S}| = \sum_{x \in \mathscr{S}} d_x
\end{equation}

We can then define the Cheeger constant, or isoperimetric number of a graph G, as the minimum isoperimetric number taken over all subsets that have less than half of the total number of vertices. 

We additionally define $\Delta(G)$ as the maximum degrees of vertices of the graph G. We can now state the Cheeger inequalities \cite{chung1997spectral,molitierno2016applications} which provide lower and upper bounds for the eigenvalue $\lambda_2$ with the isoperimetric number $h(G)$:

\begin{equation}
h(G)\geq \frac{\lambda_2}{2} \geq \frac{h^2(G)}{2\Delta(G)}
\end{equation}

This inequality is central in spectral theory as it directly relates local properties that are controlled by diffusion (\eg the rate of diffusion of information) with global properties that are set by the shape of a graph. For instance, on graphs with large $\lambda_2$, $h(G)$ is large and information flows easily because all their subsets have large boundaries and are well connected to other subsets. Conversely, if a graph $G$ is not connected, there exists two subsets that are not connected so $h(G)=0$, and the Cheeger inequality forces  $\lambda_2=0$.

\FloatBarrier\clearpage

\section{Experimental setup}

\subsection{Robots and Arenas}

\begin{figure*}[h!]

\centering
\resizebox{0.80\textwidth}{!}{%
\begin{tabular}{c}
  \includegraphics[height=0.20\textheight]{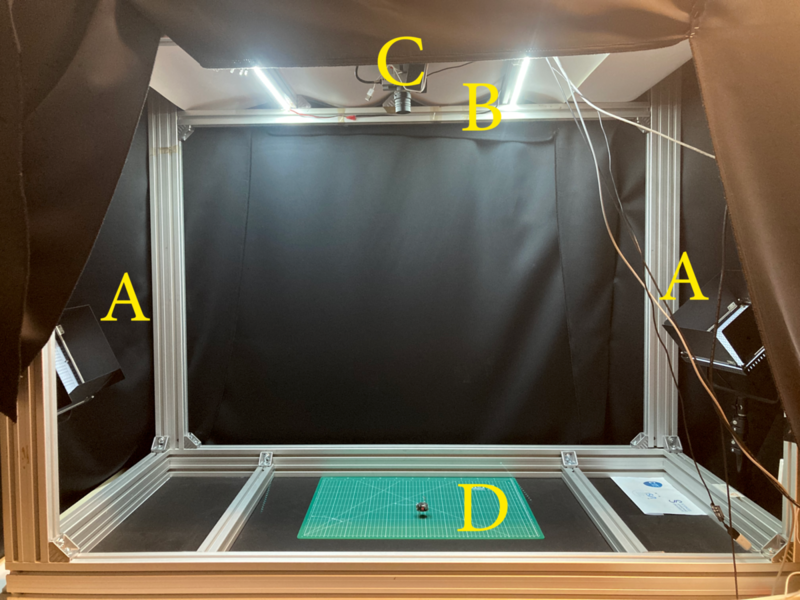}
  \includegraphics[height=0.20\textheight]{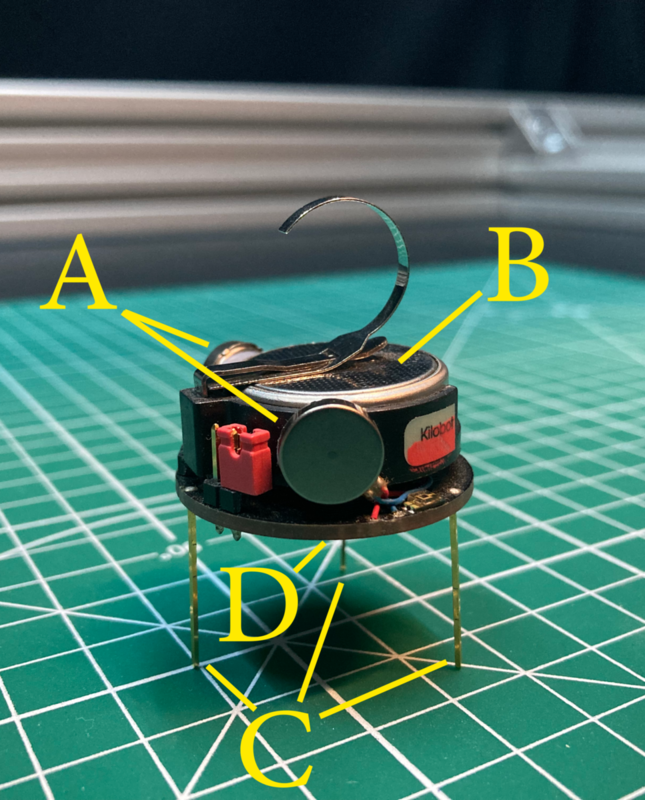} \\

  \includegraphics[height=0.20\textheight]{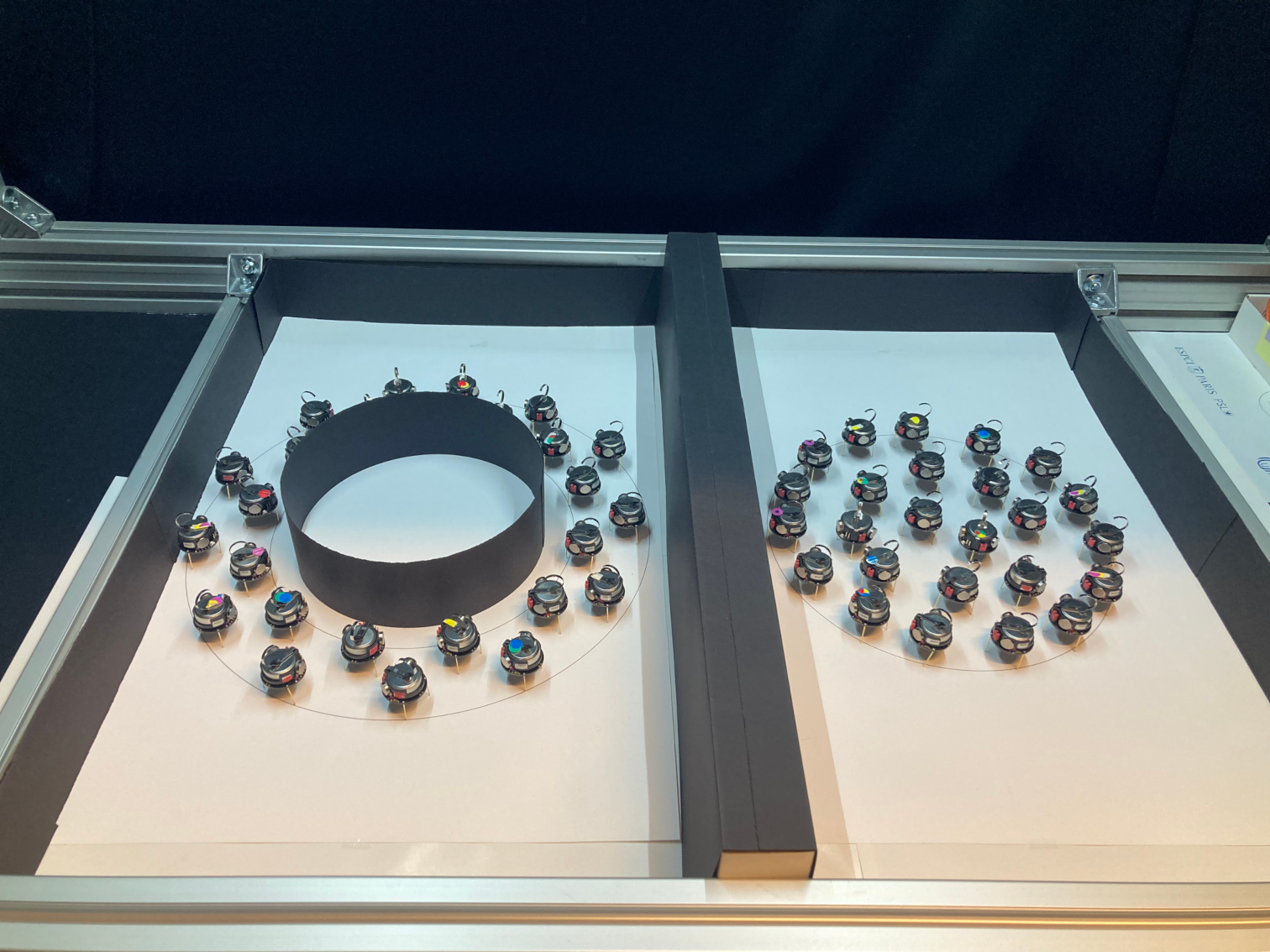}
  \includegraphics[height=0.20\textheight]{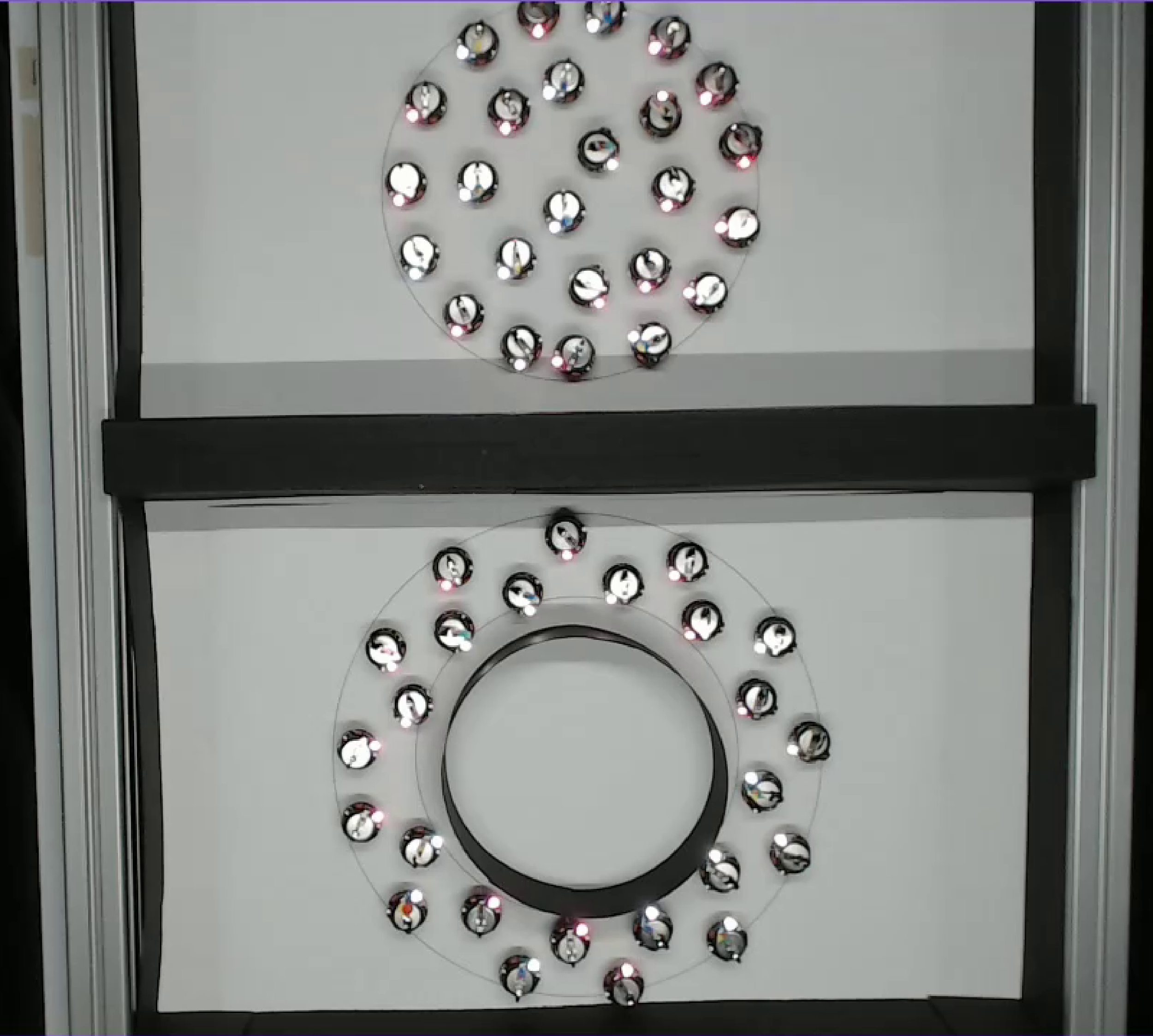} \\

  \includegraphics[height=0.20\textheight]{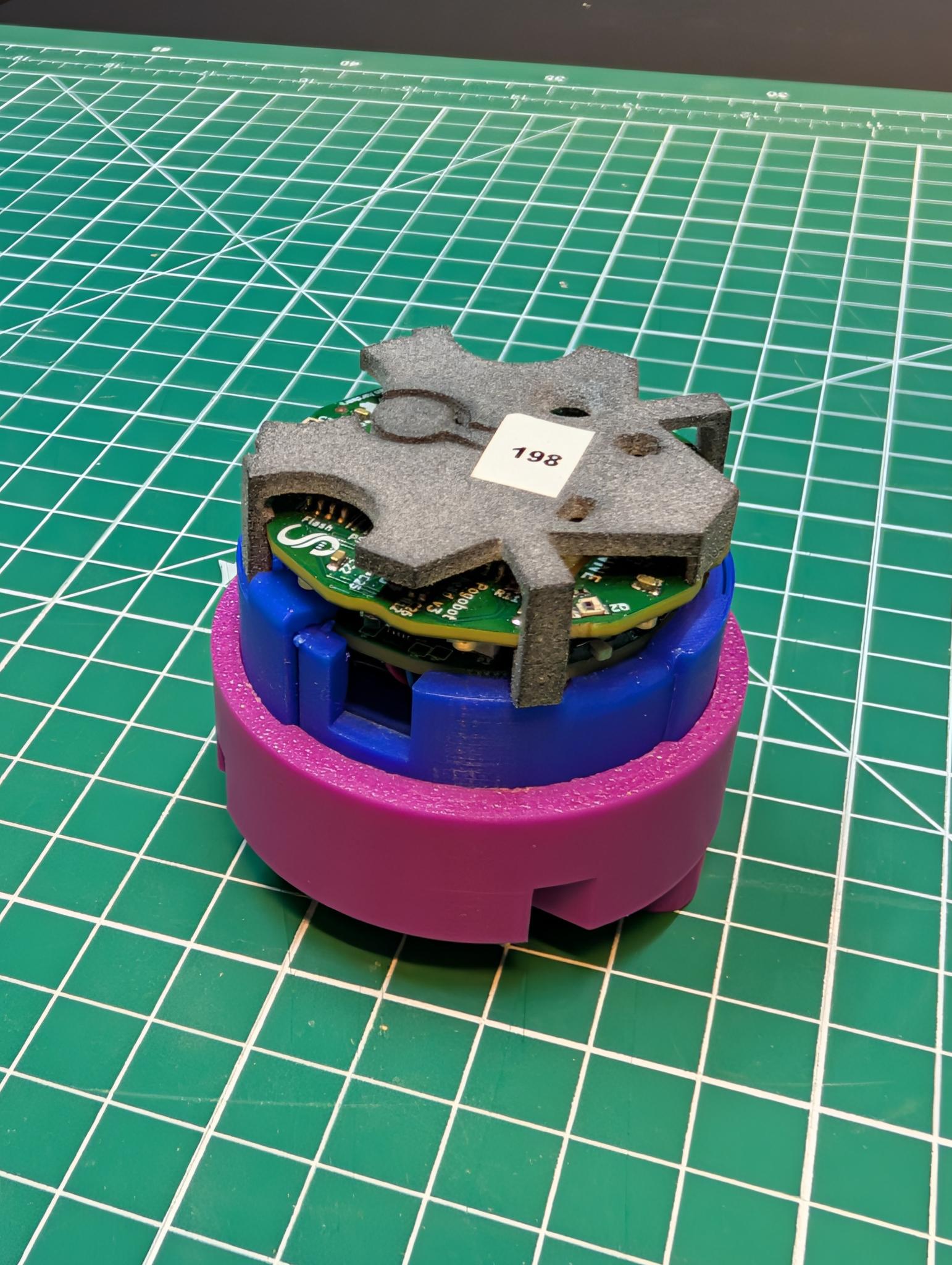}
  \includegraphics[height=0.20\textheight]{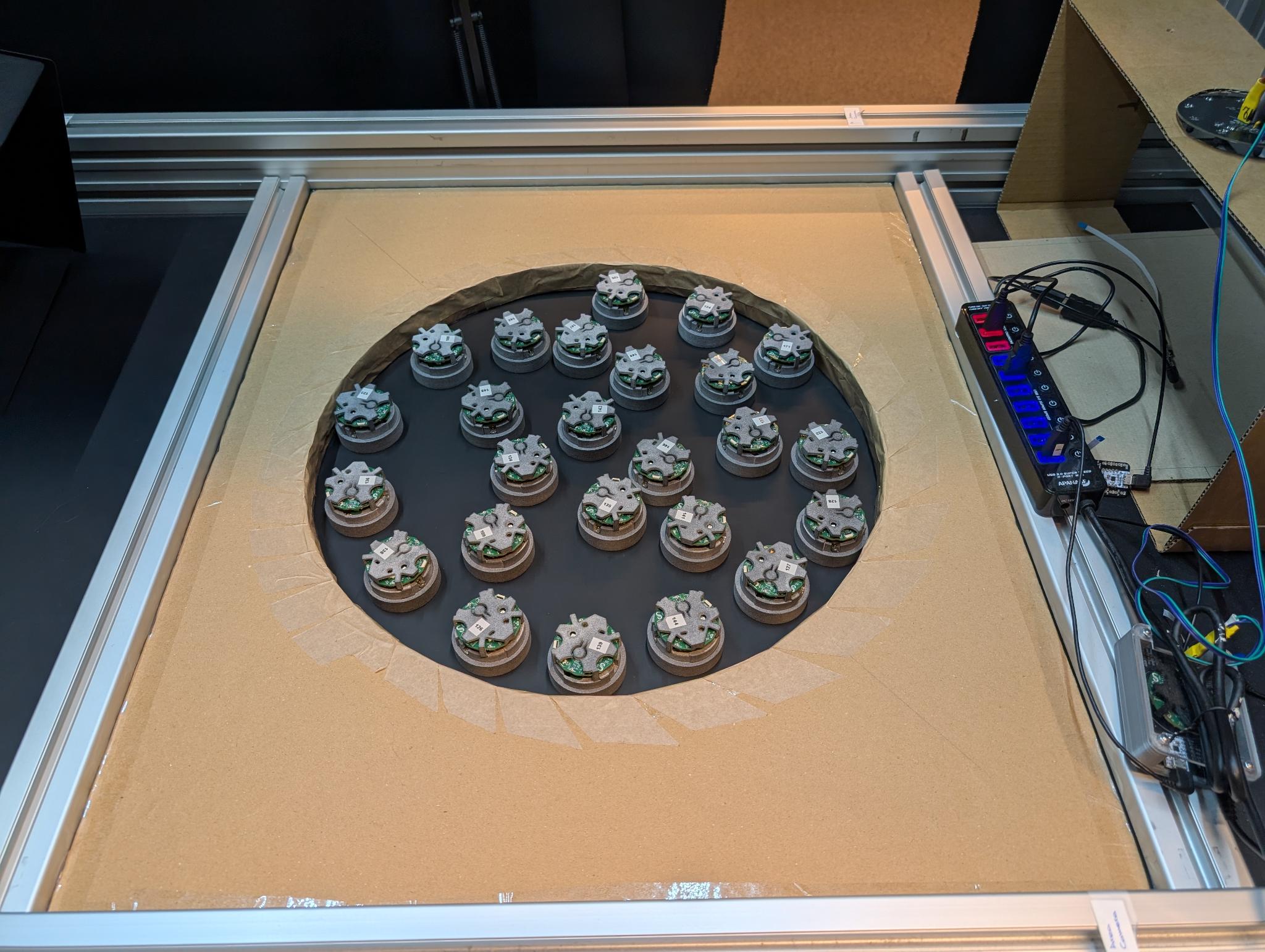}
\end{tabular}
}

\caption{General overview of the experimental setups used for the robotics experiments. \footnotesize{\textbf{Top-left}: general view of the experimental setup, which is closed using light-occluding black curtains. The robot area is 152~cm per 72~cm, within which a smaller area (88~cm per 72~cm) is physically delimited. The green floor (D) placed on the robot area is a cutting map on which stands a single Kilobot robot. Two Pixel K80 professional LED photography light projectors (A) are placed on either side of the setup, pointing upwards with an angle of $45\deg$, and complemented with two rails of LED (White 5700K) placed on the ceiling, for a global illumination of 506-584 lux (B). A Logitech C920s HD Pro Webcam (C) and a Pixelink PL-D734MU camera (not used here) are placed in the center of the ceiling (115~cm above the robot area) and point downwards. The ceiling is covered in white matte paper \textbf{Top-right}: a Kilobot robot (diameter$=33$mm). The Kilobot features two vibrating motors (A), one 3.3V rechargeable battery, three metallic rods (C) and an infrared LED transmitter and photodiode receiver (D) placed underneath the robot and pointing downwards. \textbf{Middle-left}: the annulus (left) and disk (right) configurations, using $25$ Kilobots. The disk configuration has a radius of 150mm and a surface of $\approx 70685.8$ mm$^2$. The annulus configuration has an external radius of $200mm$, internal radius of $133mm$ and a surface of $\approx 70092.1 \texttt{mm}^2$. \textbf{Middle-right}: same as above, acquired from the overhead webcam. A matte white floor is used, as well as a black disk shape for the annulus configuration used to physically forbid communication between robots from the opposite region of the annulus.
\textbf{Bottom-left}: a Pogobot robot (diameter$=53$mm). The Pogobot features modular motors and 4 high-throughput directional IR emitters and receivers placed around the robot body. \textbf{Bottom-right}: the annulus configuration with 25 Pogobot robots. Both the annulus and disk arenas have a normalized surface (accessible to robots, within walls) of $\approx 1.5e5$ mm$^2$.}}
\label{fig:viewSetup}
\end{figure*}

Supplementary Figure~\ref{fig:viewSetup} shows the experimental setup used at Sorbonne Universit\'{e} for conducting all robotic experiments. We use two arenas (disk and annulus), each populated with $25$ robots. The robotic platforms used are: (1) Kilobots~\cite{rubenstein2012kilobot,rubenstein2014programmable}, originally designed at Harvard University and purchased from K-Team SA; and (2) Pogobots~\cite{loi2025pogobot}, a recent swarm robotics platform developed at Sorbonne Universit\'{e} (ISIR laboratory).

Kilobots were designed to minimize cost and maximize scalability for thousand-robot collectives~\cite{rubenstein2012kilobot}, whereas Pogobots extend this philosophy toward higher locomotion speed, directional high-throughput infrared (IR) communication, FPGA-based computation, and programmable active-matter experiments~\cite{loi2025pogobot}. 

The main technical differences are as follows. Pogobots embed a significantly more powerful processing architecture (FPGA with a 32-bit RISC-V softcore) compared to the 8-bit microcontroller used in Kilobots; they support longer IR communication range and substantially higher communication bandwidth (accounting for packet collisions, typically $\approx 3$--$9$ bytes/s in dense Kilobot collectives versus several kB/s for Pogobots under comparable conditions). Conversely, unlike Kilobots, Pogobots do not estimate neighbor distance from IR signal intensity, as their communication system prioritizes directional high-speed transmission rather than analog intensity-based ranging.

Supplementary Table~\ref{tab:kilobot_pogobot_comparison} provides a detailed comparison of the two robotic platforms.

\begin{table}[htbp]
\centering
\small
\begin{tabular}{p{0.32\linewidth}p{0.31\linewidth}p{0.31\linewidth}}
\toprule
\textbf{Property} & \textbf{Kilobot} & \textbf{Pogobot} \\
\midrule
Diameter & 33 mm & 53 mm \\
Mass & $\sim$16 g & $\sim$43--56 g \\
Locomotion & Vibration-based (slip–stick) & Modular: wheels or vibration-based \\
Max linear speed & $\sim$1 cm/s & up to 25 cm/s (wheels); $\sim$7 cm/s (TPU legs) \\
Max angular speed & $\sim$0.78 rad/s & up to 15 rad/s (nylon vibration mode) \\
Communication type & Infrared (1 emitter + 1 receiver, downward reflection) & 4 directional IR emitter/receiver pairs \\
Communication carrier & $\sim$38 kHz IR modulation & 1–10 MHz optical carrier \\
Max communication rate & 18 bytes per second (max rate with only 2 robots), around 3-9 bytes per second in dense clusters & several kB per second in dense clusters \\
Max payload size & few bytes (9-byte packets typical) & up to 391 bytes payload \\
Communication range & up to 8-10 cm center to center & up to 10 cm (border to border) \\
Distance estimation & Yes (via IR intensity, $\pm$2 mm accuracy) & No \\
Directional sensing & No (isotropic reflection) & Yes (4 IR sensors) \\
Ambient light sensing & Yes (1 sensor) & Yes (3 photosensors: 2 front, 1 rear) \\
IMU & No & Yes (6-axis IMU) \\
Onboard computation & Atmega328 @ 8 MHz, 32 kB flash & FPGA (ICE40UP5K) + 32-bit RISC-V softcore \\
Battery & 3.4 V, 160 mAh Li-ion & 3.3 V LiFePO$_4$ CR2 \\
Autonomy & 3--24 h (activity-dependent) & 60--180 min (activity-dependent) \\
Charging & Collective conductive charging surface & Dedicated Pogocharger (up to 9 robots per unit) \\
Scalable programming & Overhead IR controller (bootloader) & Cable or IR (Pogoshower, directional IR programming, Pogowall) \\
Open-source & Yes & Yes \\
Retail price & 100 € & 250 € \\
\bottomrule
\end{tabular}
\caption{Comparison between Kilobot and Pogobot platforms~\cite{rubenstein2012kilobot,loi2025pogobot}.}
\label{tab:kilobot_pogobot_comparison}
\end{table}

All experiments are conducted while the whole setup is isolated from external perturbation using light-occluding black curtains placed on all sides and above the setup. The whole setup is enclosed in a construction of an aluminum section that supports the camera and lighting (see Caption of Supplementary Fig.~\ref{fig:viewSetup} for an exhaustive description).

Setting up each experiment starts with removing the dust on the arena by using a wipe with a very small quantity of water to ensure maximum ground reflection. The target area (Disk or Annulus) is drawn on the floor to help position the robots before the experiment starts. Kilobots are then initially placed randomly in the target area, broadly respecting a uniform distribution (\ie clusters are avoided), with random orientation. Initial positions thus vary from one run to another. Each experiment starts with all robots at full charge. The experiment is started using the infrared overhead controller, and the black curtains are closed within 10 seconds. 

Monitoring is performed with a dedicated webcam connected to a desktop computer. The image is intentionally blurred so that each Kilobot's overhead LED stands out as a blurred circle of color that corresponds to the current state of the robot.

\subsection{Controller design and implementation of the algorithms on kilobots}

We implement the SSR algorithm in Kilobots and Pogobots simulations using respectively two simulators: Kilombo~\cite{jansson2015kilombo} and Pogosim~\cite{cazenille2025pogosim}. They both allow the same respective C code to be used both in simulations and as controllers of actual robots, without any modification. Both simulators model collisions through a physics engine, and are relatively realistic, reproducing communication loss, sensor and motor biases.
The C source code used for simulations and experiments can be found in a supplementary information archive and in the repository: \url{https://github.com/leo-cazenille/spectral-swarm-robotics}.
Supplementary Table~\ref{tab:LEDcolors} lists the colors displayed by the robots LEDs depending on their current behavior.

\begin{table*}[h]
\begin{center}
\resizebox{0.40\textwidth}{!}{%
\renewcommand{\arraystretch}{1.2}
\begin{tabular}{r|l|l|}
\cline{2-3}
& LED color & Description \\
\cline{2-3}
\multirow{2}{*}{Dispersion}
& Red & Tumble behavior \\
& Green & Run behavior \\
\cline{2-3}
\multirow{4}{*}{Diffusion}
& White & Pre-diffusion stage \\
& Blue & $\texttt{sign(}s_i^t\texttt{)} > 0$ \\
& Red & $\texttt{sign(}s_i^t\texttt{)} < 0$ \\
& Grey & $\texttt{sign(}s_i^t\texttt{)} = 0$ \\
\cline{2-3}
\multirow{7}{*}{$\lambda_2$ estimation}
& Cyan      & \textbf{Disk} detected \\
& Orange    & \textbf{Square} detected \\
& Green     & \textbf{Arrow} detected \\
& Red       & \textbf{Star} detected \\
& Gold      & \textbf{Triangle} detected \\
& Brown     & \textbf{Stop} detected \\
& Violet    & \textbf{Annulus} detected \\
\cline{2-3}
\end{tabular}
}
\caption{Table of LED colors displayed by the robots depending on their current behavior.}
\label{tab:LEDcolors}
\end{center}
\end{table*}

\subsubsection{Implementation challenges}\label{sec:impl_challenges}
There is a number of pitfalls to keep in mind when implementing the spectral swarm robotics algorithm on Kilobots and Pogobots.

First of all, robots have problems to communicate due to collisions between messages. The more neighbors there are, the more collisions there will be. Furthermore, there is variability between each robot in their ability to receive and send messages: some robots will have more powerful IR transmitters than others, or be affected by different environmental artifacts (background light, presence of ground particles, etc). This discrepancy in the ability of the robots to communicate also varies during the experiment, even if the robots are immobile. This leads to cases where a robot can communicate with a neighbor, but that neighbor cannot communicate back -- which, in turn, precipitates the emergence of asymmetries in the Laplacian matrix (which are supposed to be symmetrical). If the number of asymmetries is sufficiently large, several robots may have diffusion sessions that become divergent (\ie they no longer decay exponentially). Such errors will rapidly impact the entire swarm by being propagated by the diffusion process, biasing the results of the algorithm. This problem is partially alleviated through the Handshake mechanism of the spectral swarm robotics algorithm, as described in Sec.~\ref{sec:handshakes}. However, it is often not sufficient to avoid the presence of asymmetries: cases where a robot will not be able to communicate with a given neighbor only for a small number of steps still remain.
We have found that working with low values of $\tau$ (\eg $\tau = 1/50 s$) decreases the influence of asymmetries on the diffusion process, as it reduces the global rate of the diffusion process. Furthermore, the amount of time robots have to broadcast and receive messages at each step needs to be adjusted in order to leave enough time for the robots to communicate with all their neighbors (taking into account message collisions).

Second, the robots need to be temporally synchronized so that they are able to all execute the same stages of the algorithm at the same time. On \textbf{Kilobots}, we rely on the internal ``kiloticks" counter: it is initialized at 0 at the start of an experiment, and incremented at a frequency of $31$ Hz. It is thus important that all robots start an experiment at the exact same time. There are small variabilities in the clock frequencies of the robots, and they all have a small time lag. During the diffusion, some robots may be 1-4 steps behind the others. Again, this can be partially alleviated by using small values of $\tau$, as it will reduce the impact of bias and errors in the diffusion process. Furthermore, variabilities in clock frequency are influenced by the charge of the Kilobots battery -- as such, we start each experiment with robots that are all fully charged.
On \textbf{Pogobots}, the increased message size allows the robots to directly synchronize their internal clock by broadcasting their local time in each message. This alleviates most time synchronization problems (cf Sec.~\ref{sec:noiseRes}).

Third, the diffusion can degenerate after a certain time-step (\eg if time lags or large number of asymmetries in the Laplacian corrupted some robots' internal state). In this case, the degenerated part of the diffusion should not be taken into account for the $\texttt{Indiv}\lambda_{2_i}$ calculation. This is implemented by computing $\texttt{Indiv}\lambda_{2_i}$ at each step of the diffusion by using OLS, and computing the mean squared error (MSE) of the fit. At the end of the diffusion, only the fit with the lowest MSE is kept as the value of $\texttt{Indiv}\lambda_{2_i}$.

\subsubsection{Communication between robots}

Kilobots communication is very slow: they can broadcast at most $9$ bytes of payload per messages, twice per second ($18$ bytes per second). This is a theoretical maximum rate that will rarely be reached. When the amount of neighbors increases, so does the amount of message collisions, resulting in a very low number of correctly received messages in such setting: $\approx 10-30\%$ with 8 neighbors \ie $\approx 1.8-5.4$ bytes per second.

In turn, Pogobot communication is far faster: up to 120 messages of 20-100 bytes per second up to $135$mm (center to center), and are $\approx 30-500$ times faster than Kilobots. In turn, Pogobots lack the ability to estimate communication distance between neighbors, a feature implemented by the Kilobots -- it is thus not possible to control the field of perception $\sigma$ with Pogobots.

Supplementary Table~\ref{tab:packets} describes the content of the messages broadcasted by the robots for each stage of the algorithm.

In the \textit{handshake} stage, each robot keep a list of known neighbors. The payload of the messages contains the unique identifiers (UID) of 3 selected known neighbors. The selection of known neighbors from the list iterates from message to message, in a round-robin fashion.

In the \textit{diffusion} stage, the messages contain the values of $s_i^t(a,d)$ for agent $i$, at diffusion step $t$, during iteration $a$, and for all $P = 3$ diffusion sessions $d$ computed in parallel. Due to size constraints of the message, the values of $s_i^t(a,d)$ are transmitted as half-precision float, while they are stored as single-precision float in the robots.

In the \textit{collective averaging} stage, the messages contain the current $\texttt{Consensus}\lambda_{2_i}$ value of agent $i$.

\begin{table*}[h]
\begin{center}

\resizebox{1.00\textwidth}{!}{%
\renewcommand{\arraystretch}{1.1}
\begin{tabular}{l |l | p{3.5cm} | l |p{10cm}|}
\cline{2-5}
\parbox[t]{3mm}{\multirow{8}{*}{\rotatebox[origin=c]{90}{Handshake}}}
& Variable name & Type & Size (bits) & Description \\
\cline{2-5}
& kilo\_uid & uint16 & 16 & Kilobot unique identifier \\
& known\_uid0 & uint16 & 16 & Unique identifier of already known neighbor \\
& known\_uid1 & uint16 & 16 & Unique identifier of already known neighbor \\
& known\_uid2 & uint16 & 16 & Unique identifier of already known neighbor \\
& - & - & 8 & Empty \\
\cline{2-5}
& type & uint8 & 8 & Type of message (\textbf{handshake}, diffusion, coll. averaging, etc) \\
& CRC & uint16 & 16 & Cyclic redundancy checksum \\
\cline{2-5}
\end{tabular}
}
\\[0.25cm]

\resizebox{1.00\textwidth}{!}{%
\renewcommand{\arraystretch}{1.1}
\begin{tabular}{l |l | p{3.5cm} | l |p{10cm}|}
\cline{2-5}
\parbox[t]{3mm}{\multirow{8}{*}{\rotatebox[origin=c]{90}{Diffusion}}}
& Variable name & Type & Size (bits) & Description \\
\cline{2-5}
& kilo\_uid & uint16 & 16 & Kilobot unique identifier \\
& $s_i^t(a,0)$ & half-precision float & 16 & Diffused value, for iteration $a$ and diffusion session $0$ \\
& $s_i^t(a,1)$ & half-precision float & 16 & Diffused value, for iteration $a$ and diffusion session $1$ \\
& $s_i^t(a,2)$ & half-precision float & 16 & Diffused value, for iteration $a$ and diffusion session $2$ \\
& - & - & 8 & Empty \\
\cline{2-5}
& type & uint8 & 8 & Type of message (handshake, \textbf{diffusion}, coll. averaging, etc) \\
& CRC & uint16 & 16 & Cyclic redundancy checksum \\
\cline{2-5}
\end{tabular}
}
\\[0.25cm]

\resizebox{1.00\textwidth}{!}{%
\renewcommand{\arraystretch}{1.1}
\begin{tabular}{l |l | p{3.5cm} | l |p{10cm}|}
\cline{2-5}
\parbox[t]{3mm}{\multirow{6}{*}{\rotatebox[origin=c]{90}{Coll. avg.}}}
& Variable name & Type & Size (bits) & Description \\
\cline{2-5}
& kilo\_uid & uint16 & 16 & Kilobot unique identifier \\
& $\texttt{Consensus}\lambda_{2_i}$ & single-precision float & 32 & Broadcasted consensus value \\
& - & - & 24 & Empty \\
\cline{2-5}
& type & uint8 & 8 & Type of message (handshake, diffusion, \textbf{coll. averaging}, etc) \\
& CRC & uint16 & 16 & Cyclic redundancy checksum \\
\cline{2-5}
\end{tabular}
}

\caption{Description of the data packets transmitted from robot to robot during the 
\textbf{handshake}, \textbf{diffusion} and \textbf{collective averaging} stages of the spectral swarm robotics algorithm. The maximal size of kilobots messages payload (\ie data, without type and CRC) is 72 bits (9 bytes). }
\label{tab:packets}
\end{center}
\end{table*}

\subsection{Handshake messages}\label{sec:handshakes}

Kilobots and Pogobots communicate by broadcasting a message to their neighborhood through infrared signals, resulting in potential conflicts if two or more robots are broadcasting at the same time. Message integrity is ensured through a CRC mechanism. However, the portion of correctly received messages depend on the number of neighbors. The proportion of correctly received messages can be quite low ($\approx 10-30\%$ when 4-8 Kilobots broadcast messages in the same neighborhood. 

Agents will employ a handshaking process to identify their ``stable" neighbors, or the collection of neighbors $k$ for a focal agent $i$, such that, in most cases, $i$ can transmit messages to $k$, and $k$ can reciprocate. Indeed, the actual field of perception $\sigma$ of robots can vary due to minor physical distinctions (\eg infrared sensors) or environmental factors (\eg marginally increased IR luminescence in certain parts of the arena). This may lead to situations where agent $i$ can communicate with $k$, but not the other way around. However, our algorithm relies upon the Laplacian of the communication graph, which is assumed to be symmetrical. Consequently, it is crucial to establish which neighbors are reliable.

In our approach, each focal agent $i$ will continuously broadcast handshake messages containing its own identifier, and the identifier of agents that it has previously successfully received messages from. As such, if agent $i$ receive a message from agent $k$ confirming that $k$ can receive a message from $i$, it means that there is a bidirectional communication channel between $i$ and $k$.
This process is repeated several time: at each new round of handshaking, agents will first empty their list of neighbors, then continuously broadcast for some time. After a fixed budget of time, agents stop broadcasting handshake messages, and will assess which agents they have a stable bidirectional communication channel with, on all handshaking rounds. On each agent, the latter are regrouped into a list of stable neighbors.

\subsection{Analysis and plotting scripts}
All analysis and plotting scripts were written in Python 3.10. The following libraries were used to perform analyses: NumPy~\cite{harris2020array}, Pandas~\cite{mckinney2010data}, NetworkX~\cite{hagberg2020networkx}, SciPy~\cite{2020SciPy-NMeth}, Geopandas~\cite{kelsey_jordahl_2020_3946761} and shapely~\cite{shapely2007}. Plotting was realized through the Matplotlib~\cite{Hunter:2007} and Seaborn~\cite{Waskom2021} libraries.
Pairwise comparisons of estimated $\lambda_2$ distributions (Main Text Fig.~\ref{fig:statsSimuVsExpes} and Supplementary Fig.~S14) and spatial-distribution metrics (node degree, normalized nearest-neighbor distance $d_{\min}/\sigma$, and Voronoi cell area; Supplementary Fig.~S7) were performed using Mann--Whitney--Wilcoxon tests with Bonferroni correction for multiple comparisons. Classification performance was summarized using accuracy scores and confusion matrices. Sample sizes, units of observation, and definitions of plotted variability are specified in the corresponding figure legends (generally 64 runs, with some cases using 32). Source data, individual $P$ values, and the comparison groups included in each Bonferroni correction are available in data file S1 accompanying the published article (\url{https://doi.org/10.1126/scirobotics.adr9578}).

\FloatBarrier\clearpage

\section{Supplementary Results}

Here, we provide additional results that provide a more detailed analysis compared to those of the main text. Specifically, we investigate in simulations:
\begin{itemize}
    \item the influence of the number of iterations on the accuracy of the spectral swarm robotics algorithm, first in the case with 7 arenas and then focusing on the 2 most different arenas in terms of $\texttt{Final}\lambda_2$.
    \item the influence of the simulation parameters (number of agents $N$, field of perception $\sigma$) on the order of the $\texttt{Final}\lambda_2$ values relative to the shape of arenas.
    \item how the dispersion scheme used and initial position of the agents can influence the results of the algorithm, and whether having moving agents is necessary.
\end{itemize}

Supplementary Table~\ref{tab:cases} lists the simulation and experimental cases conducted in this paper. The SI present the results of the five first cases (Supplementary Figs.~\ref{fig:regimes_simlarge7}-~\ref{fig:regimes_simsmall2uniform}). The main text presents the results of the first case (Main Text Figs.~4, 5, and 6) as well as the remaining four cases: \textbf{sim-2}, \textbf{expe-2}, \textbf{pogo-sim-2}, \textbf{pogo-expe-2} (Main Text Figs. 6, 7 and 8).

The parameters values used for all cases are listed in Supplementary Table~\ref{tab:params}.

\begin{table*}[h]
\begin{center}
\resizebox{1.00\textwidth}{!}{%
\renewcommand{\arraystretch}{1.5}
\begin{tabular}{|l |p{2.0cm} |p{2.0cm} |p{3.0cm} |p{1.9cm} |p{3.0cm} |p{1.0cm} |p{1.4cm} |l |p{1.1cm} |}
\hline
Case name & Robot type & Simulations or experiments & Dispersion alg. & Surface of arenas $S$ & Arenas & Nr. of runs & Nr. of agents $N$ & Lenght scale $l$ & Nr. of iterations $I$ \\
\hline
\hline
sim-large-7 & Kilobots & Simulations & run-and-tumble & $500000 mm^2$ & Disk, Square, Arrow, Star, Triangle, Stop, Annulus & 64 & $50 - 550$ & $100.0 - 30.15$ & 1-30 \\
sim-large-2 & Kilobots & Simulations & run-and-tumble & $500000 mm^2$ & Disk, Annulus & 64 & $50 - 550$ & $100.0 - 30.15$ & 1-30 \\
\hline
sim-small-2-dispersion & Kilobots & Simulations & run-and-tumble & $70000 mm^2$ & Disk, Annulus & 64 & $10 - 80$ & $83.67 - 29.58$ & 1-10 \\
sim-small-2-random & Kilobots & Simulations & None - immobile robots initialized at random positions& $70000 mm^2$ & Disk, Annulus & 64 & $10 - 80$ & $83.67 - 29.58$ & 1-10 \\
sim-small-2-uniform & Kilobots & Simulations & None - immobile robots initialized uniformly in the arenas& $70000 mm^2$ & Disk, Annulus & 64 & $10 - 80$ & $83.67 - 29.58$ & 1-10 \\
\hline
sim-2 & Kilobots & Simulation & None - immobile robots initialized uniformly in the arena & $70000 mm^2$ & Disk, Annulus & 64 & 25 & $52.91$ & 1 \\
expe-2 & Kilobots & Experiments & None - immobile robots initialized uniformly in the arena & $70000 mm^2$ & Disk, Annulus & 15 & 25 & $52.91$ & 1 \\
\hline
pogo-sim-2 & Pogobots & Simulation & None - immobile robots initialized uniformly in the arena & $150000 mm^2$ & Disk, Annulus (main text) and Triangle (SI only) & 64 & 25 & $77.46$ & 3 \\
pogo-expe-2 & Pogobots & Experiments & None - immobile robots initialized uniformly in the arena & $150000 mm^2$ & Disk, Annulus (main text) and Triangle (SI only) & 15 & 25 & $77.46$ & 3 \\
\hline
\end{tabular}
}
\caption{List of all considered cases in simulation and experiments.}
\label{tab:cases}
\end{center}
\end{table*}

\begin{table*}[b!]
\begin{center}
\resizebox{0.95\textwidth}{!}{%
\renewcommand{\arraystretch}{1.5}
\begin{tabular}{|p{0.2cm} |p{4.5cm} |p{2.0cm} |p{2.0cm} |p{2.0cm} |p{2.0cm} |p{2.0cm} |p{2.0cm}| p{2.0cm}| p{2.0cm}| p{2.0cm}|}
\hline
& Parameter Name & sim-large-7 & sim-large-2 & sim-small-2-dispersion & sim-small-2-random & sim-small-2-uniform & sim-2 & expe-2 & pogo-sim-2 & pogo-expe-2  \\
\hline
\hline
\parbox[t]{2mm}{\multirow{15}{*}{\rotatebox[origin=c]{90}{Base}}}
 & Robot type & \multicolumn{7}{|c|}{Kilobots} & \multicolumn{2}{|c|}{Pogobots} \\
\cline{2-11}
 & Nr. of runs & \multicolumn{6}{|c|}{$64$} & 15 & 64 & 15 \\
\cline{2-11}
 & Arena Surface $S$ & \multicolumn{2}{|c|}{$500000 mm^{2}$} & \multicolumn{5}{|c|}{$70000 mm^{2}$} & \multicolumn{2}{|c|}{$150000 mm^{2}$} \\
\cline{2-11}
 & Agent diameter & \multicolumn{7}{|c|}{$33$ mm} & \multicolumn{2}{|c|}{$53$ mm}\\
\cline{2-11}
 & Field of Perception $\sigma$ (center to center) & \multicolumn{5}{|c|}{$30$mm-$320$mm} & \multicolumn{2}{|c|}{$\approx85$ mm} & \multicolumn{2}{|c|}{$\approx135$ mm} \\
\cline{2-11}
 & Nr. of agents $N$ & \multicolumn{2}{|c|}{50-550} & \multicolumn{3}{|c|}{10-80} & \multicolumn{4}{|c|}{25} \\
\cline{2-11}
 & Initial $s^0$ value & \multicolumn{9}{|c|}{$-1$ or $1.$ chosen randomly, uniform distribution is *not* enforced} \\
\cline{2-11}
 & $\tau$ & \multicolumn{5}{|c|}{$1.0 / 15.0$s} & \multicolumn{2}{|c|}{$1.0 / 50.0 s$} & \multicolumn{2}{|c|}{$0.15$s} \\
\cline{2-11}
 & Number of iterations $I$ & \multicolumn{2}{|c|}{$1$-$30$} & \multicolumn{3}{|c|}{$1$-$10$} & \multicolumn{2}{|c|}{$1$} & \multicolumn{2}{|c|}{$3$} \\
\cline{2-11}
 & Initial position of agents & \multicolumn{3}{|c|}{Packed in the center of the arena} & Random positions & \multicolumn{5}{|c|}{Uniformly distributed (equidistant to neighbors)} \\
\cline{2-11}
 & Prop. of correctly received messages between neighbors & \multicolumn{5}{|c|}{$100\%$} & \multicolumn{2}{|c|}{$\approx 10\%$} & \multicolumn{2}{|c|}{$\approx 20\%$} \\
\cline{2-11}
 & Number of diffusion sessions computed in parallel per it. & \multicolumn{9}{|c|}{3} \\
\cline{2-11}
 & Max nr. of neighbors $H$ & \multicolumn{9}{|c|}{20} \\
 
\hline
\hline
\parbox[t]{2mm}{\multirow{19}{*}{\rotatebox[origin=c]{90}{Time}}}
 & Simulation/Experiment time & \multicolumn{2}{|c|}{$1143900$ kt $\approx 615$ min} & \multicolumn{1}{|c|}{$257300$ kt $\approx 138$min} & \multicolumn{2}{|c|}{$148800$ kt = $80$ min} & \multicolumn{2}{|c|}{ $146630$ kt $\approx 79$min} & \multicolumn{2}{|c|}{$960$ s = $16$ min}  \\
\cline{2-11}
 & Initial dispersion duration & \multicolumn{3}{|c|}{$46500$ kt = $25$ min} & \multicolumn{6}{|c|}{$0$ kt}  \\
\cline{2-11}
 & Iteration total duration & \multicolumn{2}{|c|}{$36580$ kt $\approx 20$ min} & \multicolumn{1}{|c|}{$21080$ kt = $680$s} & \multicolumn{2}{|c|}{$14880$ kt = $480$s} & \multicolumn{2}{|c|}{$146630$ kt = $79$min} & \multicolumn{2}{|c|}{$320$ s $\approx 5$ min} \\
\cline{2-11}
 & Iteration waiting time & \multicolumn{7}{|c|}{$930$ kt = $30$ s} & \multicolumn{2}{|c|}{$2.5$ s} \\
\cline{2-11}
 & Iteration handshake & \multicolumn{5}{|c|}{$310$ kt = $10$ s} & \multicolumn{2}{|c|}{$6200$ kt = $200$ s} & \multicolumn{2}{|c|}{$25$ s} \\
\cline{2-11}
 & Iteration handshake step dur. & \multicolumn{5}{|c|}{$31$ kt = $1$ s} & \multicolumn{2}{|c|}{$248$ kt = $8$ s} & \multicolumn{2}{|c|}{$2.5$ s} \\
\cline{2-11}
 & Iteration dispersion duration & \multicolumn{3}{|c|}{$6200$ kt = $200$ s} & \multicolumn{6}{|c|}{$0$ kt} \\
\cline{2-11}
 & Iteration diffusion/pre-diffusion duration & \multicolumn{2}{|c|}{$13950$ kt = $450$ s} & \multicolumn{3}{|c|}{$6200$ kt = $200$ s} & \multicolumn{2}{|c|}{$54250$ kt = $29$ min} & \multicolumn{2}{|c|}{$145$ s} \\
\cline{2-11}
 & Iteration diffusion/pre-diffusion nr. of steps & \multicolumn{2}{|c|}{$450$} & \multicolumn{3}{|c|}{$200$} & \multicolumn{2}{|c|}{$218$} & \multicolumn{2}{|c|}{$70$} \\
\cline{2-11}
 & Iteration diffusion/pre-diffusion step duration & \multicolumn{5}{|c|}{$31$ kt = $1$ s} & \multicolumn{2}{|c|}{$248$ kt = $8$ s} & \multicolumn{2}{|c|}{$2.5$ s} \\
\cline{2-11}
 & Iteration diffusion burn-in & \multicolumn{2}{|c|}{$5000$ kt $\approx 161$ s} & \multicolumn{3}{|c|}{$620$ kt = $20$ s} & \multicolumn{2}{|c|}{$10000$ kt $\approx 323$ s} & \multicolumn{2}{|c|}{$50$ s} \\
\cline{2-11}
 & Iteration collective avg. duration & \multicolumn{5}{|c|}{$620$ kt = $20$ s} & \multicolumn{2}{|c|}{$15500$ kt = $500$ s} & \multicolumn{2}{|c|}{$5$ s}  \\
\cline{2-11}
 & Iteration collective avg. nr. of steps & \multicolumn{5}{|c|}{$20$} & \multicolumn{2}{|c|}{$62$} & \multicolumn{2}{|c|}{$10$} \\
\cline{2-11}
 & Iteration collective avg. step dur. & \multicolumn{5}{|c|}{$31$ kt = $1$ s} & \multicolumn{2}{|c|}{$248$ kt = $8$ s} & \multicolumn{2}{|c|}{$1.25$ s} \\

\hline
\hline
\parbox[t]{2mm}{\multirow{4}{*}{\rotatebox[origin=c]{90}{Dispersion}}}
 & Step duration & \multicolumn{3}{|c|}{$15$ kt $\approx 0.48$s} & \multicolumn{6}{|c|}{-} \\
\cline{2-11}
 & Tumble duration & \multicolumn{3}{|c|}{$93 + \mathcal{N}(0,1) * 31$ kt $= 3 + \mathcal{N}(0,1)$ s} & \multicolumn{6}{|c|}{-} \\
\cline{2-11}
 & Tumble duration domain & \multicolumn{3}{|c|}{$0 - 124$ kt = $0 - 4$s} & \multicolumn{6}{|c|}{-} \\
\cline{2-11}
 & Run duration & \multicolumn{3}{|c|}{$19$ kt = $0.6$s} & \multicolumn{6}{|c|}{-} \\

\hline
\end{tabular}
}
\caption{List of parameter values used with all cases. One kilotick (kt) corresponds to $1/31 \approx 0.03$ seconds.}
\label{tab:params}
\end{center}
\end{table*}

\subsection{Kilobot simulations: arenas with surface $S = 500000 \texttt{mm}^2$}\label{sec:bigarenas}

Here, we consider in simulation the cases \textbf{sim-large-7} (with 7 arenas) and \textbf{sim-large-2} (with only 2 arenas: disk and annulus). Both cases use simulated Kilobots and have arenas with a surface of $S = 500000 \texttt{mm}^2$.
Main Text Figs.~4 and 5 present results from the \textbf{sim-large-7} case, after 30 iterations. That number of iterations was selected because, for all parameters considered, the algorithm needs at most 30 iterations to converge when considering arenas of that size.

\begin{figure*}[h]
\begin{center}
\begin{tikzpicture}
\node[anchor=north] (acc1it) at (0,0) {
    \includegraphics[width=1.0\textwidth]{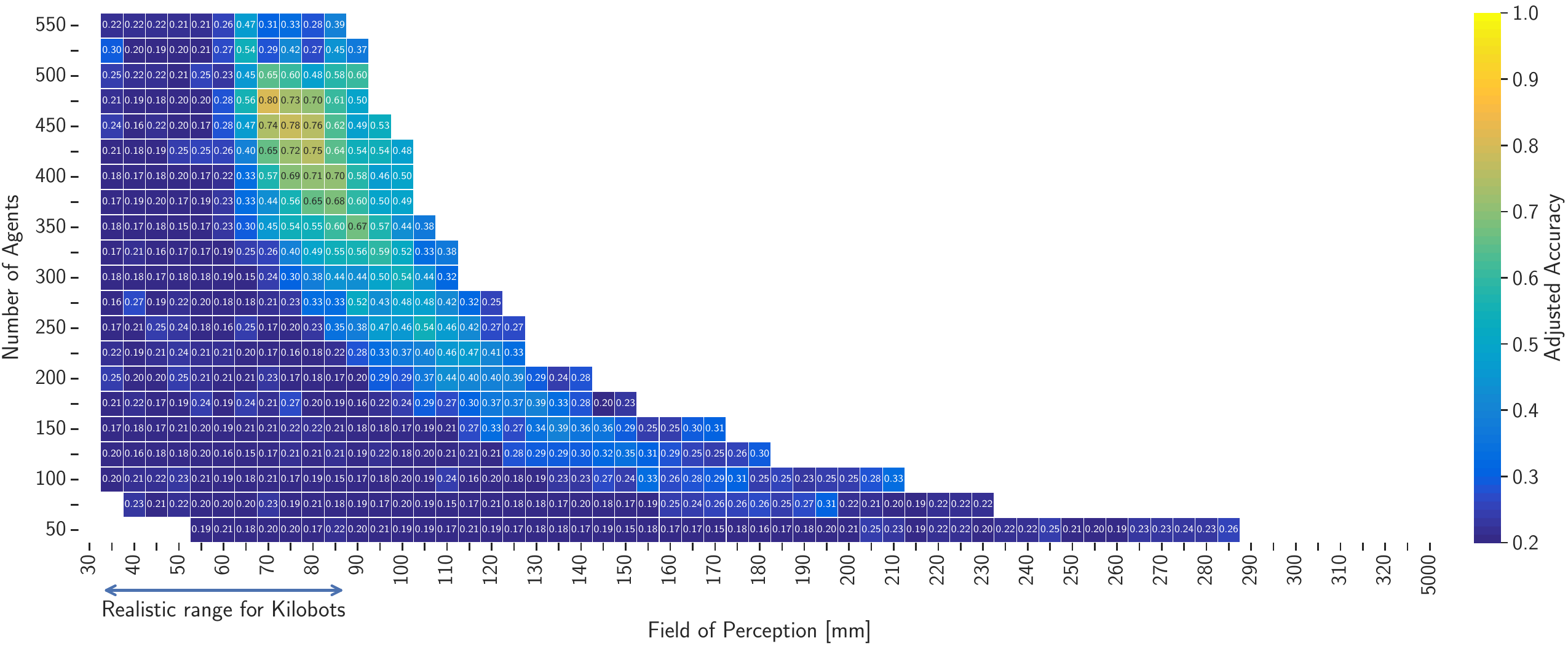}
};

\node[anchor=south] at (2,-1.8) {\large\textbf{sim-large-7}, only 1 iteration};

\node[below=0mm of acc1it] (acc30it) {
    \includegraphics[width=1.0\textwidth]{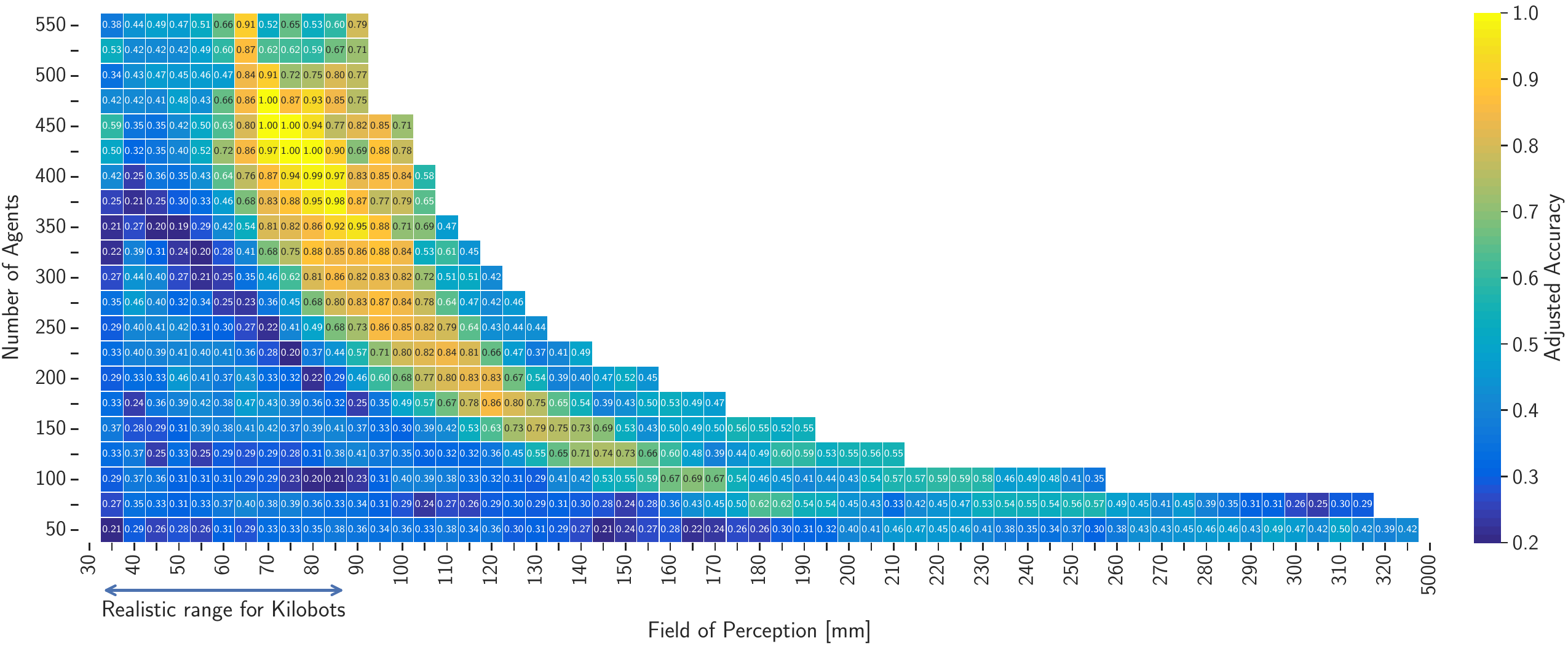}
};

\node[anchor=south] at (2,-10) {\large\textbf{sim-large-7}, 30 iterations};
\end{tikzpicture}
\caption{Evolution of the accuracy with respect to the field of perception of each agent and the number of agents, for the case \textbf{sim-large-7}. Here, we consider 7 arenas with a surface $S = 500000 mm^2$. All results are computed over 64 simulation runs. We adjust the accuracy score to only show results where the simulated diffusion sessions in the swarm did not diverge: empty bins correspond to results where there are more than $50\%$ of runs exhibiting divergences. \textbf{Top:} results obtained after only 1 iteration. \textbf{Bottom:} results obtained after 30 iterations.}
\label{fig:regimes_simlarge7}
\end{center}
\end{figure*}

\begin{figure*}[h]
\begin{center}
\includegraphics[width=0.8\textwidth]{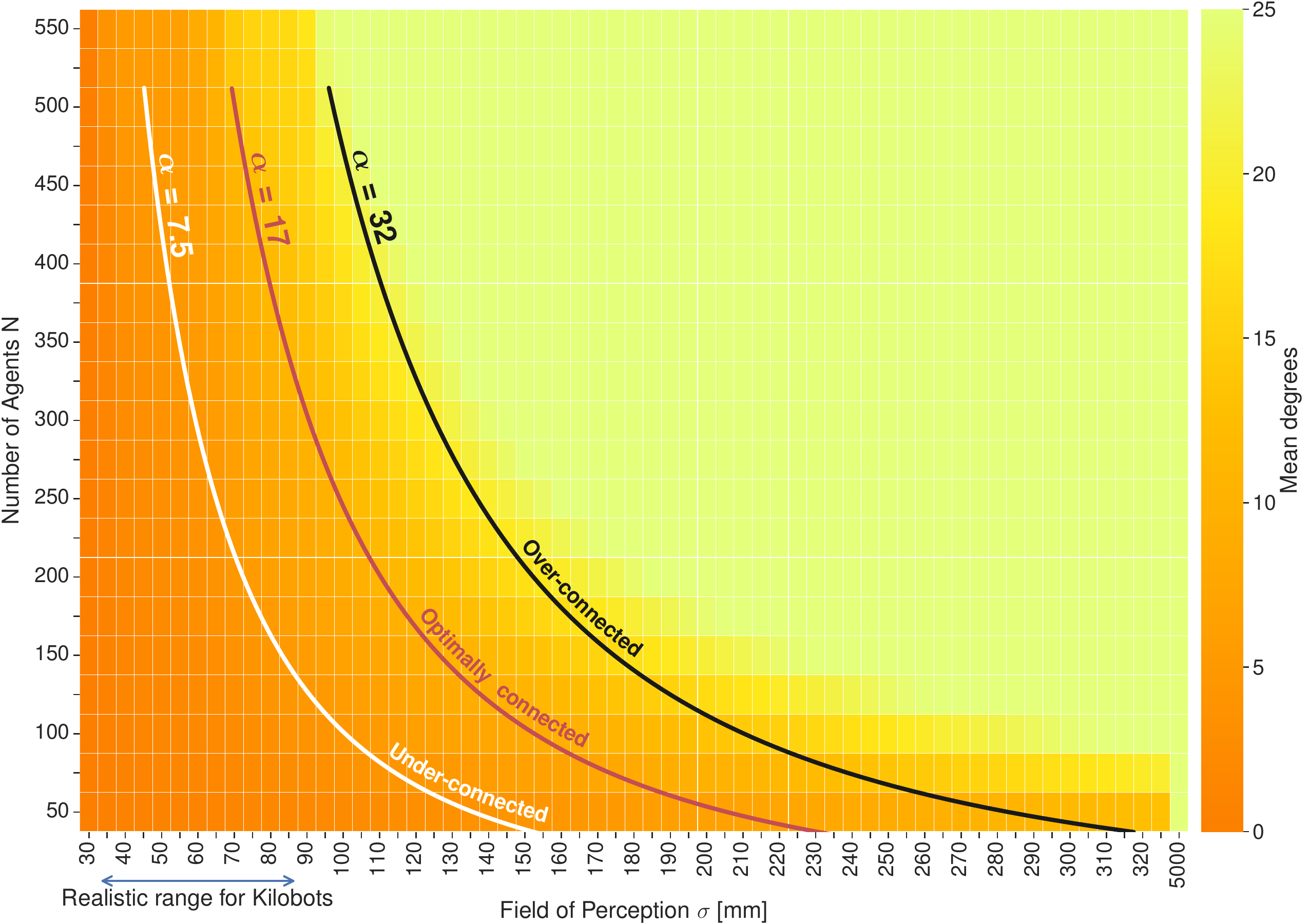}

\includegraphics[width=0.8\textwidth]{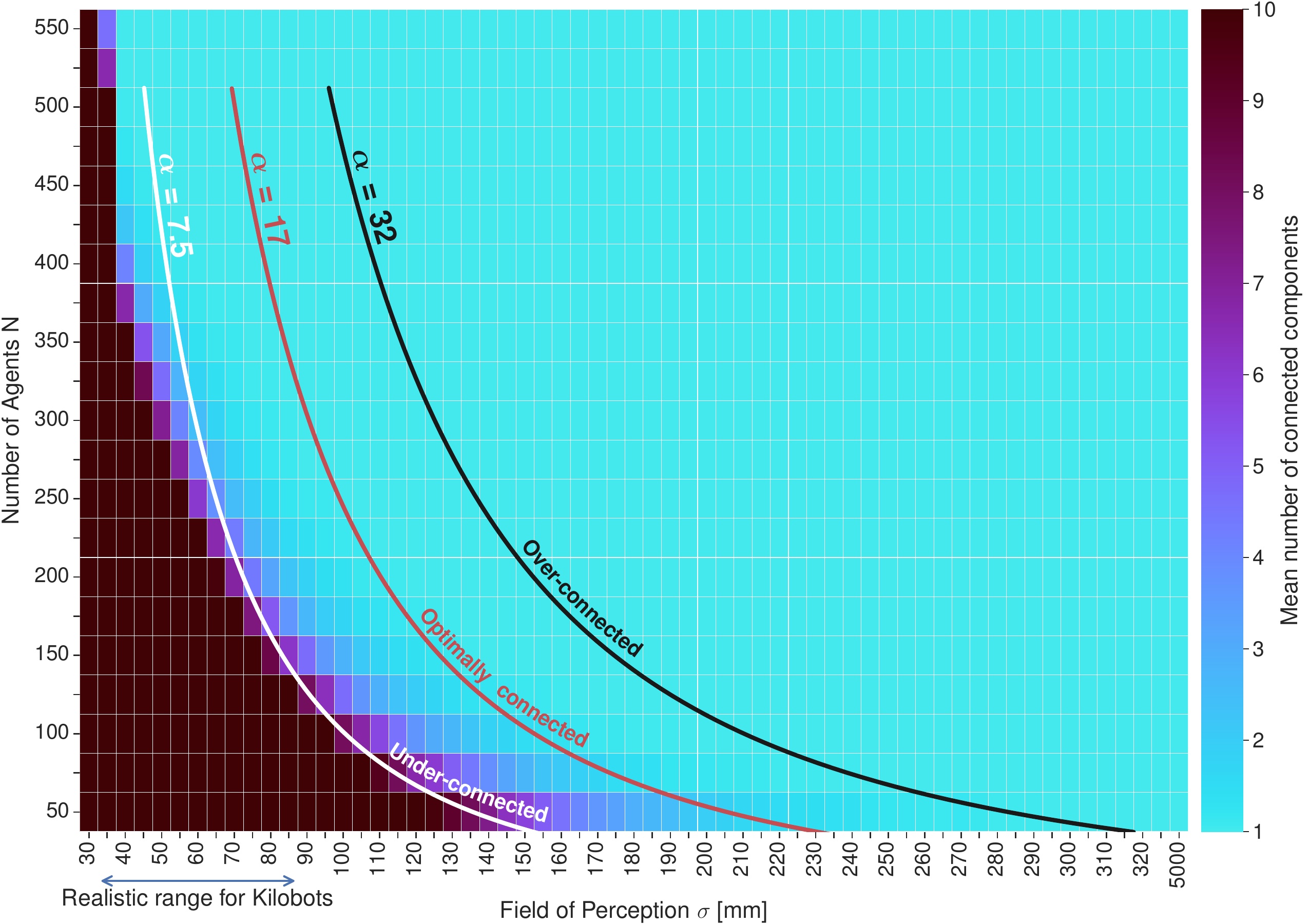}
\end{center}
\caption{\textbf{Top:} Mean number of neighbors of agents (\ie mean degree of the communication graphs), and \textbf{Bottom}: mean number of connected graph components in the \textbf{sim-large-7} case, after 30 iterations.}
\label{fig:mean_conn_comp}
\end{figure*}

\begin{figure*}[h]
\centering
\includegraphics[width=1.0\textwidth]{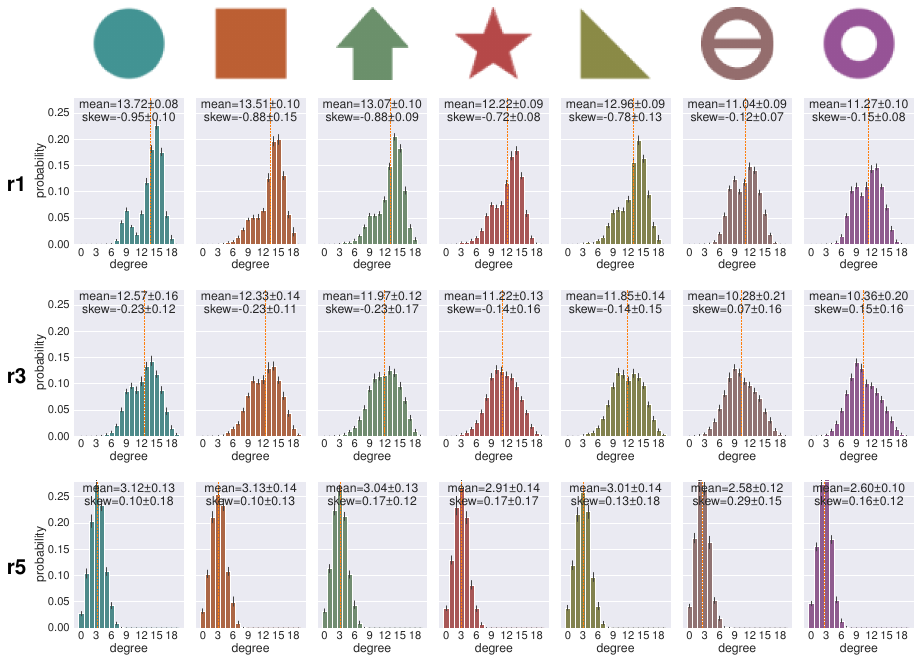}
\caption{Distribution of degrees of the regimes \textbf{r1, r3, r5}, as defined in Fig.~5 of the Main Text, and for the \textbf{sim-large-7} case, after 30 iterations. Results are obtained over 32 runs, with the variability in values represented by the error bars in each histogram bin. In each plot, an orange dotted line corresponds to the distribution mean. Values of mean and skewness are provided for each plot. Regimes \textbf{r1} and \textbf{r3} both show different values mean and skewness for each arena. The skewness of the distribution is affected by the prevalence of borders in the arenas: agents close to a wall will have less neighbors than agents far for them. In the case of the disk, the disparity in degree is particularly high, with agents close to walls having far less neighbors than agents in the center. Conversely, the annulus has far less disparity in degree, as most agents will be close either to the outer wall or the inner wall. As such, there are no real center. The other arenas degree distributions fall between the two extremes cases of the disk and annulus.}
\label{fig:distrib_degrees}
\end{figure*}

Supplementary Fig.~\ref{fig:regimes_simlarge7} compares the results with respect to accuracy after respectively 1 and 30 iterations. Both configurations show a clear impact of the simulation parameters ($N$ and $\sigma$) on the accuracy. These results validate the conclusions from Main Text Fig.~5: a higher number of iterations translate into a higher accuracy. For both 1 and 30 iterations, we observe similar regimes as in Main Text Fig.~5, with the optimal solutions found in the hyperbole $N = \alpha S / (\pi \sigma^2)$ with $\alpha = 17$.

\FloatBarrier\clearpage

\subsection{Kilobot simulations: influence of the dispersion algorithm}\label{sec:dispersionRes}

We want to study how the dispersion algorithm and the initial position of the agents influence the behavior of the spectral swarm robotics algorithm and associated results in terms of shape classification accuracy. We consider the following three cases:
\begin{description}
    \item[\textbf{sim-small-2-dispersion}] where agents are initially packed at the center of the arena and move according to a run-and-tumble motion during both the ``seeding" and the ``short random walk" stages of the spectral swarm robotics algorithm. As such, agents first disperse throughout the arena and then move at the beginning of each iteration of the algorithm to change their neighbors.
    \item[\textbf{sim-small-2-random}] where agents are initially placed at random positions of the arena and where they remain static: the ``seeding" stage of the spectral swarm robotics algorithm is skipped and agent do not move during the ``short random walk" stage. As such, they always keep the same neighbors for all iterations.
    \item[\textbf{sim-small-2-uniform}] where agents are initially optimally distributed throughout the arena so that the distance between each pair of neighboring agents is the same across the entire swarm. This ensures that all parts of the arena are identically discretized by the communication graph of the agents. The optimal positions of the agents are found by using K-Means to find $N$ clusters from a large ($100000$) number of points randomly placed in a space with the boundaries as the arenas.
\end{description}
All three cases are performed in simulation, with two arenas (disk and annulus) of surface $S = 70000\texttt{mm}^2$ -- in turn, we will use a smaller number of agents compared to the cases of the previous section (Supplementary Table~\ref{tab:cases}).

While cases \textbf{sim-small-2-dispersion} and \textbf{sim-small-2-random} do not assume any prior knowledge about the geometry of the arenas, the \textbf{sim-small-2-uniform} case assumes that it is possible to distribute uniformly the agent in the arenas before the start of the algorithm. In experiments with Kilobots, this can be achieved by having the human experimenter positioning the robots so that they are evenly distributed before starting the experiment.

\begin{figure*}[h]
\begin{center}
\vspace{1cm}
\includegraphics[width=1.00\textwidth]{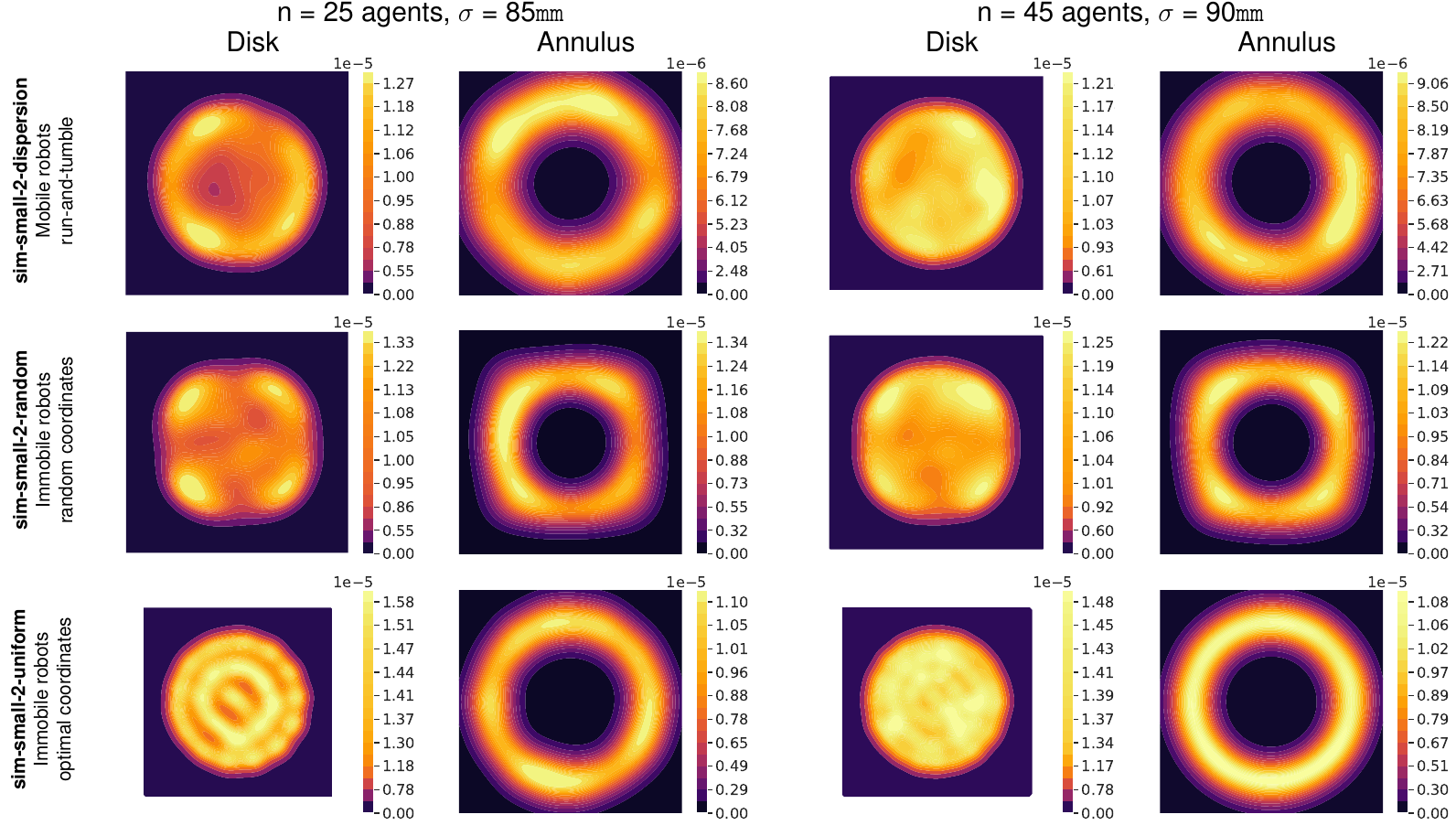}%
\caption{Density of agents in the arenas after the initial seeding, for cases with arena surface $S = 70000\texttt{mm}^2$ and for two different configurations (either 25 or 45 agents). Results were obtained over 64 runs.}
\label{fig:dispersionDensities}
\end{center}
\end{figure*}

Supplementary Figure~\ref{fig:dispersionDensities} shows the densities of agents after the seeding stage over 64 runs. In all cases, for both arenas, the agents reach all parts of the arenas, with no visible gap.

\begin{figure*}[h!]
\begin{center}
\vspace{0.3cm}
\includegraphics[width=0.99\textwidth]{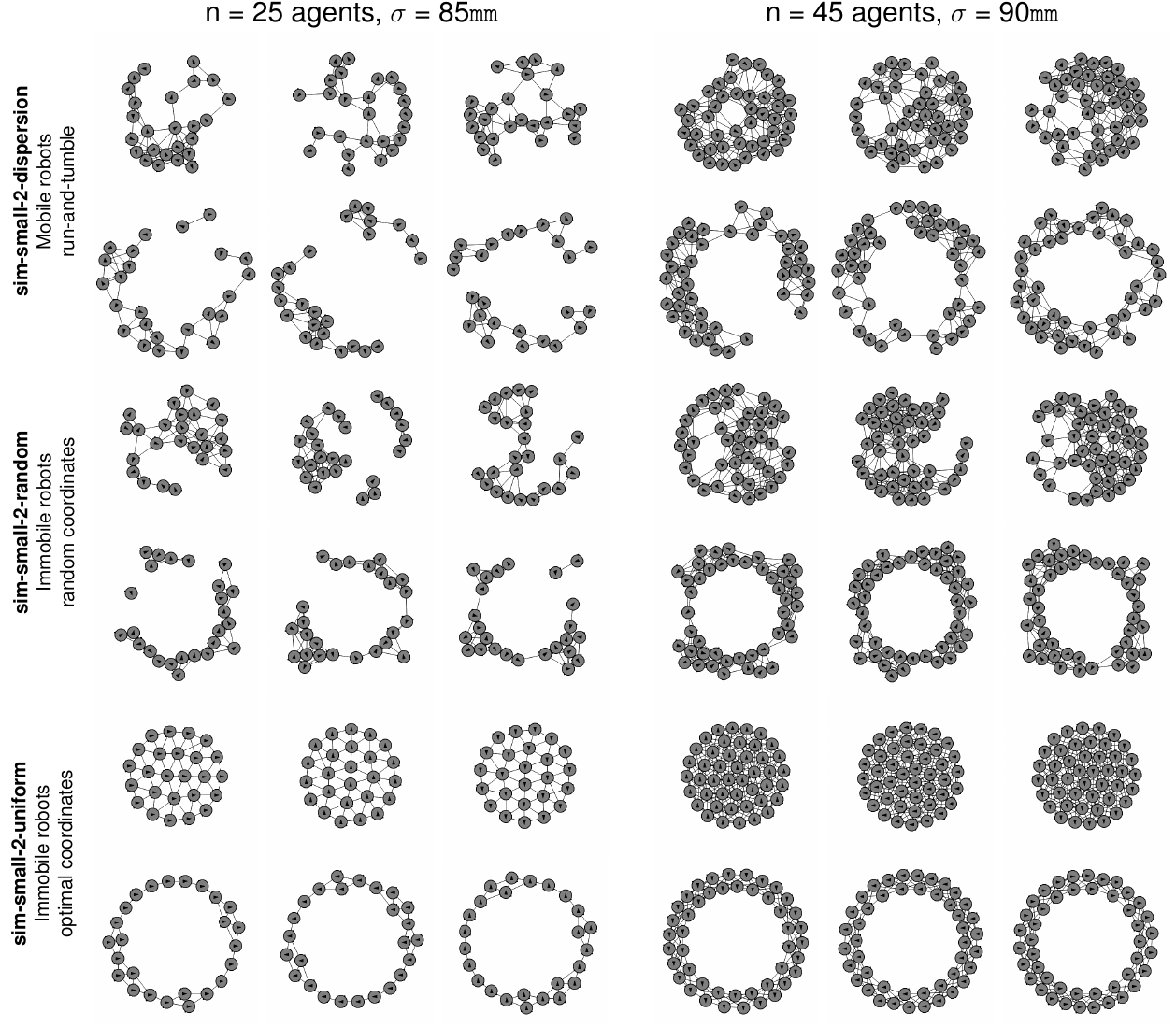}%
\caption{Examples of position of the agents after the initial seeding, for cases with arena surface $S = 70000\texttt{mm}^2$ and for two different configurations (either 25 or 45 agents).}
\label{fig:dispersionTraces}
\end{center}
\end{figure*}

Supplementary Figure~\ref{fig:dispersionTraces} shows representative examples of agents position after the ``initial seeding" stage for the three cases. 

\begin{figure*}[h!]
\begin{center}
\includegraphics[width=1.00\textwidth]{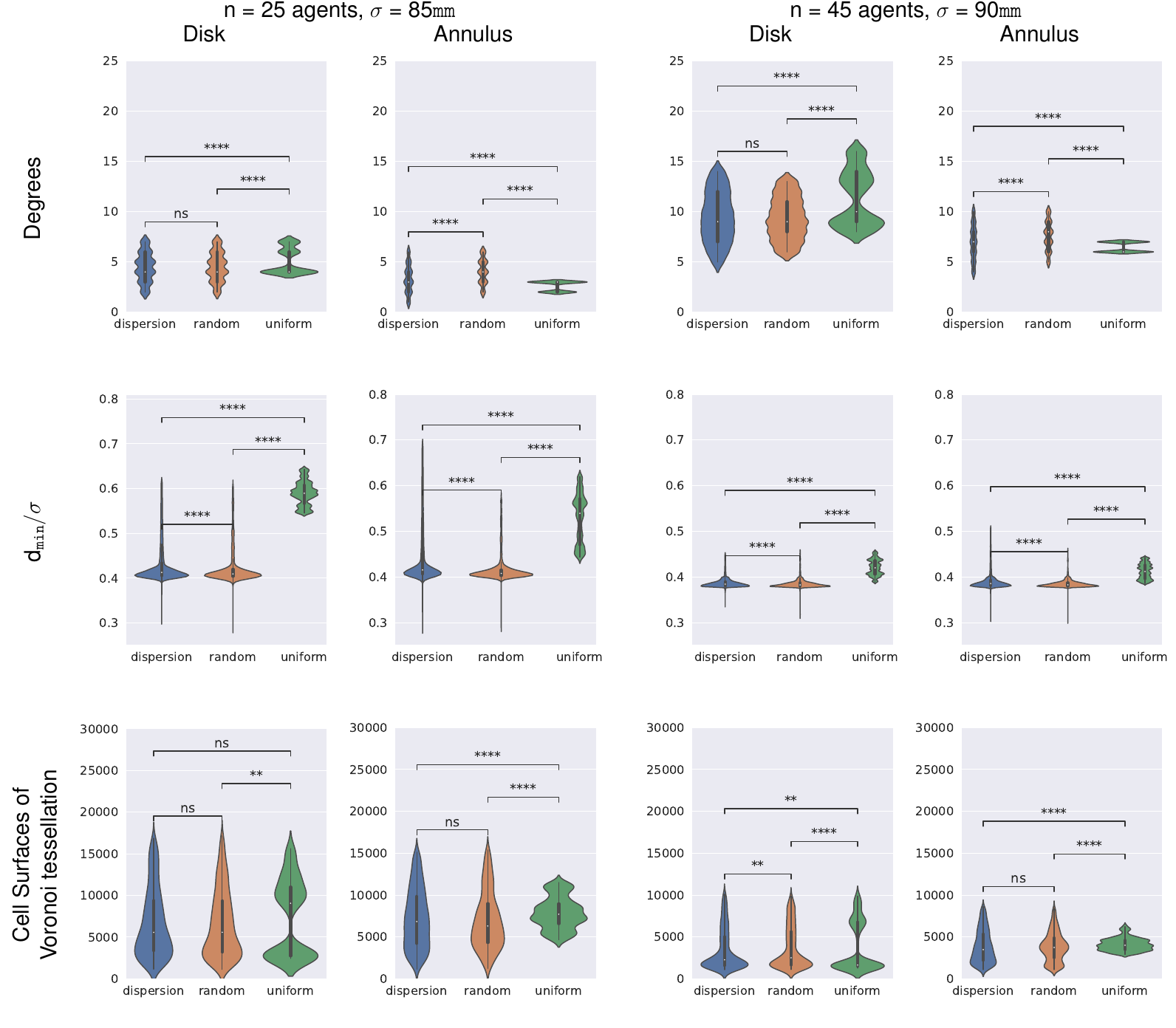}%
\caption{Statistical analysis of the different types of agent dispersion in arenas of surface $S = 70000\texttt{mm}^2$. We consider the \textbf{sim-small-2-dispersion}, \textbf{sim-small-2-random}, and \textbf{sim-small-2-uniform} cases -- named in the figure ``dispersion", ``random" and ``uniform" respectively. Two configurations are considered, using 25 and 45 agents respectively. Statistical significance is computed using the two-sided Mann-Whitney-Wilcoxon test with Bonferroni correction. Significant pairs are designated as follows: ns: $0.05 < p \leq 1.00$, *: $0.01 < p \leq 0.05$, **: $0.001 < p \leq 0.01$, ***: $0.0001 < p \leq 0.001$, ****: $p \leq 0.0001$.}
\label{fig:dispersionStats}
\end{center}
\end{figure*}

Supplementary Figure~\ref{fig:dispersionStats} characterizes the spacial distribution of agents after the seeding stage for the three cases considered here. We consider the following three complementary statistics and assess their distribution using violin plots:
\begin{description}
    \item[Degrees] The distribution of degrees in the communication graphs. This metric only takes into account the communication graph $\mathcal{G}$ itself, without considering the distance between agents.
    \item[$d_{\texttt{min}}/\sigma$] The distance between the focal agent and its closest neighbors divided by $\sigma$, the field of perception of the agents. This metric only takes into account the distance to one neighbor, without involving the communication graph $\mathcal{G}$.
    \item[Cell Surface of Voronoi tessellation] The area of each cells obtained by computing a Voronoi tessellation of the positions of the agents in the arena (constrained by the geometry of the arena). This metric only takes into account the spatial size of neighborhoods, without involving the communication graph $\mathcal{G}$. The Voronoi tessellation is computed in our Python scripts using the Geopandas and shapely packages~\cite{kelsey_jordahl_2020_3946761,shapely2007}.
\end{description}
In both configurations, the results of cases \textbf{sim-small-2-dispersion} and \textbf{sim-small-2-random} are generally similar for the degrees and cell surface metrics and differ from the results of case \textbf{sim-small-2-uniform}. The results of the $d_{\texttt{min}}/\sigma$ metric suggest that agents tend to be more packed together in the \textbf{sim-small-2-dispersion} and \textbf{sim-small-2-random} cases compared to the \textbf{sim-small-2-uniform} -- this matches the examples of Supplementary Fig.~\ref{fig:dispersionTraces}. The cell surfaces in the uniform case has generally less variability compared to the other cases (at least for the annulus): indeed all agents are equidistant to their neighbors in this case. In turn, the discretization of the shape is more uniform in the \textbf{sim-small-2-uniform} case.

Note however, when the number of agents is sufficiently low, it is very possible to obtain a graph of agents with multiple graph components (cf Supplementary Fig.~\ref{fig:dispersionTraces}, \ie a graph that is not fully connected. This will have a different set of consequences depending on the case:
\begin{description}
    \item[\textbf{sim-small-2-dispersion}] The agents move and change their neighbors every iteration. Consequently, the graph components do not stay static over iterations: they may sometimes merge together or split; some agents may go from one component to another. In turn, the estimations of $\lambda_2$ will be periodically transferred from component to component, making the system able to correctly (but imprecisely) estimate $\lambda_2$ while sometimes having more than one graph component. That means that it is still possible to have a classification with high-performing accuracy even if there are more than one component. Of course, if the number of components is too high (\ie the number of agents is too low), such computation will be impossible: the diffusion sessions will diverge, and/or the information from one component to another won't be transferred, resulting in an inaccurate classification.
    \item[\textbf{sim-small-2-random}] The agents are immobile and always keep the same neighbors. Therefore, isolated agents will always remain isolated and graph components will not be able to communicate with each other over multiple iterations. This will results in each component doing its own independent estimation of $\lambda_2$. As this estimation assumes that the agents are distributed across the entire arena, this process will almost always results in an incorrect (rather than imprecise) estimation, only taking into account the spaces covered by each component.
    \item[\textbf{sim-small-2-uniform}] The agents are immobile and equidistributed in the arena. As such, if there are several components, it follows that there are actually $N$ components, that is, the agents are all too distant from each other to communicate. In this case, all agents are isolated and the diffusion stages cannot be computed, resulting in undecided classification results.
\end{description}

\begin{figure*}[h]
\begin{center}
\begin{tikzpicture}
\node[anchor=north] (accRTvsRandom) at (0,0) {
    \includegraphics[width=0.9\textwidth]{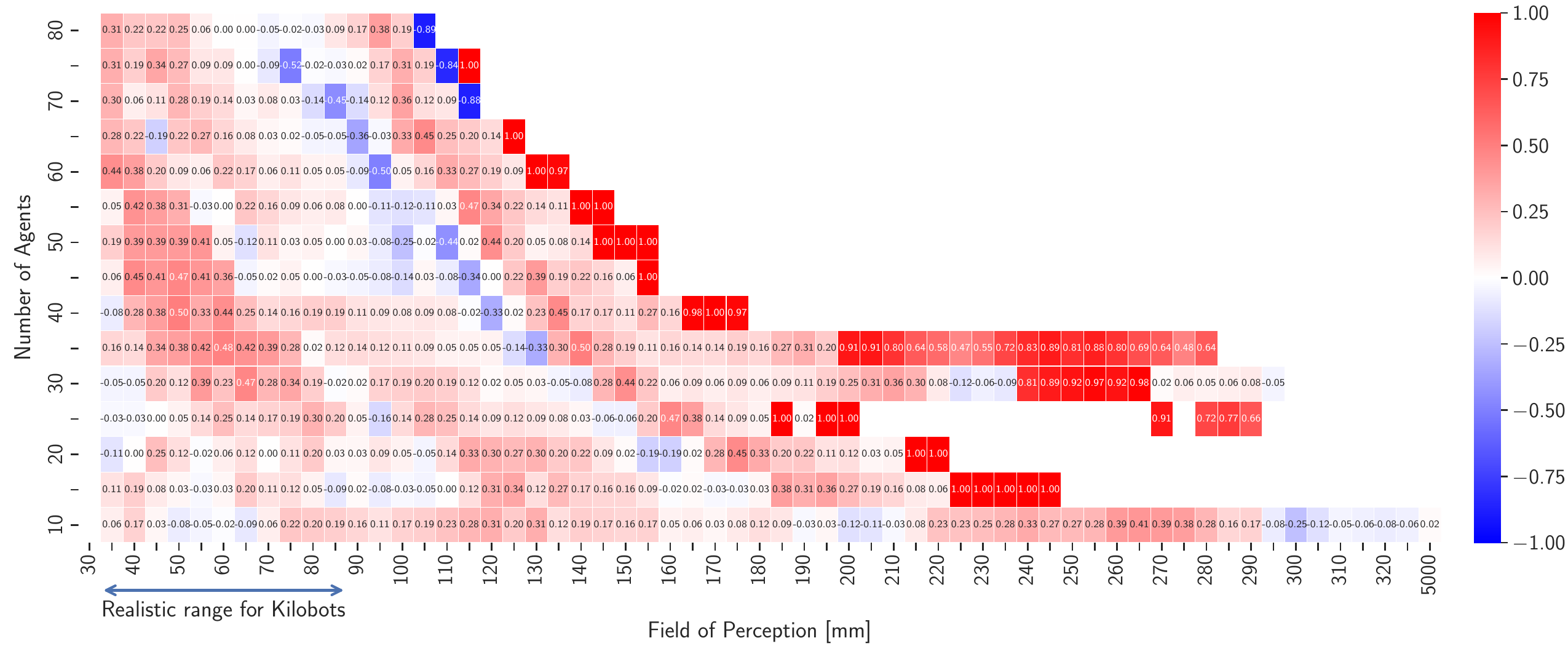}
};

\node[anchor=south] at (2.4,-1.8) {\large\textbf{sim-small-2-dispersion} - \textbf{sim-small-2-random}};
\node[anchor=south] at (2.4,-2.3) {\large 10 iterations};

\node[below=0mm of accRTvsRandom] (accRTvsUniform) {
    \includegraphics[width=0.9\textwidth]{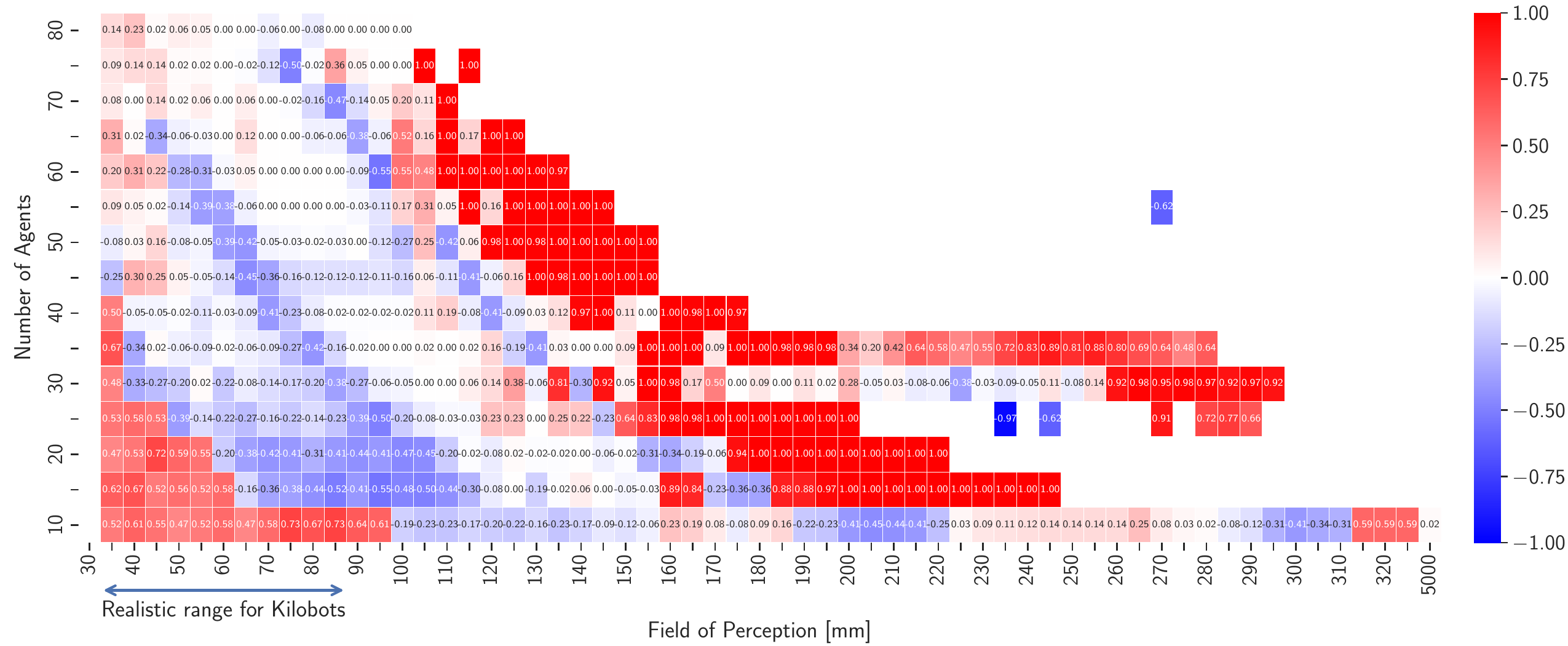}
};

\node[anchor=south] at (2.4,-9.5) {\large\textbf{sim-small-2-dispersion} - \textbf{sim-small-2-uniform}};
\node[anchor=south] at (2.4,-10.0) {\large 10 iterations};

\end{tikzpicture}
\caption{Difference in accuracy scores between cases \textbf{sim-small-2-dispersion}, \textbf{sim-small-2-random} and \textbf{sim-small-2-uniform}, depending on the number of agents and their field of perception. All results are obtained in simulations over 10 iterations and over 64 runs. Each bin of the 2D histograms corresponds to the difference between the accuracy score of \textbf{sim-small-2-dispersion} and that of either \textbf{sim-small-2-random} (top) or \textbf{sim-small-2-uniform} (bottom). We adjust the accuracy score to only show results where the simulated diffusion sessions in the swarm do not diverge: empty bins correspond to results where there are more than $50\%$ of the runs exhibiting divergences.}
\label{fig:diffRegimes}
\end{center}
\end{figure*}

These findings are corroborated by Supplementary Fig.~\ref{fig:diffRegimes}, showing the difference in accuracy scores between the three cases after 10 iterations (the number of 10 iterations was selected so that $\texttt{Final}\lambda_2$ estimations have converged for all settings). The case \textbf{sim-small-2-dispersion} has generally higher (or at least equivalent) accuracy than the \textbf{sim-small-2-random} case: as explained before, it may be attributed to the capacity of the former to have agents that change neighbors across iterations. In particular, that is the case when $\sigma$ is small ($\leq 85 \texttt{mm}$, \ie the realistic range for Kilobots), placing the system below the optimal $\alpha = 17$ hyperbole or to the right of this hyperbole: in these settings, the dispersion case having far higher scores than the random case. 

Both the \textbf{sim-small-2-dispersion} and \textbf{sim-small-2-uniform} cases have similar ($100\%$) accuracy scores close to the optimal $\alpha = 17$ hyperbole (1). The latter has better scores than the former immediately below $p$ (2), while the dispersion case outperforms the uniform case above and to the right of $p$ (3), and with settings with low values of $N$ ($\leq 40$) and $\sigma$ ($\leq 85 \texttt{mm}$) (4). The good results of the uniform case in (1) and (2) can be explained by considering an uniform distribution of agents in the arena: it both ensures that the graph is connected (except for settings where the agents are too far from each other) and that the discretization of the shape is uniform (each point occupies a Voronoi cell that all have similar size).
However, the agents in the uniform case do not move and always keep the same neighbors at each iteration.
while this lowers the variability in $\texttt{Final}\lambda_2$ estimations, it also makes the estimation less resilient to graph artifacts that could bias such estimations (\eg a small number of agents diverging and affecting their entire neighborhood).
This may also be the case in the (3) settings where the uniform case may produce graphs that are too dense, increasing the probability of divergences and computation anomalies during the diffusion stages. This issue is alleviated in the dispersion case by changing neighbors at each iteration.
In the (4) settings, the uniform distribution results in agents being too far apart to communicate, whereas the dispersion case still produces clusters of agents sufficiently close to each other to estimate $\texttt{Final}\lambda_2$, possibly relying only on local geometry, as the graph becomes too small to match the entire surface of the arenas with those settings.

\begin{figure*}[h]
\begin{center}
\resizebox{0.9\textwidth}{!}{
\begin{tikzpicture}
\node[anchor=north] (acc1it) at (0,0) {
    \includegraphics[width=0.9\textwidth]{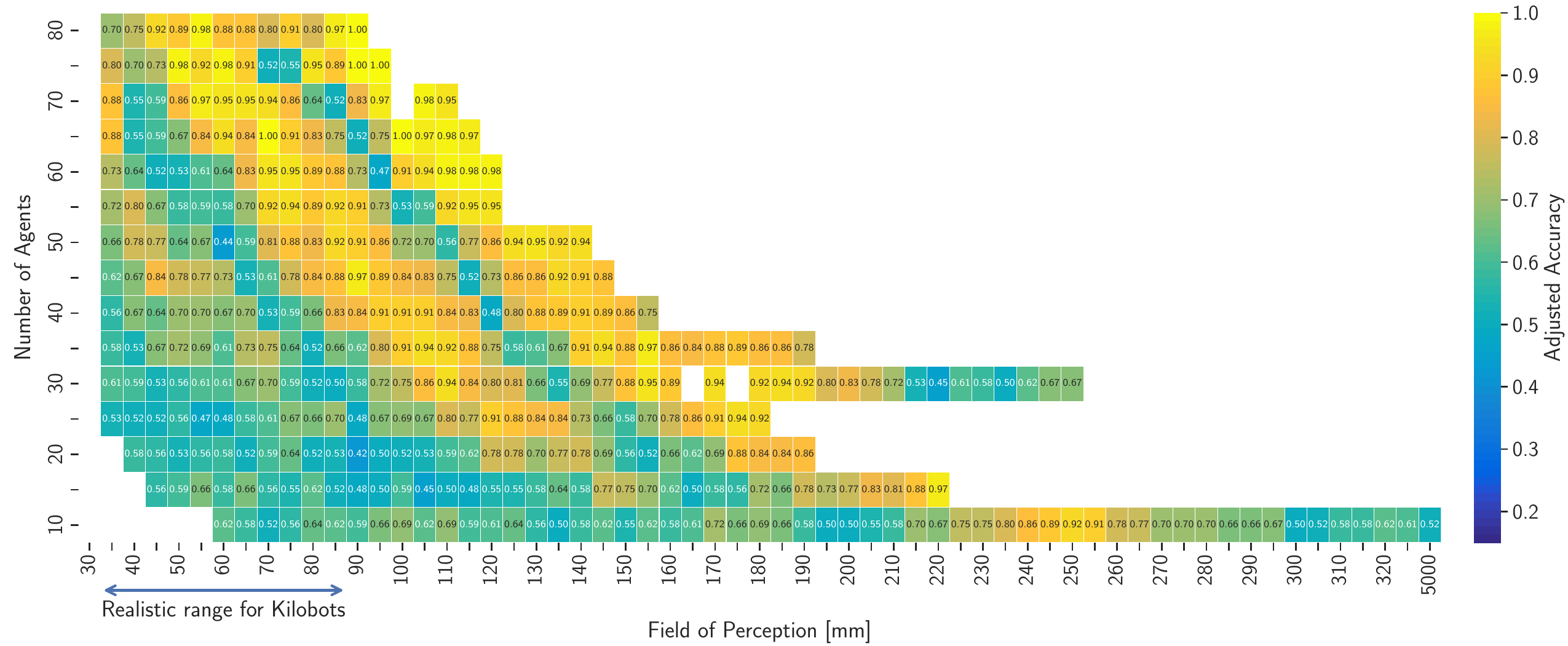}
};

\node[anchor=south] at (2,-1.8) {\large\textbf{sim-small-2-dispersion}, only 1 iteration};

\node[below=0mm of acc1it] (acc30it) {
    \includegraphics[width=0.9\textwidth]{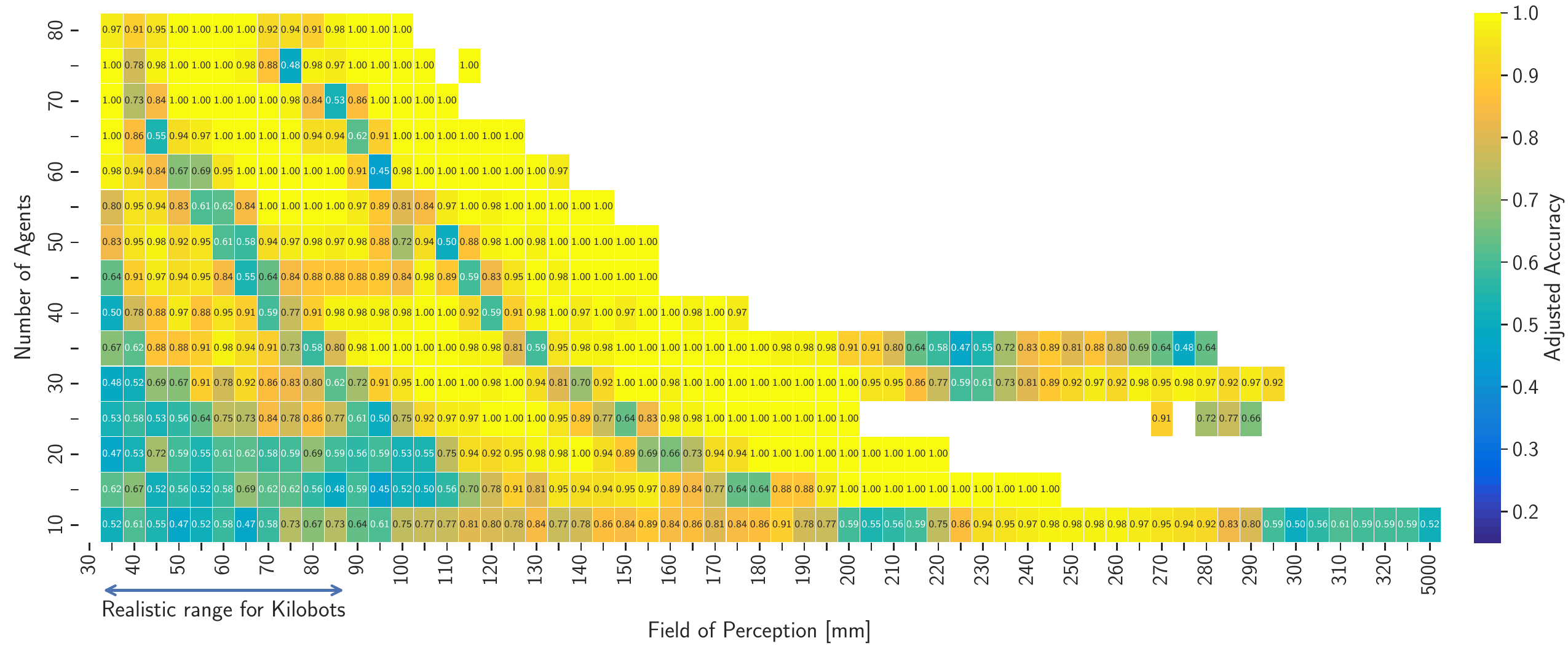}
};

\node[anchor=south] at (2,-9) {\large\textbf{sim-small-2-dispersion}, 10 iterations};

\node[below=0mm of acc30it] (rank_disk) {
    \includegraphics[width=0.9\textwidth]{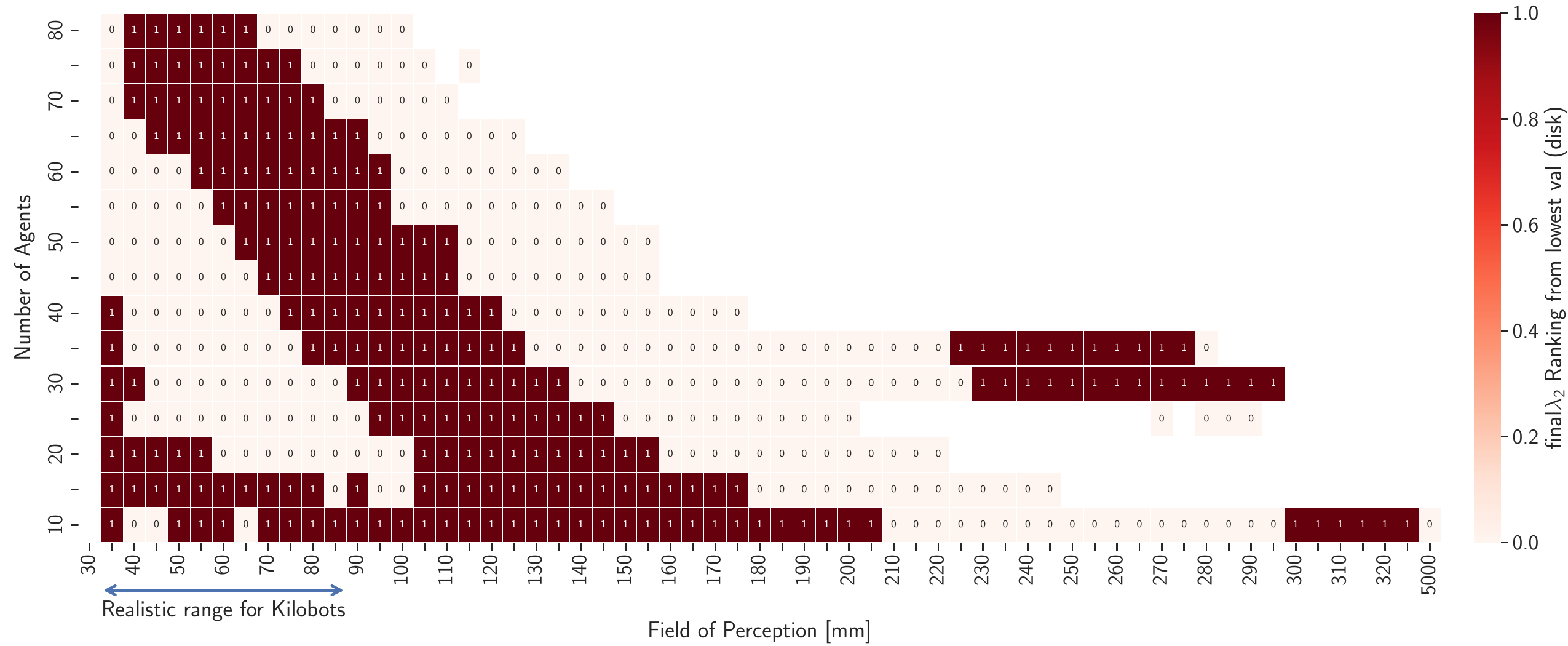}
};

\node[anchor=south] at (2,-16.5) {\large\textbf{sim-small-2-dispersion}, 10 iterations, Ranking disk};

\end{tikzpicture}
}
\caption{Evolution of the accuracy depending on the number of agents and their field of perception, for the case in simulation \textbf{sim-small-2-dispersion}. In this case, we consider 2 arenas (disk vs annulus) of surface $S = 70000 mm^2$ and the robots use the run-and-tumble dispersion algorithm to move throughout the arenas between each iteration. All results are computed over 64 runs of simulation. We adjust the accuracy score to only show results where the simulated diffusion sessions in the swarm do not diverge: empty bins correspond to results where there are more than $50\%$ of runs exhibiting divergences. \textbf{Top:} results with only 1 iteration. \textbf{Middle:} results obtained after 10 iterations. \textbf{Bottom:} Ranking of the order of values of the Final$\lambda_2$ centroids for the disk arena.}
\label{fig:regimes_simsmall2dispersion}
\end{center}
\end{figure*}

\begin{figure*}[h]
\begin{center}
\resizebox{0.9\textwidth}{!}{
\begin{tikzpicture}
\node[anchor=north] (acc1it) at (0,0) {
    \includegraphics[width=0.9\textwidth]{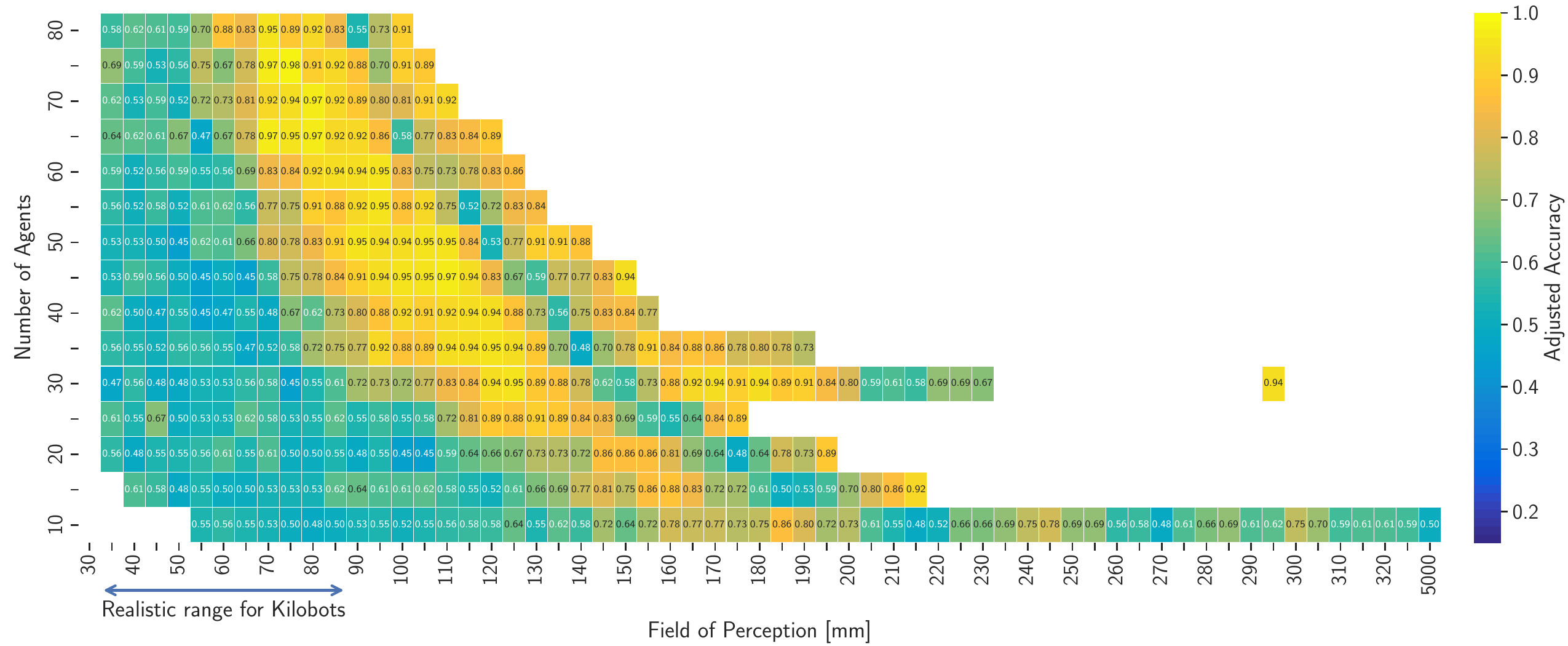}
};

\node[anchor=south] at (2,-1.8) {\large\textbf{sim-small-2-random}, only 1 iteration};

\node[below=0mm of acc1it] (acc30it) {
    \includegraphics[width=0.9\textwidth]{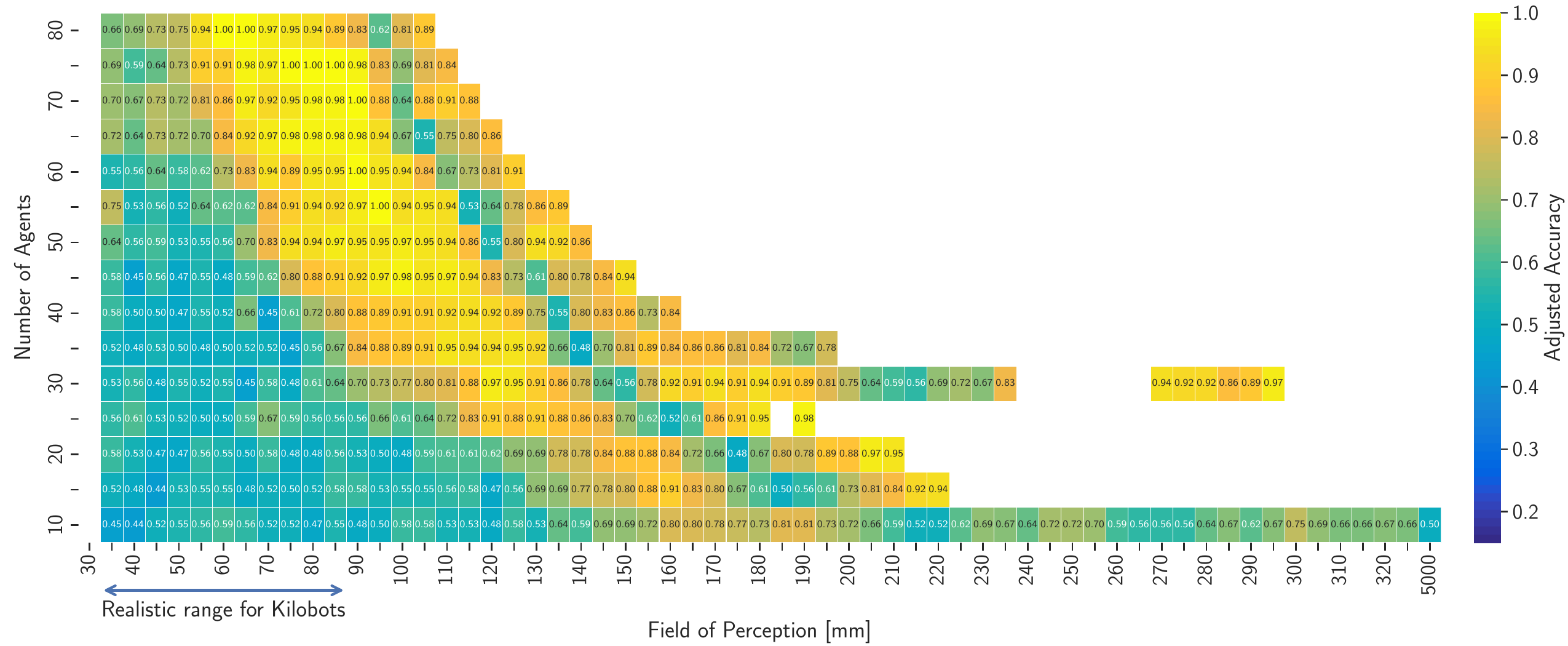}
};

\node[anchor=south] at (2,-9) {\large\textbf{sim-small-2-random}, 10 iterations};

\node[below=0mm of acc30it] (rank_disk) {
    \includegraphics[width=0.9\textwidth]{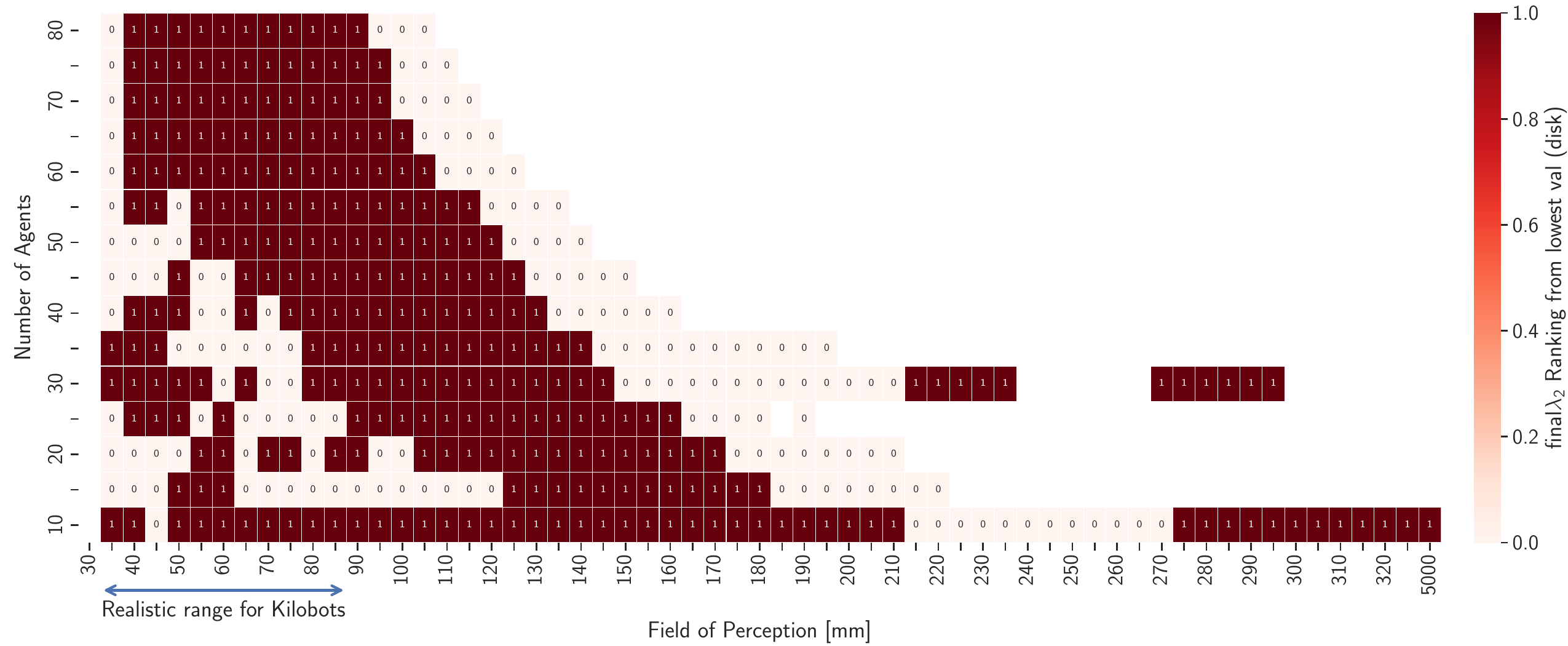}
};

\node[anchor=south] at (2,-16.5) {\large\textbf{sim-small-2-random}, 10 iterations, Ranking disk};
\end{tikzpicture}
}
\caption{Evolution of the accuracy depending on the number of agents and their field of perception, for the case in simulation \textbf{sim-small-2-random}. In this case, we consider 2 arenas (disk vs annulus) of surface $S = 70000 mm^2$ and the robots always stay immobile during the simulation. The initial position of the robots inside the arenas are random (2D uniform distribution). All results are computed over 64 runs of simulation. We adjust the accuracy score to only show results where the simulated diffusion sessions in the swarm do not diverge: empty bins correspond to results where there are more than $50\%$ of runs exhibiting divergences. \textbf{Top:} results after 1 iteration. \textbf{Middle:} results after 10 iterations. \textbf{Bottom:} Ranking of the values of the Final$\lambda_2$ centroids for the disk arena.}
\label{fig:regimes_simsmall2random}
\end{center}
\end{figure*}

\begin{figure*}[h]
\begin{center}
\resizebox{0.9\textwidth}{!}{
\begin{tikzpicture}
\node[anchor=north] (acc1it) at (0,0) {
    \includegraphics[width=0.9\textwidth]{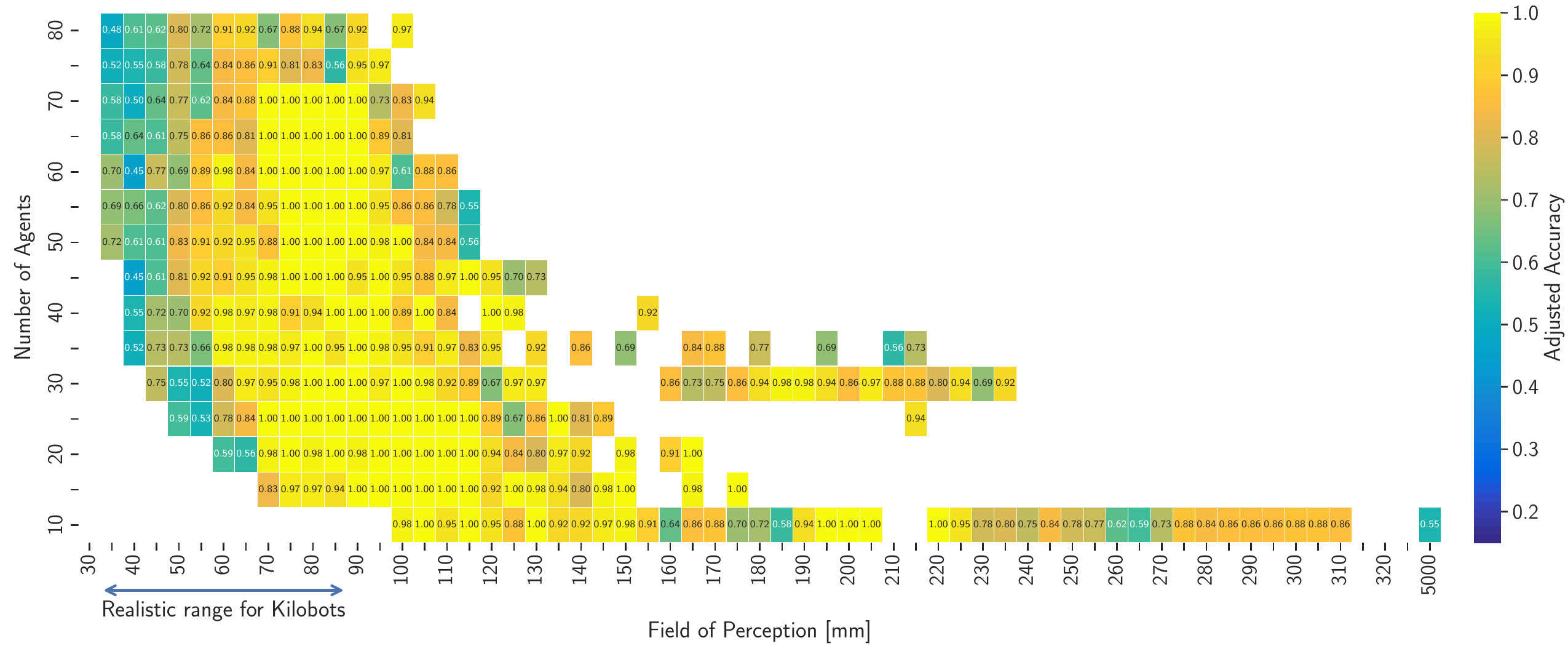}
};

\node[anchor=south] at (2,-1.8) {\large\textbf{sim-small-2-uniform}, only 1 iteration};

\node[below=0mm of acc1it] (acc30it) {
    \includegraphics[width=0.9\textwidth]{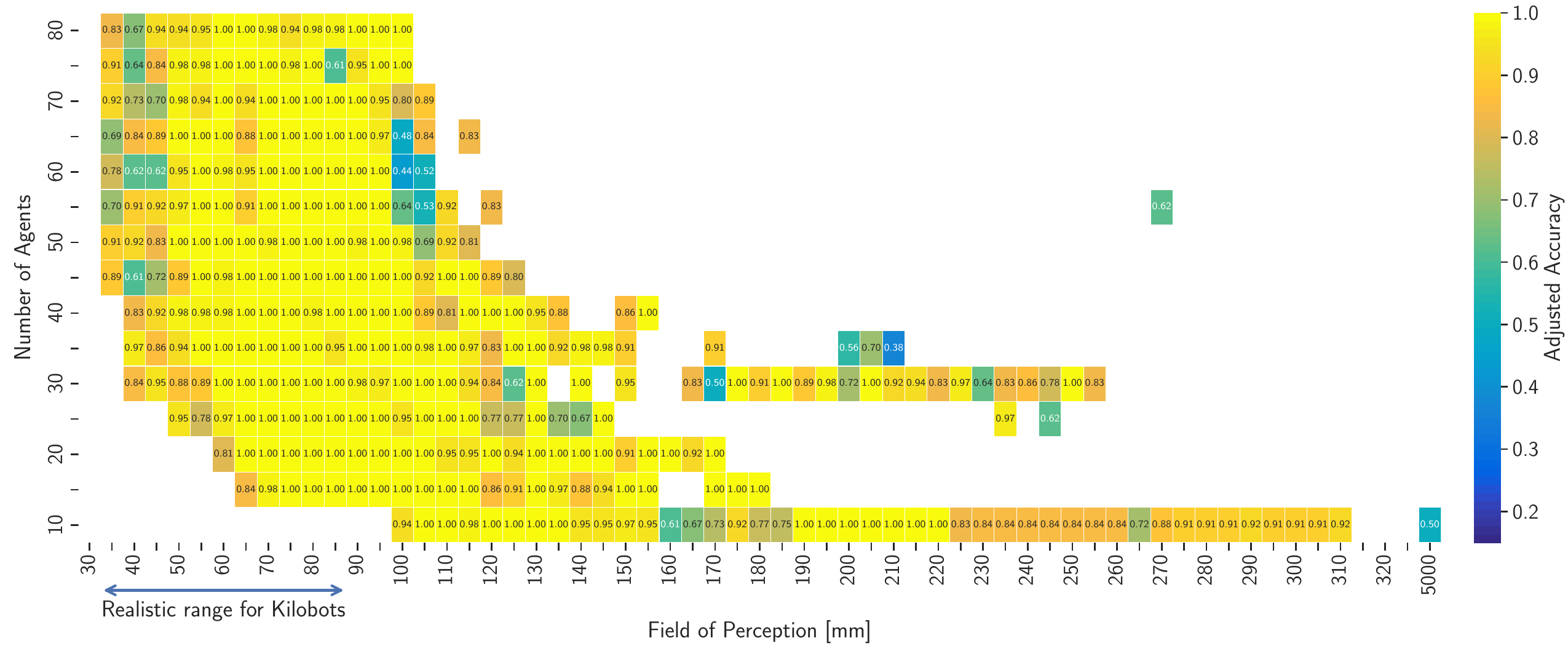}
};

\node[anchor=south] at (2,-9) {\large\textbf{sim-small-2-uniform}, 10 iterations};

\node[below=0mm of acc30it] (rank_disk) {
    \includegraphics[width=0.9\textwidth]{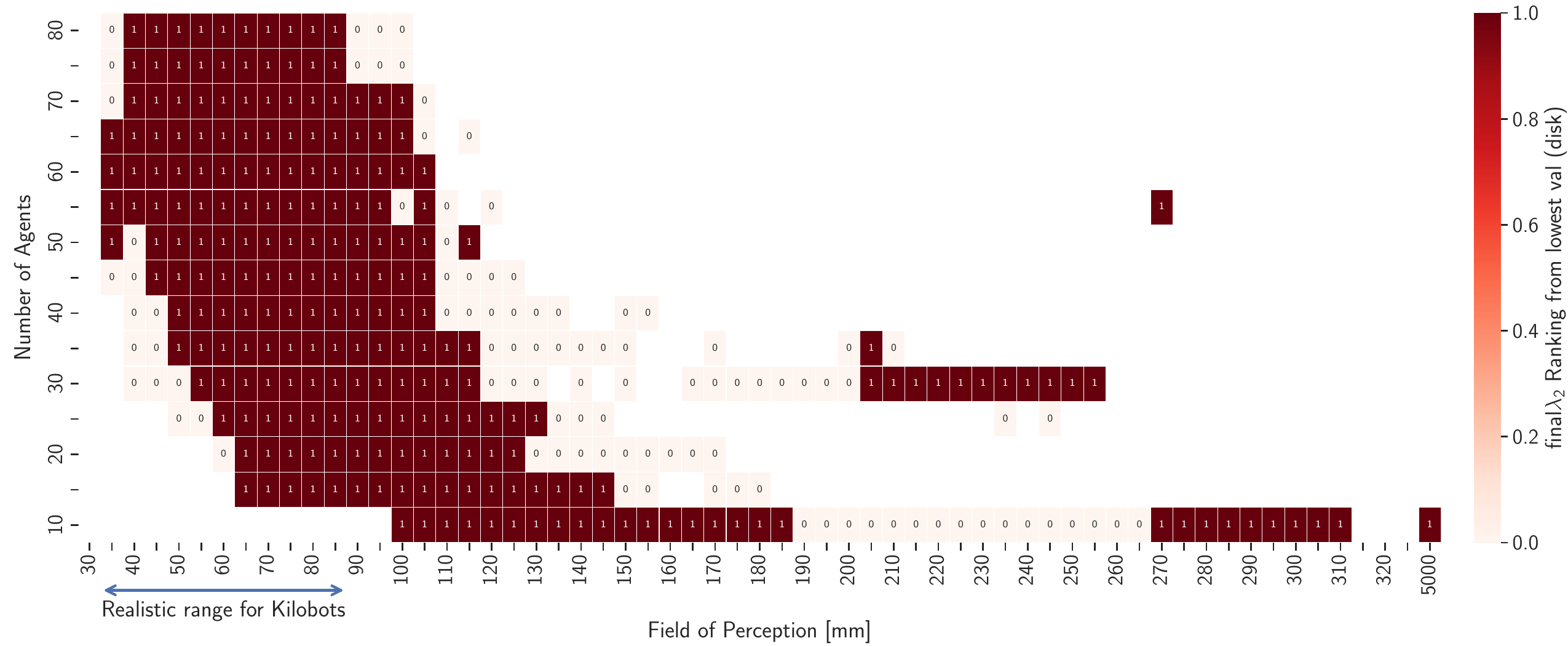}
};

\node[anchor=south] at (2,-16.5) {\large\textbf{sim-small-2-uniform}, 10 iterations, Ranking disk};
\end{tikzpicture}
}
\caption{Evolution of the accuracy with respect to the number of agents and their field of perception, for the case in simulation \textbf{sim-small-2-uniform}. In this case, we consider 2 arenas (disk vs annulus) of surface $S = 70000 mm^2$ and the robots always stay immobile during the simulation. Robots are distributed uniformly in the arenas, so that they are equidistant to their neighbors (obtained through Voronoi tessellation). All results are computed over 64 runs of simulation. We adjust the accuracy score to only show results where the simulated diffusion sessions in the swarm do not diverge: empty bins correspond to results where there are more than $50\%$ of runs exhibiting divergences. \textbf{Top:} results after one iteration. \textbf{Middle:} results after 10 iterations. \textbf{Bottom:} Ranking of the values of the Final$\lambda_2$ centroids for the disk arena.}
\label{fig:regimes_simsmall2uniform}
\end{center}
\end{figure*}

Finally, Supplementary Figs.~\ref{fig:regimes_simsmall2dispersion},~\ref{fig:regimes_simsmall2random} and~\ref{fig:regimes_simsmall2uniform} show the accuracy results after 1 and 10 iterations for cases \textbf{sim-small-2-dispersion}, \textbf{sim-small-2-random} and \textbf{sim-small-2-uniform}. They also show the influence of the parameters $N$ and $\sigma$ on the ordering of the values of $\texttt{Final}\lambda_2$ corresponding to the two arenas (cf Sec.~\ref{sec:bigarenas} for details). The results are consistent with our previous analysis in Supplementary Fig.~\ref{fig:diffRegimes}, with rankings showing clear demarcation lines between regimes.

In the end, the \textbf{sim-small-2-dispersion} and \textbf{sim-small-2-uniform} cases are complementary and may be selected depending on $N$ and $\sigma$. They exhibit similar results for the optimal $\alpha = 17$ hyperbole.

\FloatBarrier\clearpage

\subsection{Pogobot simulations: influence of communication and temporal noise}\label{sec:noiseRes}

\begin{figure*}[bp!]
\begin{center}
\includegraphics[width=0.50\textwidth]{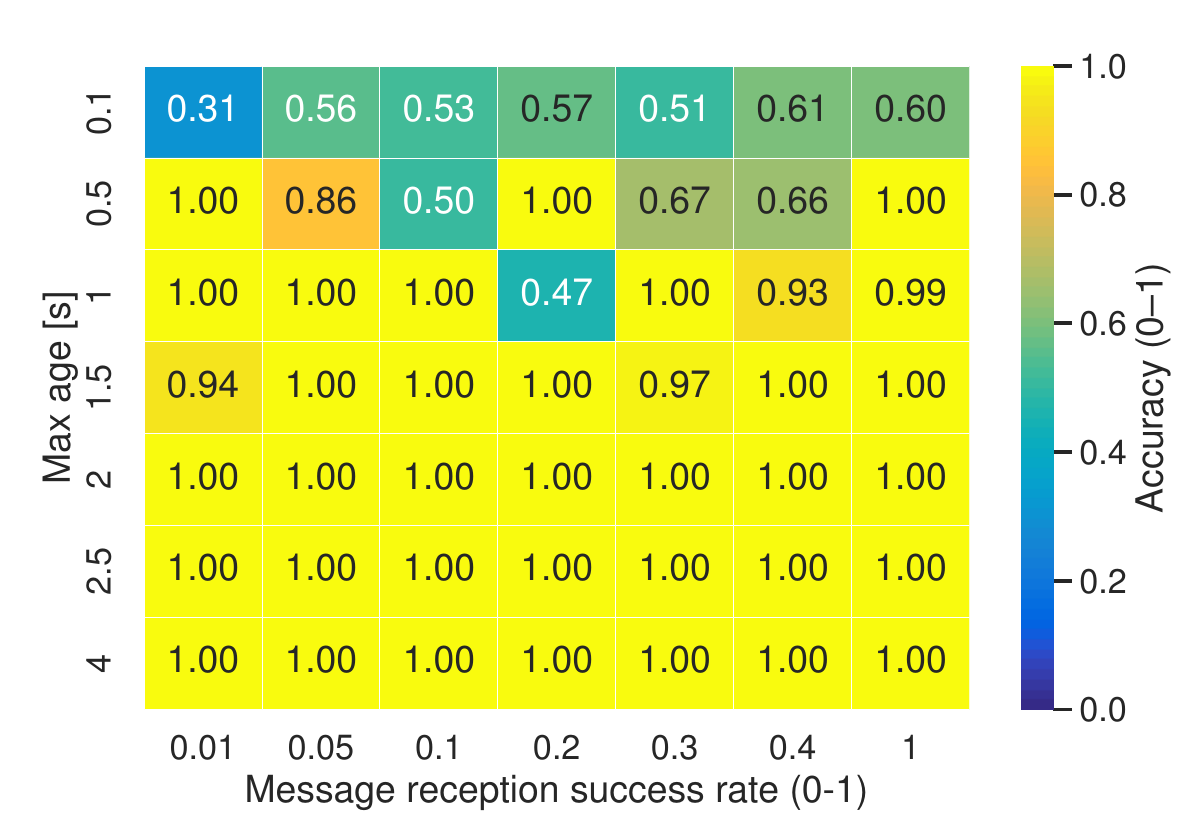}%
\caption{\textbf{Impact of communication noise on accuracy}, in simulations of one iteration of the SSR algorithm (32 runs), with 25 simulated Pogobot robots, and the same algorithm parameters as results in main text Figs.~6, 7 and 8. We vary two parameters: 1) the probability of a focal robot to correctly receive neighbors' messages; 2) the amount of time the neighbors messages are retained in the focal robot (after that, they are discarded). }
\label{fig:simulation_noise}
\end{center}
\end{figure*}

We investigated the impact of communication noise (\ie the proportion of correctly received messages by each robot) on the accuracy of the SSR algorithm.
Figure~\ref{fig:simulation_noise} shows that communication noise can be fully mitigated by selecting appropriate parameters. In particular, “max age” corresponds to the retention time of neighbors' messages in a focal robot. Increasing this parameter allows more time between diffusion steps, ensuring that more neighbors' messages for the current step are received. This increases the total time required to complete one iteration of the algorithm, but it reduces the likelihood of asymmetric connections appearing during a step (\ie a focal robot can send messages to a neighbor, but the neighbor cannot send messages back).

We also investigated the impact of temporal noise (\ie the temporal drift of each robots with respect to the aggregated temporal delay) on the accuracy of the SSR algorithm.
We considered the following protocol in Pogobot simulations: at the end of each main loop user code execution (\ie one pogotick) we add uncorrelated positive random values following a half-normal distribution $Z_{i,k} = |\mathcal{N}(0,\sigma^2)|$ to the current simulated time of each robot. This causes the robots' internal clocks to slowly drift apart, with increasing variance:
\[
\mathbb{E}[Z] = \sigma \sqrt{\tfrac{2}{\pi}}, \quad
\operatorname{Var}(Z) = \sigma^2\!\left(1 - \tfrac{2}{\pi}\right)
\]
For example, after 120 pogoticks (4 seconds at 30Hz, as in the experimentally tested cases):
\[
\mathbb{E}[X_{i,120}] = 120 \, \sigma \sqrt{\tfrac{2}{\pi}}, \quad
\operatorname{Var}(X_{i,120}) = 120 \, \sigma^2\!\left(1 - \tfrac{2}{\pi}\right)
\]

We tested how different levels of temporal drift variance impacted the accuracy of the SSR algorithm, using the following values of $\sigma$: $0.1$ ms, $1$ ms, $2$ ms, $10$ ms, $100$ ms.
In the first four cases, accuracy was completely unaffected (always $100\%$). Indeed, Pogobot messages contain a temporal synchronization term (cf Sec.~\ref{sec:impl_challenges}). If a focal robot is too far behind one of its neighbor, it adjusts its internal counter to the value indicated in the received message.
In the last case ($100$ ms), the algorithm could not function at all, likely because the induced drift was too large. Robots that resynchronizes their internal clocks to those of their neighbors often skipped several diffusion steps, introducing excessive asymmetries in the Laplacian matrix.

\FloatBarrier\clearpage

\subsection{Pogobot experiments/simulations over three arenas: disk, annulus, triangle}\label{sec:pogobotexp_3arenas}

Figure~\ref{fig:pogobot_3arenas} shows additional results of cases pogo-sim-2 (simulations) and pogo-expe-2 (experiments) across three arenas: disk, annulus, triangle and three iterations. This goes beyond the results provided in the main text that only focused on two arenas (Main Text Figs. 6, 7, 8) and one iteration.

We extend the 2-class results from the main text by adding a new triangular arena with a new triangle-related centroid value set in the C code controller. We use 15 runs for each experimental arena, and 64 runs for each simulated arena.
The 3-arena results are consistent with the 2-arena ones in the main text, and showcase the generalization capabilities of the algorithm to more classes than just the disk and annulus. After one iteration, we obtain an accuracy of $87.9\%$ in simulations and $75.4\%$ in experiments. After three iterations, we obtain an accuracy of $97\%$ in simulations and $93\%$ in experiments. The confusion matrices and accuracies of the 3-arena case are different from the 2-arena ones shown in Fig. 7 of main text as it is recomputed with the new arena. Also, results are shown at both iteration 0 and 2 instead of only at iteration 0. Indeed, as the problem is more complex with three arenas than with two arenas, a larger number of iterations is needed to converge to stable and precise $\lambda_2$ estimates and obtain the best results.

\begin{figure*}[htp!]
\begin{center}
\includegraphics[height=5cm]{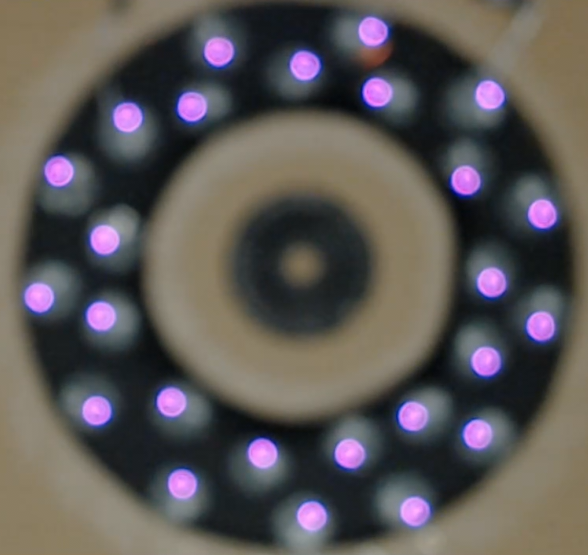}%
\includegraphics[height=5cm]{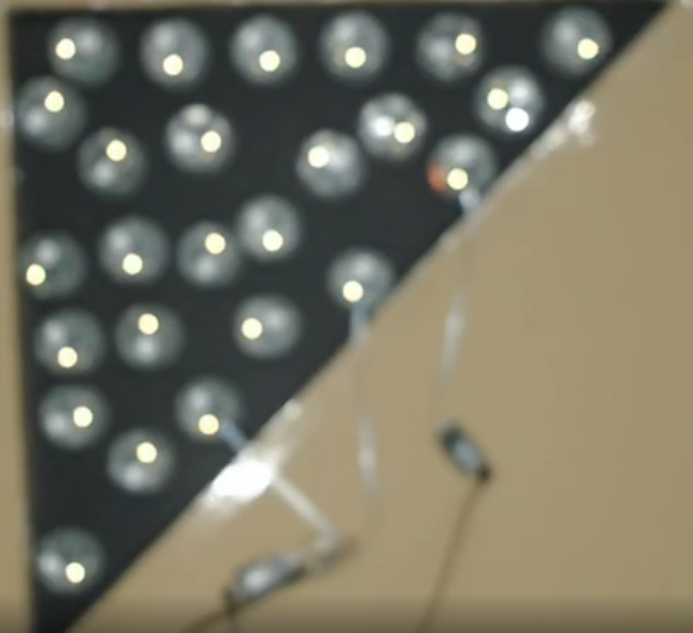}%
\includegraphics[height=5cm]{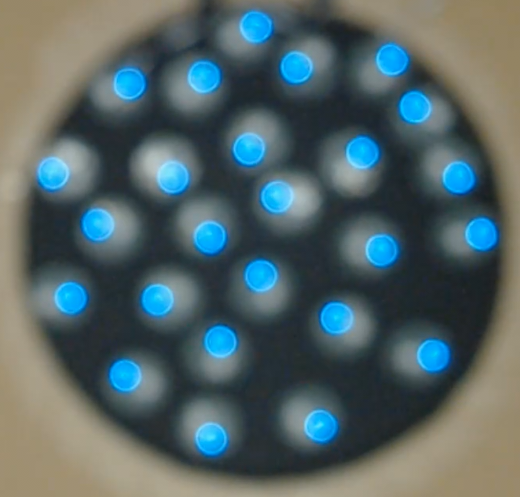}%
\caption{\textbf{Examples of experimental state at the end of an iteration}, with 25 Pogobot robots that light LED colors that correspond to their detected arena geometry: purple for the annulus, yellow for the triangle, cyan for the disk. The photos are blurred to ease visualization of robot LED colors.}
\label{fig:pogobot_3arenas_examples_expes}
\end{center}
\end{figure*}

\begin{figure*}[htp!]
\centering
\resizebox{0.80\textwidth}{!}{%
\begin{tikzpicture}[
panel_label/.style={
font=\large\bfseries,
anchor=north west,
fill=white,
fill opacity=0.75,
text opacity=1,
inner sep=1.5pt
},
panel_title/.style={
font=\large\bfseries,
anchor=south,
inner sep=1pt
}
]

% -------------------------------------------------------------------------
% Panel a
% -------------------------------------------------------------------------
\node[inner sep=0pt] (comp_it0) at (0,0) {%
\includegraphics[width=0.9\textwidth]{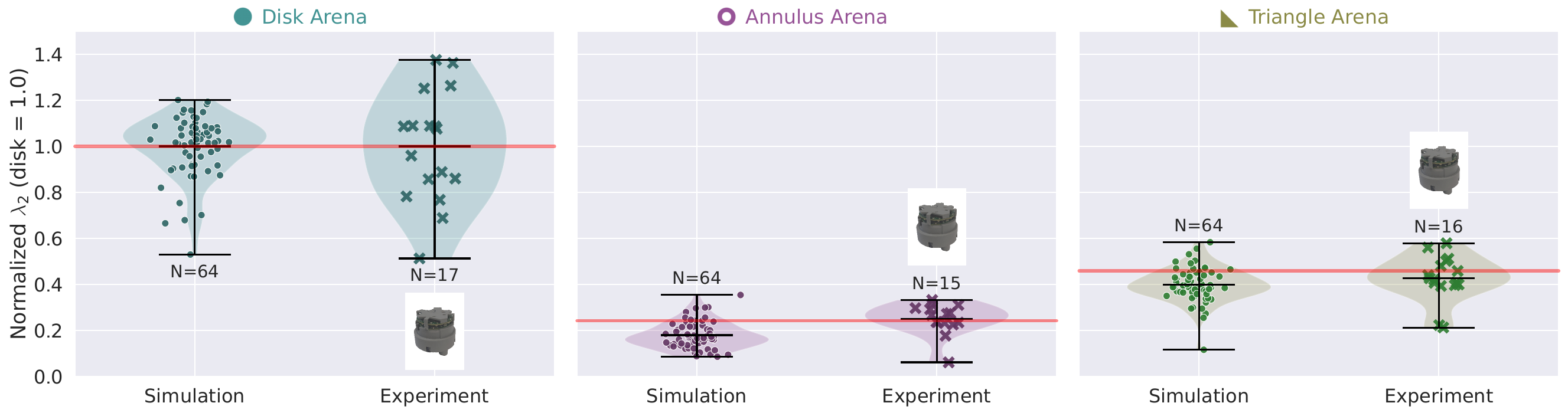}
};

\node[panel_label] at ([xshift=2mm,yshift=-2mm]comp_it0.north west) {a};
\node[panel_title] at ([yshift=2mm]comp_it0.north) {iteration 0};

% -------------------------------------------------------------------------
% Panel b
% -------------------------------------------------------------------------
\node[inner sep=0pt, below=8mm of comp_it0] (comp_it2) {%
\includegraphics[width=0.9\textwidth]{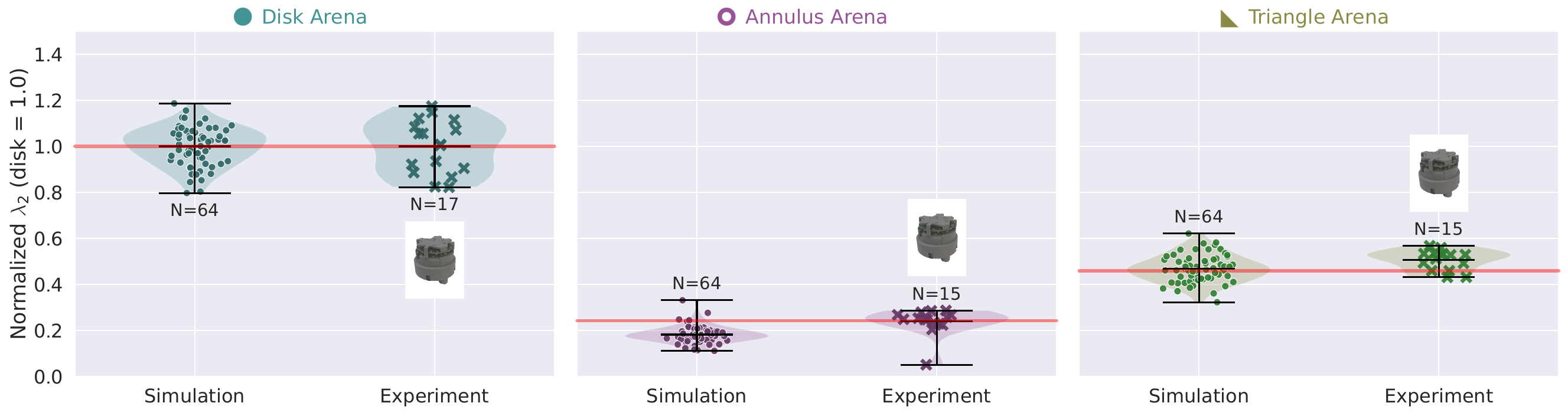}
};

\node[panel_label] at ([xshift=2mm,yshift=-2mm]comp_it2.north west) {b};
\node[panel_title] at ([yshift=2mm]comp_it2.north) {iteration 2};

% -------------------------------------------------------------------------
% Confusion matrices: iteration 0
% -------------------------------------------------------------------------
\node[inner sep=0pt, below=12mm of comp_it2.south west, anchor=north west, xshift=0.09\textwidth] (confusion_sim_it0) {%
\includegraphics[width=0.35\textwidth]{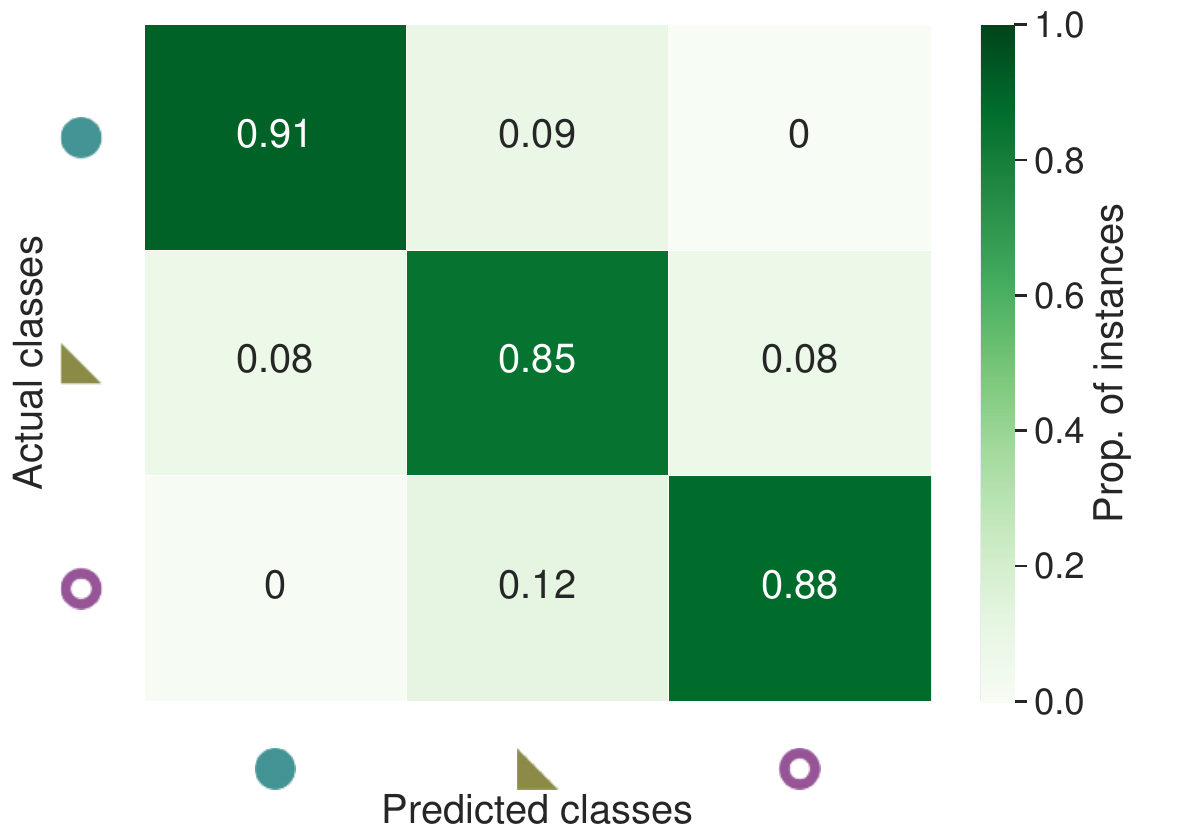}
};

\node[inner sep=0pt, right=10mm of confusion_sim_it0] (confusion_exp_it0) {%
\includegraphics[width=0.35\textwidth]{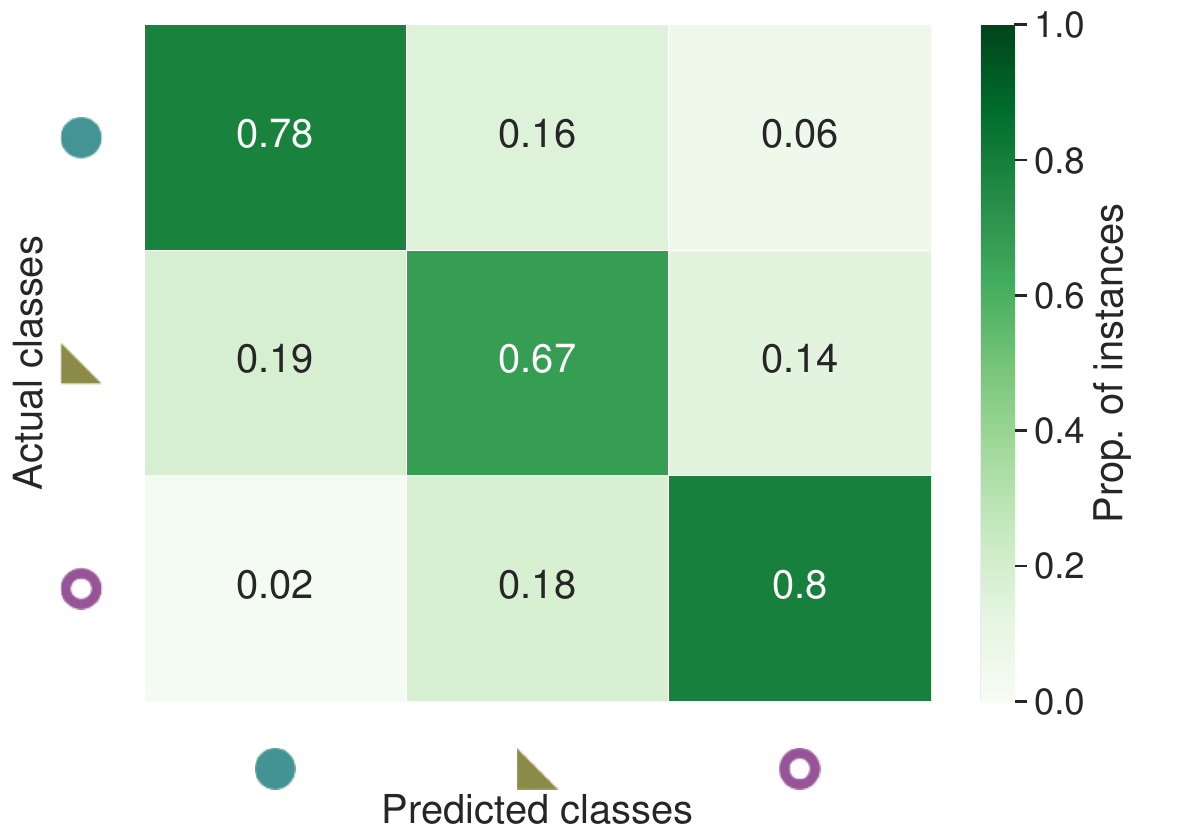}
};

\node[panel_label] at ([xshift=2mm,yshift=-2mm]confusion_sim_it0.north west) {c};
\node[panel_label] at ([xshift=2mm,yshift=-2mm]confusion_exp_it0.north west) {d};

\node[panel_title] at ([yshift=2mm]confusion_sim_it0.north) {simulations, iteration 0};
\node[panel_title] at ([yshift=2mm]confusion_exp_it0.north) {experiments, iteration 0};

% -------------------------------------------------------------------------
% Confusion matrices: iteration 2
% -------------------------------------------------------------------------
\node[inner sep=0pt, below=9mm of confusion_sim_it0] (confusion_sim_it2) {%
\includegraphics[width=0.35\textwidth]{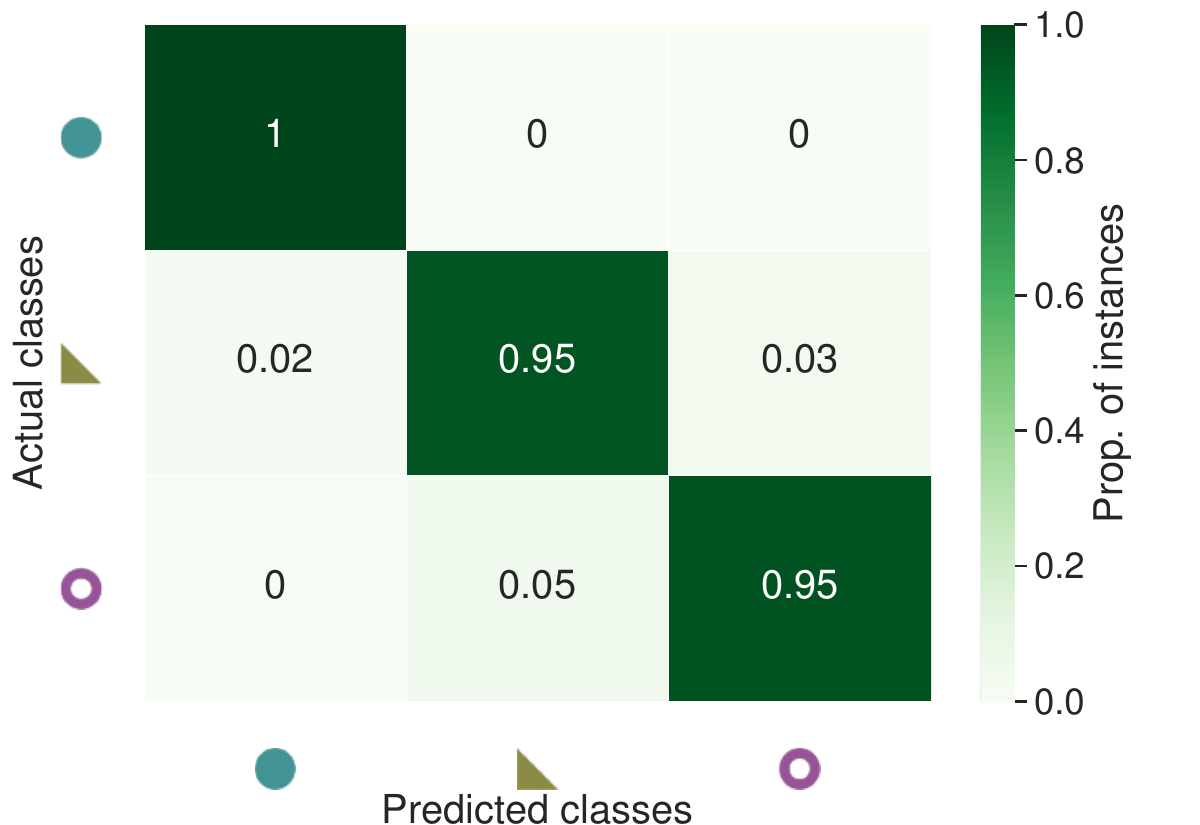}
};

\node[inner sep=0pt, right=10mm of confusion_sim_it2] (confusion_exp_it2) {%
\includegraphics[width=0.35\textwidth]{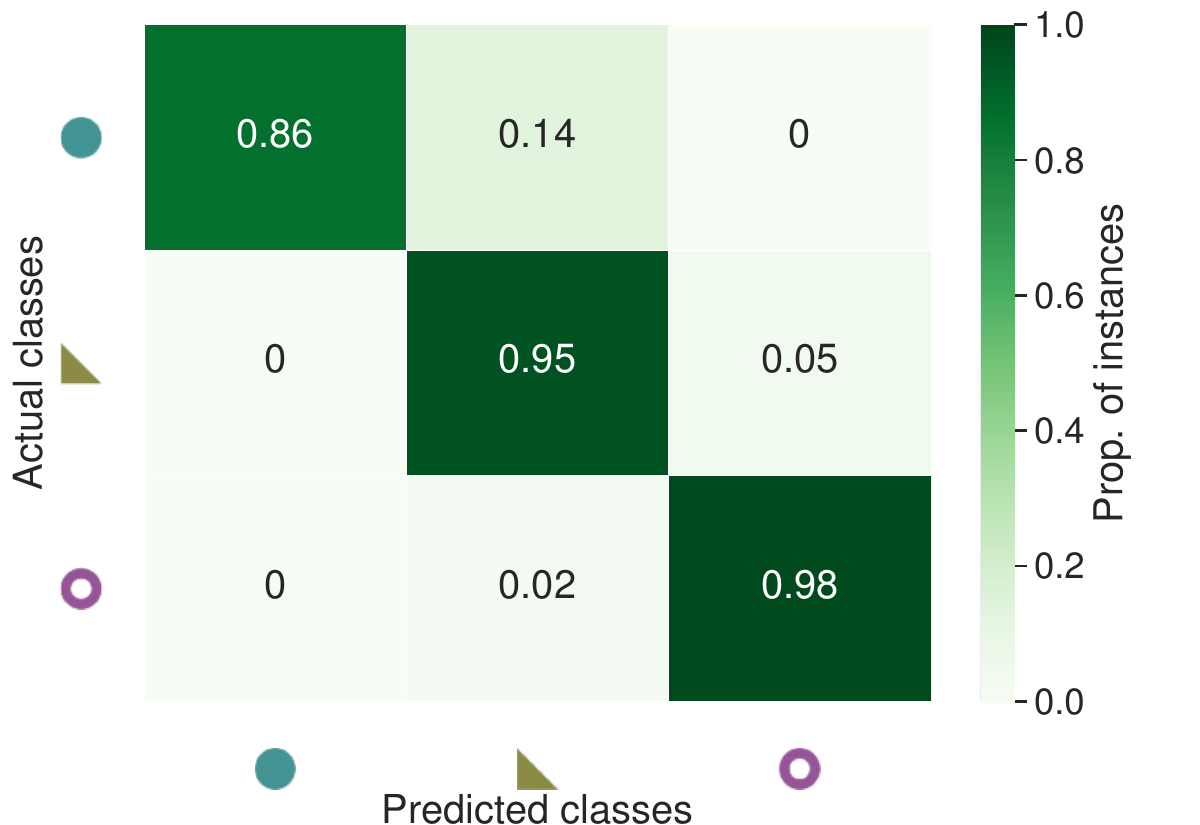}
};

\node[panel_label] at ([xshift=2mm,yshift=-2mm]confusion_sim_it2.north west) {e};
\node[panel_label] at ([xshift=2mm,yshift=-2mm]confusion_exp_it2.north west) {f};

\node[panel_title] at ([yshift=2mm]confusion_sim_it2.north) {simulations, iteration 2};
\node[panel_title] at ([yshift=2mm]confusion_exp_it2.north) {experiments, iteration 2};

\end{tikzpicture}%
}

\caption{\textbf{Comparison between theoretical, simulated, and experimental results on a three-class classification setting with three shapes: disk, annulus and triangle.} \small{
\textbf{a:}~Distribution of $\lambda_2$ values estimated on each robot after one iteration of the algorithm in simulations (pogo-sim-2 case), compared to the value computed using global information from the continuous Laplacian (red lines) and Pogobot experiments (pogo-expe-2 case). Values are normalized so that the mean of disk values is $1.0$. Comparisons used Mann--Whitney--Wilcoxon tests with Bonferroni correction; adjusted $P$ values and comparison groups are provided in data file S1 accompanying the published article.
\textbf{b:}~Distribution of $\lambda_2$ values estimated on each robot after three iterations of the algorithm in simulations, compared to the value computed using global information from the continuous Laplacian and Pogobot experiments. Values are normalized so that the mean of disk values is $1.0$. Comparisons used Mann--Whitney--Wilcoxon tests with Bonferroni correction; adjusted $P$ values and comparison groups are provided in data file S1 accompanying the published article.
\textbf{c:}~Confusion matrix of simulations over 64 runs per arena, after one iteration of the algorithm. Accuracy: $87.9\%$.
\textbf{d:}~Confusion matrix of experiments over 15 runs per arena, after one iteration of the algorithm. Accuracy: $75.4\%$.
\textbf{e:}~Confusion matrix of simulations over 64 runs per arena, after three iterations of the algorithm. Accuracy: $97\%$.
\textbf{f:}~Confusion matrix of experiments over 15 runs per arena, after three iterations of the algorithm. Accuracy: $93\%$.
}}
\label{fig:pogobot_3arenas}
\end{figure*}

Figure~\ref{fig:pogobot_3arenas_examples_expes} provides representative examples of experimental states at the end of an iteration across the three arenas, with robots displaying LED colors that correspond to their detected arena geometry.

\FloatBarrier\clearpage

\subsection{Comparison of the SSR algorithm with simpler baselines}\label{sec:baselines}

We compare the results in simulation of the SSR algorithm with three simpler baselines:
\begin{description}
    \item[average degree] where the robots estimate their number of neighbors and average it by consensus (collective averaging). This value is then used as a feature for the classification process instead of $\lambda_2$.
    \item[degree statistics] where the robots reach a consensus on more complex statistics based on local degree (number of neighbors): mean, variance, and the 25th, 75th percentiles. The classification is then performed based on the mean distance of these features compared to reference centroids.
    \item[convergence time] where the robots run a diffusion process as with SSR (without pre-diffusion) but assess the amount of time needed to reach local convergence (\ie $s_i$ reaches a plateau). This convergence time is then averaged by consensus (collective averaging) and used as a feature for the classification process.
\end{description}
Like the SSR algorithm, these baselines assume that the robots classify the arenas in a decentralized way, relying only on local robot-to-robot communication and not external sensors (\eg wall detection).
They can therefore be compared under the same sensing assumptions as the proposed algorithm.
We use the same simulation parameters as in the pogo-sim-2 case, but consider two initial swarm formations in the arena: either fully random, where the $x$, $y$, and $\theta$ values are sampled uniformly at random within the arena surface, or uniformly distributed, obtained through Lloyd-relaxation of the arena surface, \ie an iterative procedure that repeatedly uses Voronoi diagrams to make points more evenly distributed~\cite{lloyd1982least,du1999centroidal}. The initial formation matters because metrics based on average degree vary much less in a Lloyd-relaxed formation, where distances between neighboring robots are equalized, than in a random distribution, and can therefore serve as a stronger proxy for arena geometry. With a Lloyd-relaxed swarm, the ``average degree'' baseline no longer assesses whether a noisy swarm can extract shape information from local connectivity; it is closer to testing whether a well-spread sampling of the arena has a different mean neighborhood size.

\begin{figure*}[p!]
\begin{center}
\setlength{\tabcolsep}{3pt}
\renewcommand{\arraystretch}{1.0}

\newcommand{\confmat}[1]{%
    \raisebox{-.5\height}{\includegraphics[width=0.38\textwidth]{#1}}%
}
\newcommand{\formlabel}[1]{%
    \raisebox{-.5\height}{\rotatebox{90}{\textbf{#1}}}%
}

\begin{tabular}{c cc}
 & \textbf{SSR} & \textbf{Average degree} \\[0.2em]

\formlabel{Random}
&
\confmat{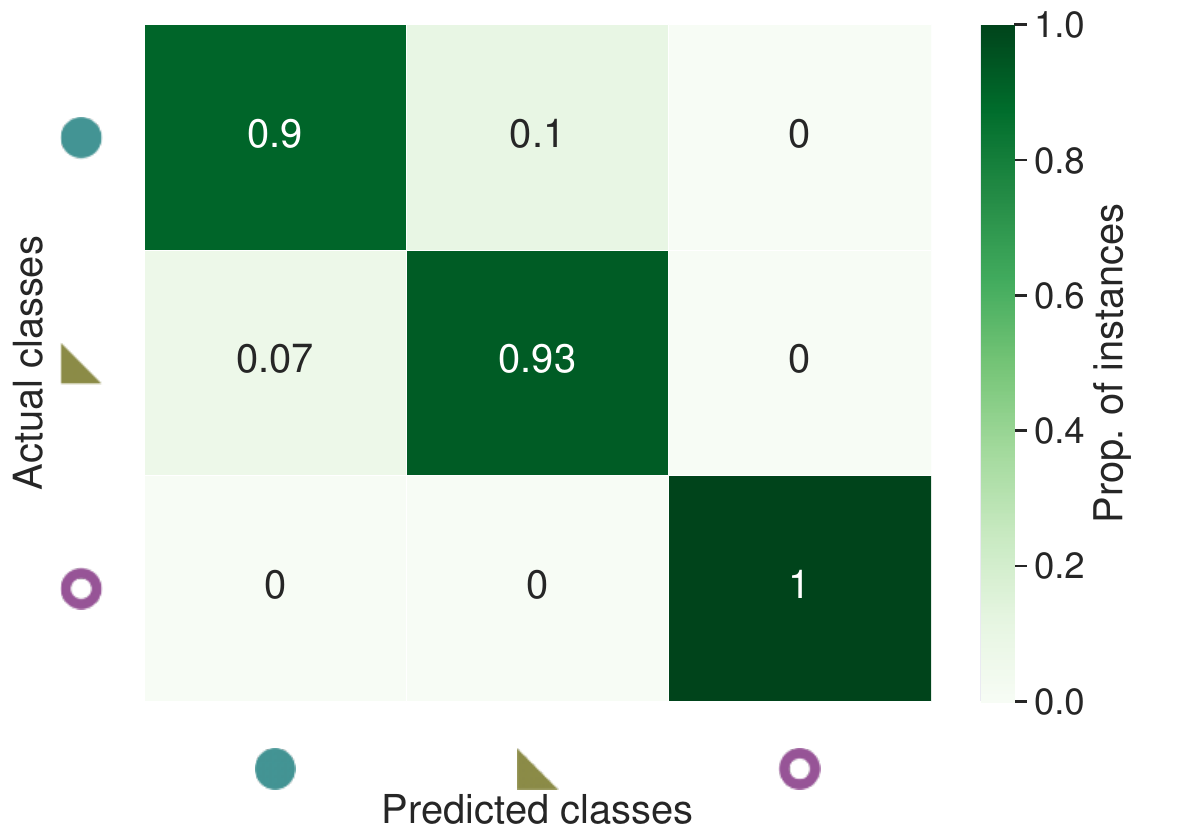}
&
\confmat{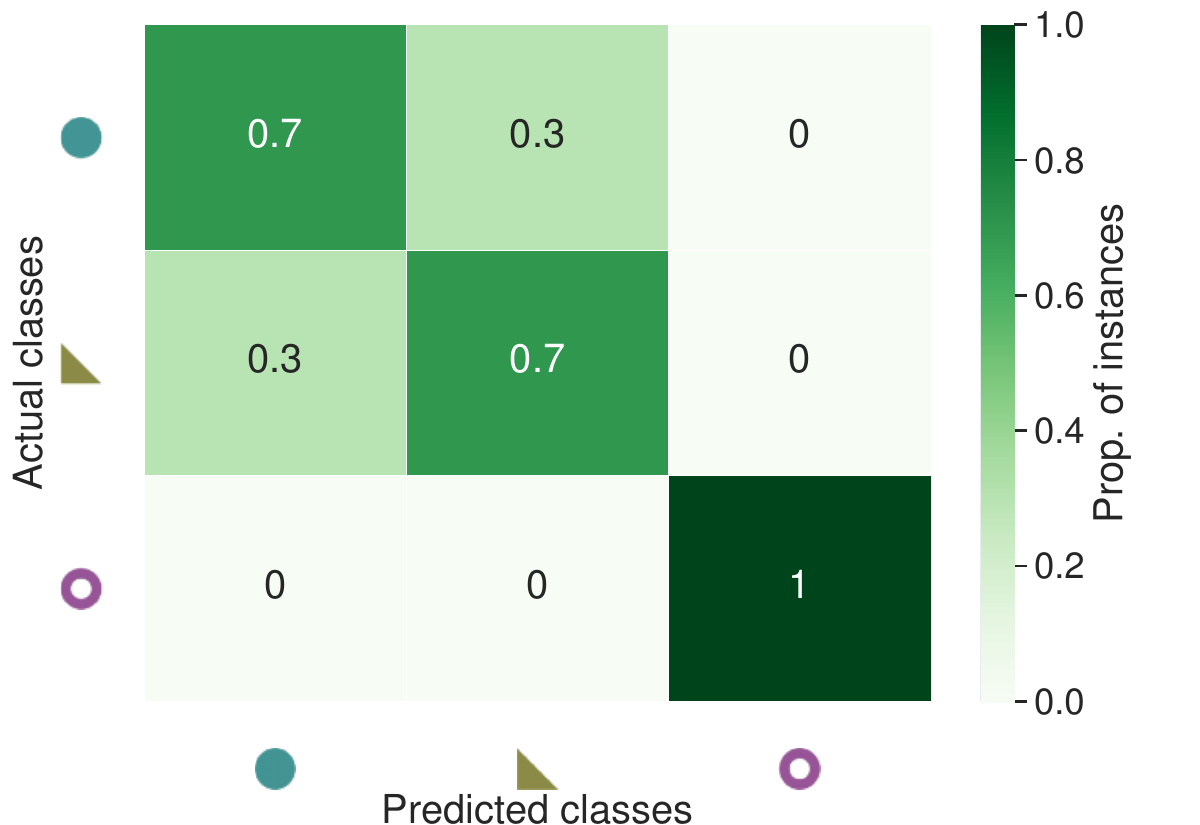}
\\[0.2em]

\formlabel{Lloyd-relaxed}
&
\confmat{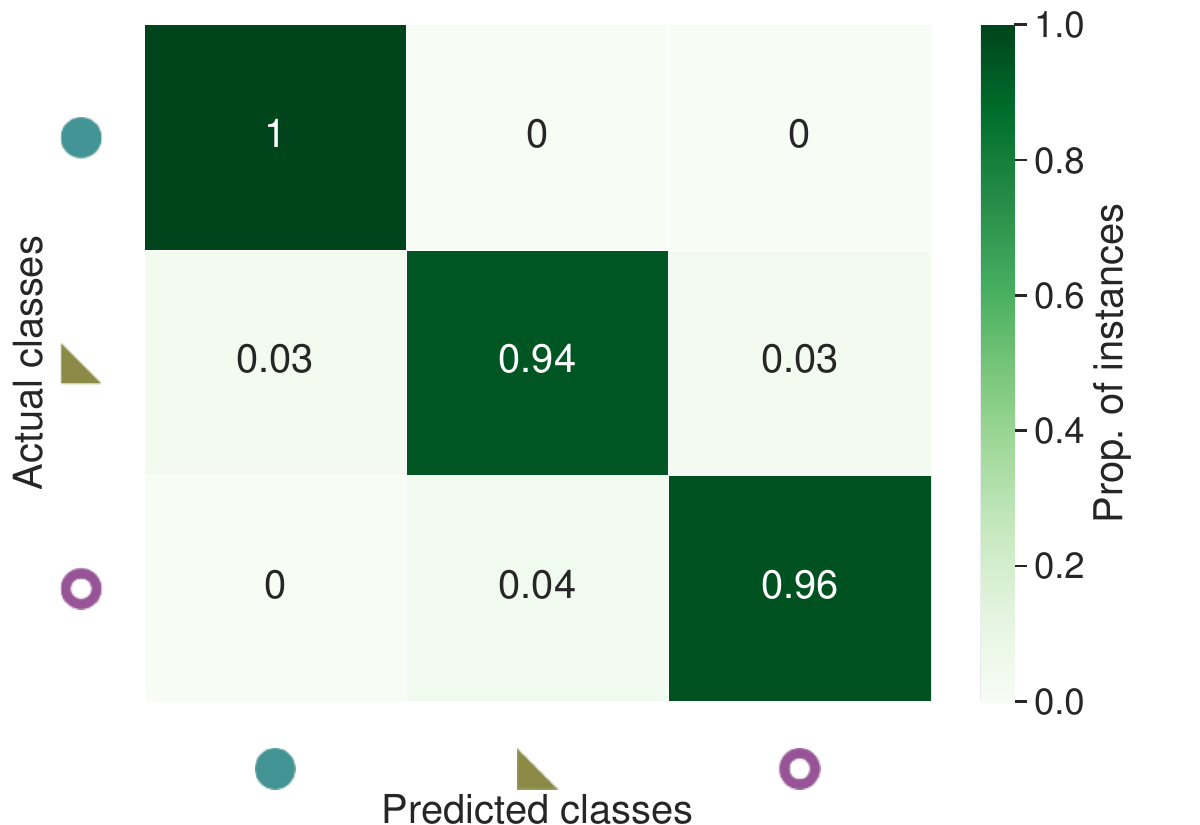}
&
\confmat{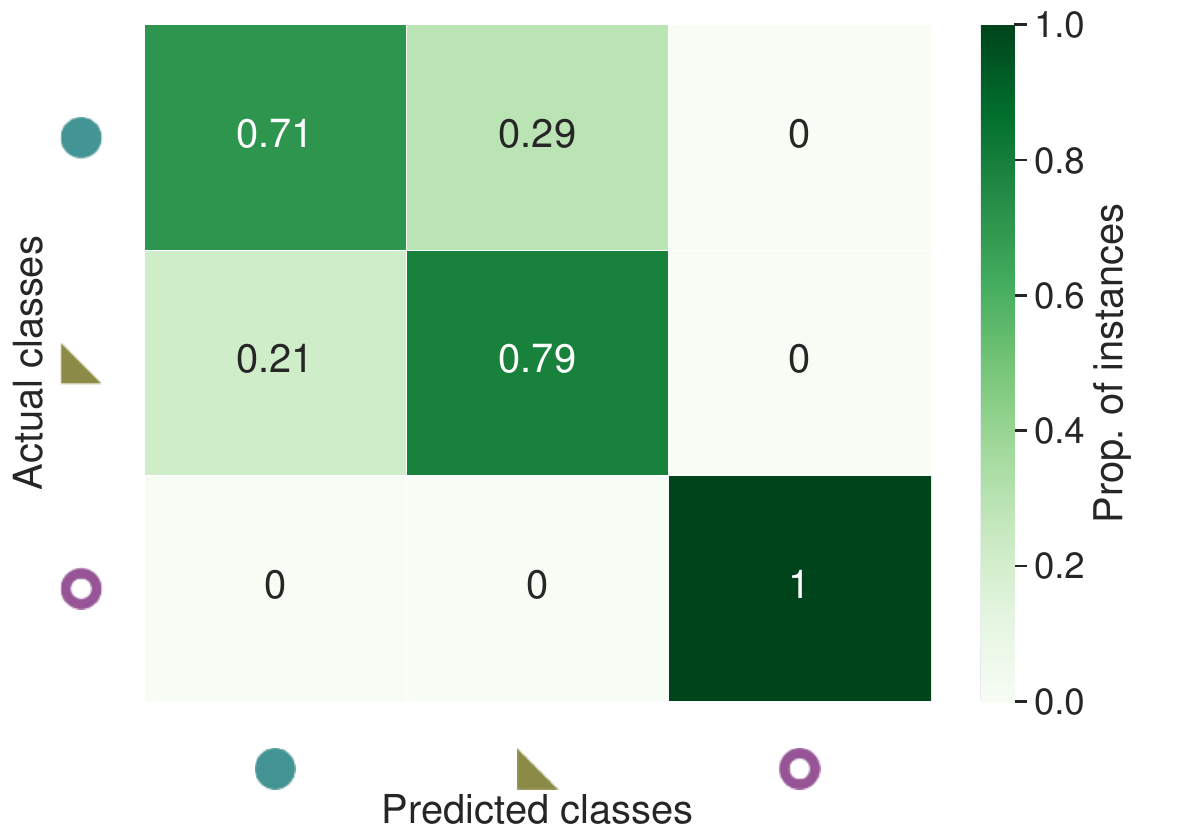}
\end{tabular}

\vspace{3.0em}

\begin{tabular}{c cc}
 & \textbf{Degree statistics} & \textbf{Convergence time} \\[0.2em]

\formlabel{Random}
&
\confmat{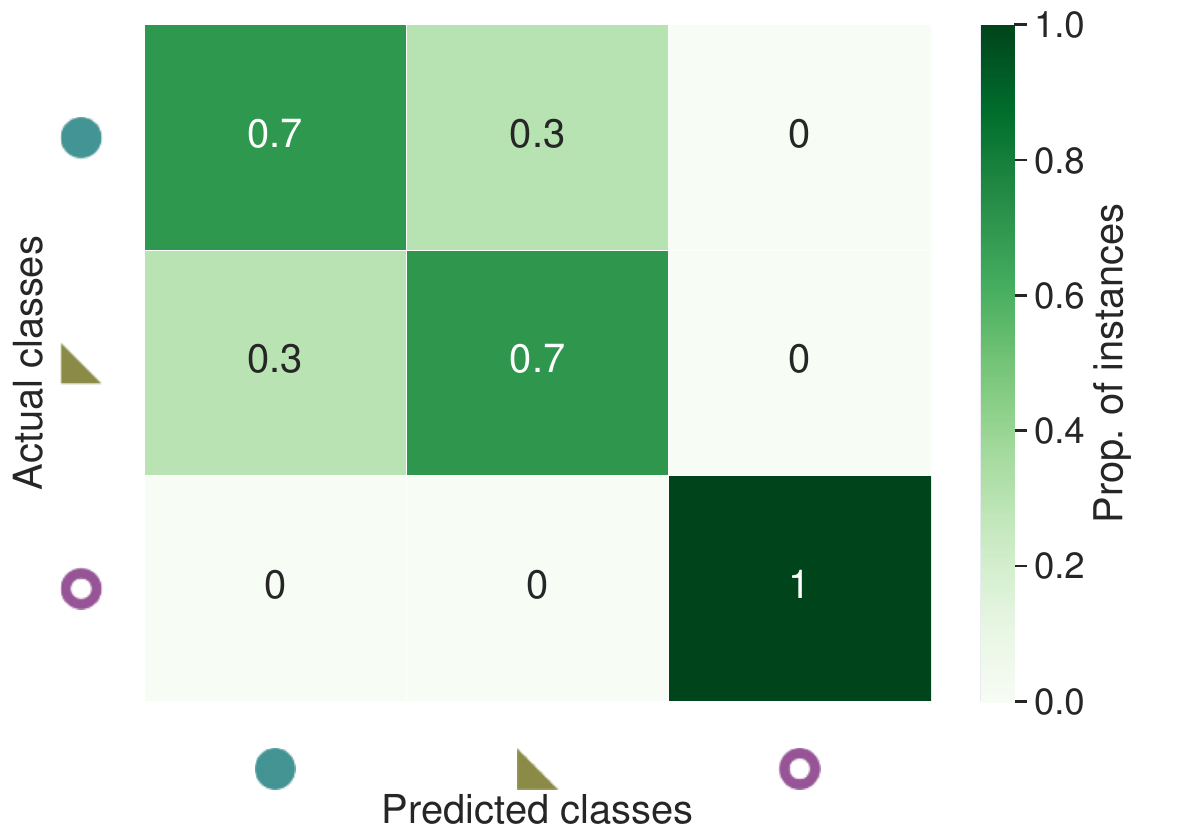}
&
\confmat{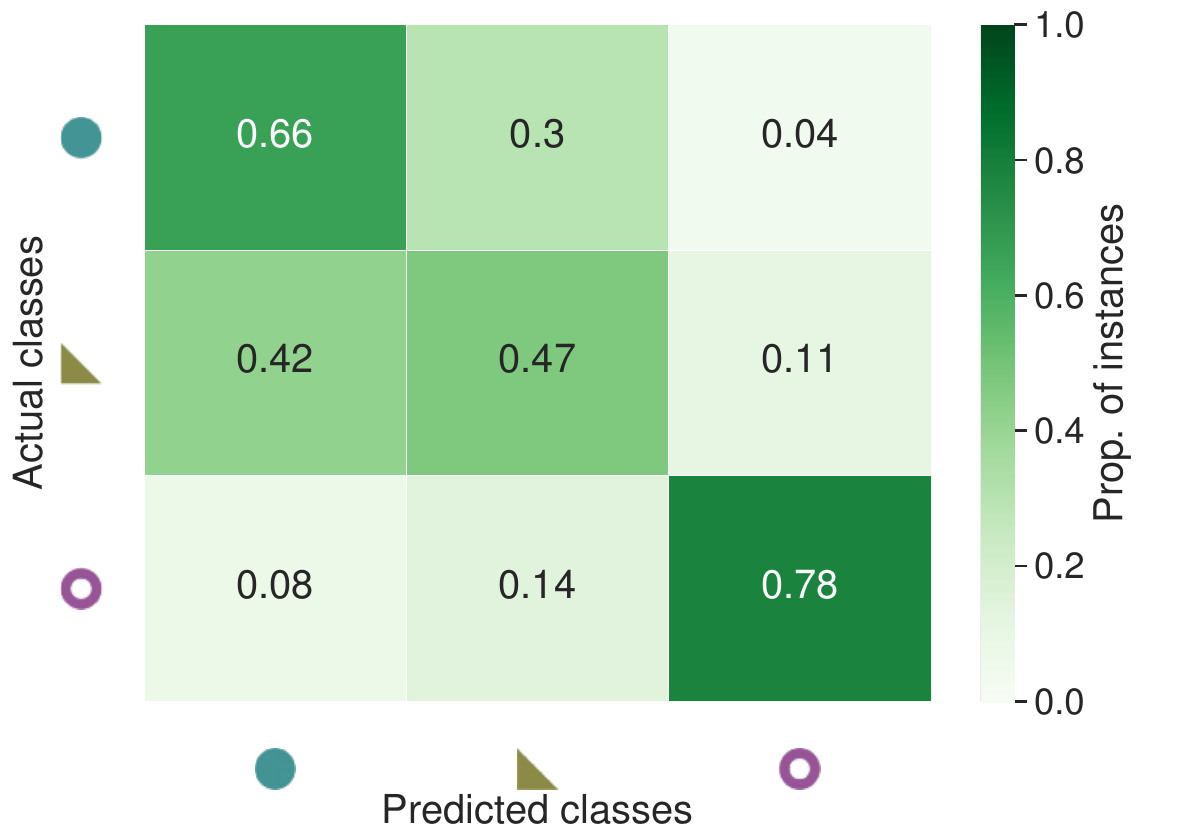}
\\[0.2em]

\formlabel{Lloyd-relaxed}
&
\confmat{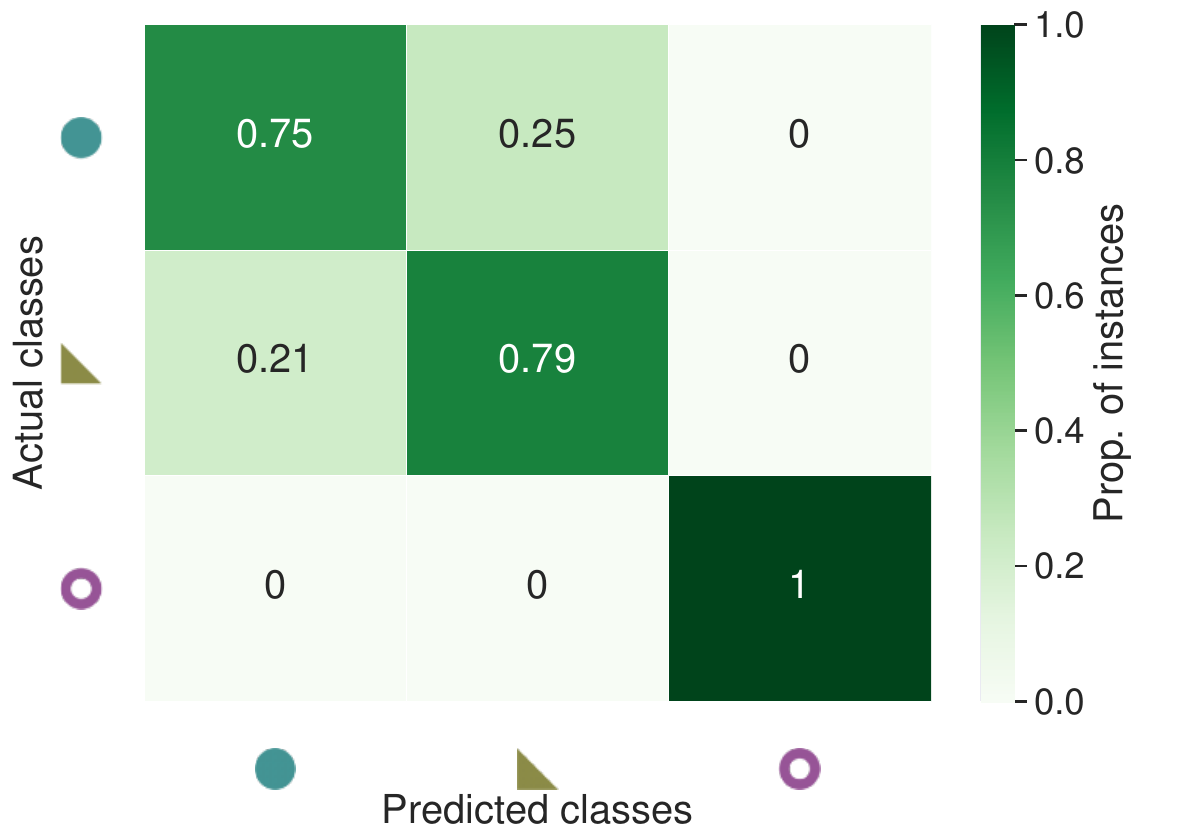}
&
\confmat{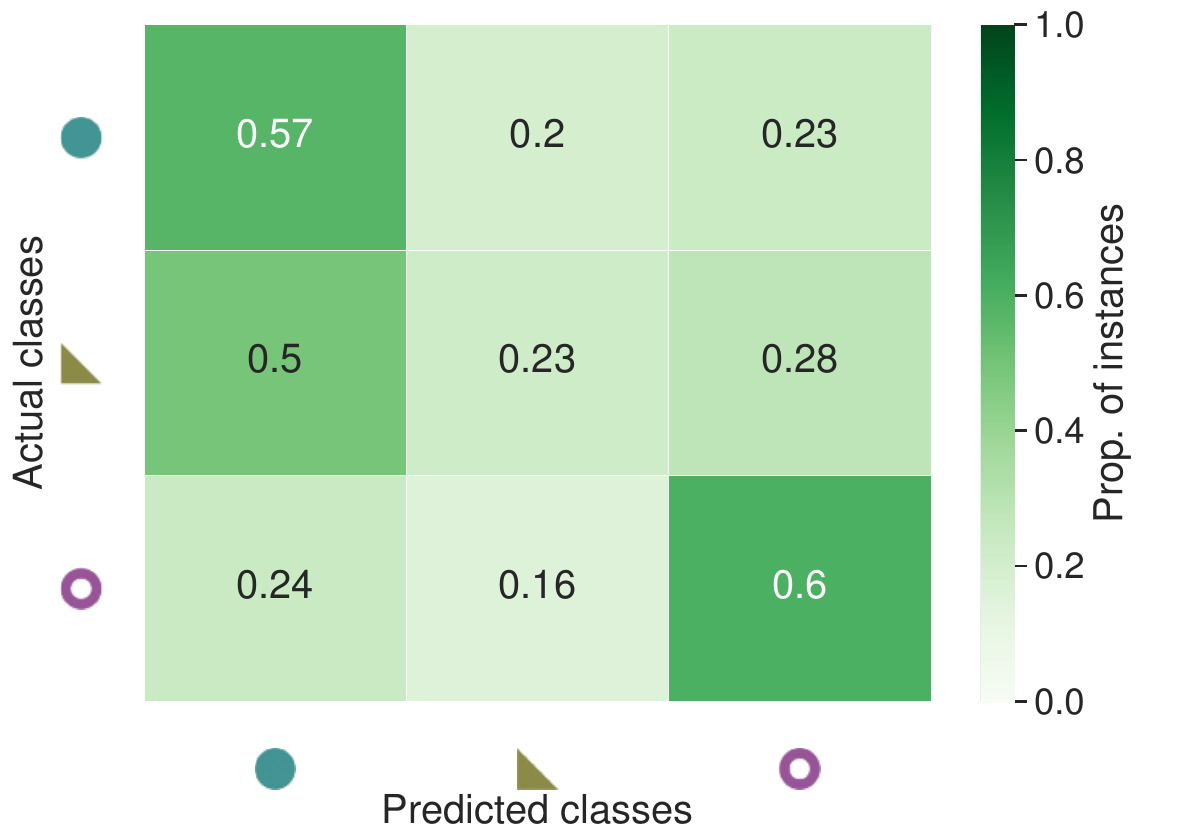}
\end{tabular}

\caption{\textbf{Confusion matrices of the baselines}, in 64-run simulations of 25 Pogobots robots with parameters from the pogo-sim-2 case, across three arenas: disk, annulus and triangle and two initial formations: random and Lloyd-relaxed (uniform). Rows correspond to the two initial formations. Columns correspond to the four classification methods: SSR, average degree, degree statistics and convergence time.}
\label{fig:baselines_confusion}
\end{center}
\end{figure*}

\begin{table*}[h!]
\begin{center}
\scriptsize
\resizebox{0.80\textwidth}{!}{%
\begin{tabular}{|c|c|c|}
\hline
Baseline name & Random formation accuracy & Lloyd-relaxed formation accuracy \\
\hline
SSR (base algorithm) & $\mathbf{93.9\%}$ & $\mathbf{96.7\%}$ \\
average degree & $80.1\%$ & $83.2\%$ \\
degree statistics & $80.6\%$ & $84.5\%$ \\
convergence time & $63.7\%$ & $46.5\%$ \\
\hline
\end{tabular}
}
\caption{Table of accuracy scores of baselines compared to the SSR algorithm (in bold), across 64-run simulations using the parameters of the pogo-sim-2 case, after 3 iterations.}
\label{tab:baselines_accuracy}
\end{center}
\end{table*}

Figure~\ref{fig:baselines_confusion} shows the confusion matrix across baselines, compared to the SSR algorithm, after 3 iterations. The overall accuracy of each method is listed in Table~\ref{tab:baselines_accuracy}.

These results show that simple local-connectivity descriptors already contain substantial information about arena shape, but fail to match the performance of SSR.
In particular, the average-degree and degree-statistics baselines reach relatively high accuracies, indicating that differences in accessible area, boundary exposure, and local robot density contribute to the classification signal.
However, both remain clearly below SSR in both initial conditions. The limited improvement of degree statistics over average degree further suggests that, for these arenas, most of the information available from static degree distributions is captured by the first moment, while higher-order degree features add only marginal discrimination.
By contrast, SSR exploits the relaxation dynamics of the communication graph and is therefore sensitive not only to the number of local connections, but also to their global organization.
The poor performance of the convergence-time baseline further indicates that a crude local plateau-detection heuristic is not sufficient to recover this information: estimating the spectral relaxation rate provides a more robust descriptor than measuring local convergence time alone.
Overall, these comparisons support the interpretation that SSR is not merely classifying arenas through mean connectivity or boundary exposure, but extracts additional global information from the distributed diffusion process.

An additional distinction is that the local-connectivity baselines are not intrinsically scale-invariant. Mean degree and degree-distribution statistics depend directly on the ratio between communication range, robot density, and arena size; changing the arena scale, the number of robots, or the communication radius can therefore modify these descriptors even when the underlying shape is unchanged. By contrast, the spectral quantity estimated by SSR has a well-defined scaling law: for a continuous domain, Laplacian eigenvalues scale as the inverse square of length, so $\lambda_2$ can be normalized by an appropriate geometric factor, such as area, to obtain a scale-independent shape descriptor. This makes the SSR descriptor more directly related to global arena geometry, whereas the simpler baselines partly conflate shape with sampling density and communication-range effects.

\FloatBarrier

\subsection{Experimental runs \& Supplementary Videos}

\begin{figure*}[p!]
\begin{center}
\includegraphics[width=1.00\textwidth]{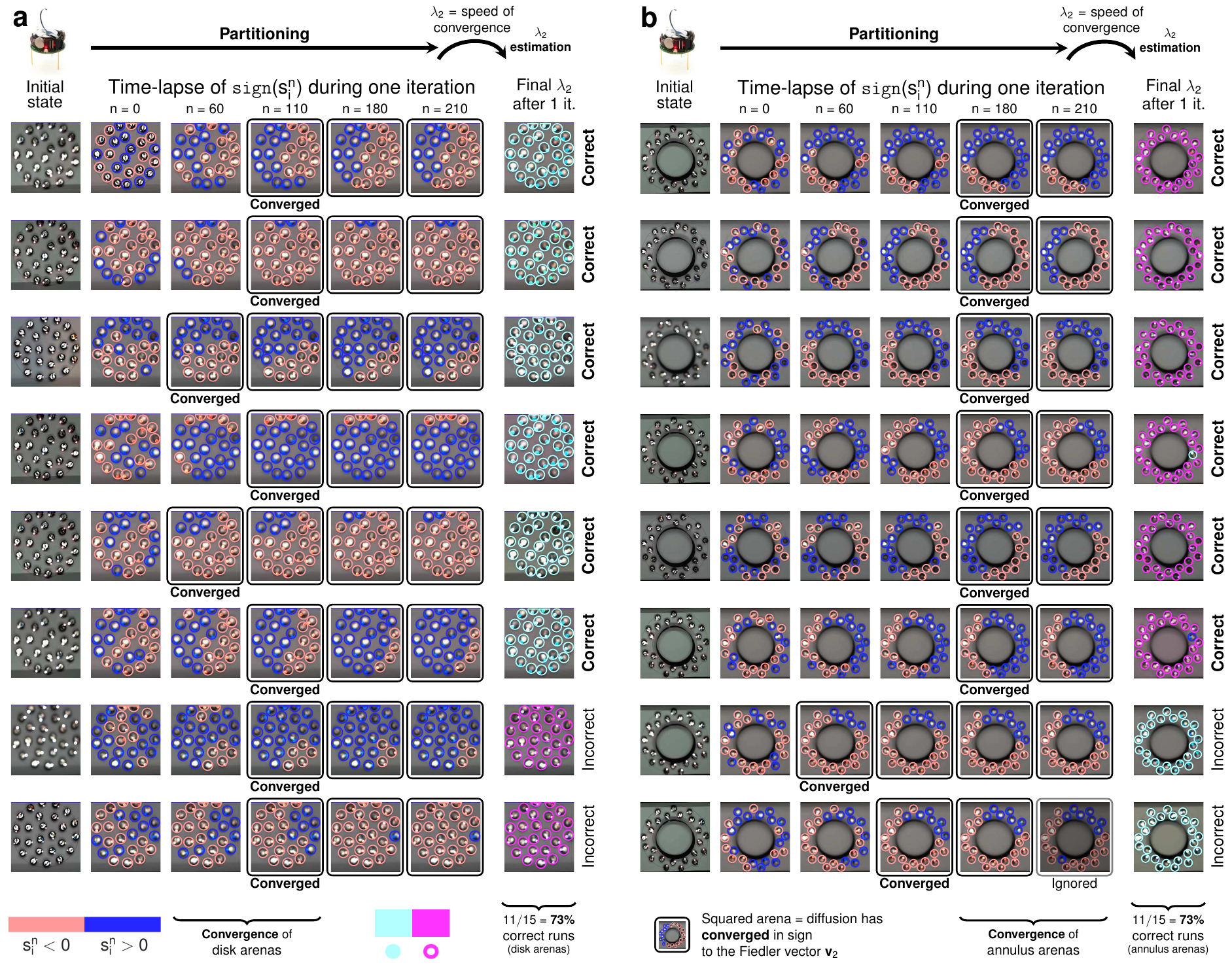}%
\caption{\textbf{Time-lapses of the experiments with Kilobots}, with 25 Kilobot robots, respectively for the disk (\textbf{a}) and annulus (\textbf{b}) arenas. \small{ Experiments use the same parameters as simulations and experiments of main text Figs.~6, 7 and 8. Only the results of 16 runs are shown (over 30 considered runs). Time-lapses of all runs are presented in Supplementary movie 1. Experimental photos are blurred to ease the visualization of LED colors (except left column). Images are re-scaled for better visualization; experiments have arenas surface approximately equal to $70000\texttt{mm}^2$.
\textbf{Left}:~Initial state of the robots before the start of experiments.
\textbf{Middle}:~diffusion, resulting in partitioning (the color of each robot corresponds to a local component of the Fiedler vector $\textbf{v}_2$). In the studied configuration, robots typically converge in the domain $n\in [60, 110]$ for the disk arenas and in the domain $n\in [180, 210]$ for the annulus arenas. In rare cases, the robots may still change state after convergence, due to computation errors, or message transmission errors -- in such case, the partitioning the robots converged into may be broken (\textit{e.g.,} last experiment on the annulus arena). The spectral swarm robotics algorithm will detect and ignore divergent cases after convergence (cf details in Supplementary Sec.~3).
\textbf{Right}:~individual $\lambda_2$ estimation on each robot from the convergence rate of diffusion, and consensus (collective averaging) over the entire swarm. Robots show a LED color-code according to the value of $\lambda_2$: cyan when a disk is detected, violet for an annulus.
The robots detected the correct shape on 22 runs over 30 runs (15 runs on each arena), which translates into an accuracy score of $22/30 = 73\%$. }}
\label{fig:expes_kilobots}
\end{center}
\end{figure*}

All Kilobot and Pogobot experiments respectively involve the experimental cases \textbf{expe-2} and \textbf{pogo-expe-2} described in Supplementary Table~\ref{tab:cases}. Individual runs of Kilobots and obtained classification results are listed in Supplementary Table~\ref{tab:expe_runs_list}. Pogobots experimental runs videos only contain a single arena (either disk, annulus or triangle), and with a very high accuracy of $100$\% for two classes (disk-annulus) and $93\%$ for three classes, the classification has very few errors, so we do not list them in a Table.
The raw video files of all Kilobots experiments are made available at the following link:~{\scriptsize \url{https://doi.org/10.6084/m9.figshare.31069147}}. The raw video files of Pogobots experiments (including all disk, annulus experiments and one example of triangle experiment) are found at:~{\scriptsize \url{https://doi.org/10.6084/m9.figshare.31027951}}.
\\

Figure~\ref{fig:expes_kilobots} shows example time-lapses of the SSR algorithm in Kilobots experiments -- similar to main text Fig.~8, but with Kilobots instead of Pogobots. Here only $73\%$ of experiments correctly classified the shape of their arena (compared to $100\%$ with Pogobots), and the experiments took significantly more time (around 1 hour per iteration, compared to 3 minutes with Pogobots).
\\

Additionally, we provide four explanatory videos as supplementary materials, available at ~{\scriptsize\url{https://doi.org/10.6084/m9.figshare.31074412}} and described as follow:

Video \textbf{SI1} presents the results of the \textbf{expe-2} case, involving experiments with 25 Kilobot robots in arenas of surface area $S = 70000 \texttt{mm}^2$. First, we display the observed behavior of the algorithm on one experimental run for both the disk and annulus arenas. This is illustrated by a time-lapse showing the active LED colors of each robot during both the ``diffusion" and ``$\lambda_2$ estimation" stages of the algorithm. We then present the results obtained across all Kilobot experimental runs listed in Supplementary Table~\ref{tab:expe_runs_list}.

Video \textbf{SI2} compares experimental results from case \textbf{expe-2} with simulation results from case \textbf{sim-2}. Both experimental and simulated agents are driven by the same controllers implementing the spectral swarm robotics algorithm on a Kilobot architecture. In both cases, we observe similar dynamics during the “diffusion” (the swarm exhibits correct partitioning) and “$\lambda_2$ estimation” (the swarm correctly classifies arenas) stages.

Video \textbf{SI3} showcases swarm behavior in the \textbf{sim-large-7} case, with parameters from regime r3 (shown in Main Text Figs.~4 and 5). It involves 300 agents with a field of perception of $85,\text{mm}$ and 7 arenas of surface area $S = 500000 \texttt{mm}^2$. This video presents a time-lapse similar to Main Text Fig.~4, for both the “diffusion” and “$\lambda_2$ estimation” stages of the algorithm.

Video \textbf{SI4} presents the results of the \textbf{pogo-expe-2} case, involving experiments with 25 Pogobot robots in arenas of surface area $S = 150000 \texttt{mm}^2$. First, we display the observed behavior of the algorithm on one experimental run for both the disk and annulus arenas. This is illustrated by a time-lapse showing the active LED colors of each robot during both the ``diffusion" and ``$\lambda_2$ estimation" stages of the algorithm. We then present the results obtained across all Pogobot experimental runs.

\begin{table*}[h!]
\begin{center}
\normalsize
\resizebox{0.80\textwidth}{!}{%
\begin{tabular}{|p{1.2cm} |p{4.4cm} |p{2.5cm} p{2.5cm} |p{6.0cm}|}
\hline
Exp. number & Associated video file & Disk classification & Annulus classification & Additional notes \\
\hline
0  & \url{2022-07-21_14-16-04.mp4}  & \textbf{Disk}     & \textbf{Annulus}       & Disk LED color is orange instead of cyan (early version of the color set), code, parameters, centroids are the same as other experiments. \\
1  & \url{2022-07-21_15-47-12.mp4}  & \textbf{Disk}     & \textbf{Annulus}       & - \\
2  & \url{2022-07-22_10-18-21.mp4}  & \textbf{Disk}     & \textbf{Annulus}       & - \\
3  & \url{2022-07-22_11-50-10.mp4}  & Annulus           & \textbf{Annulus}       & - \\
4  & \url{2022-07-22_13-33-07.mp4}  & \textbf{Disk}     & Disk                   & - \\
5  & \url{2022-07-22_15-25-54.mp4}  & \textbf{Disk}     & \textbf{Annulus}       & - \\
6  & \url{2022-07-22_17-06-29.mp4}  & \textbf{Disk}     & Disk                   & - \\
7  & \url{2022-09-15_15-45-26.mp4}  & \textbf{Disk}     & Disk                   & - \\
8  & \url{2022-09-16_10-38-16.mp4}  & \textbf{Disk}     & \textbf{Annulus}       & - \\
9  & \url{2022-09-16_12-42-32.mp4}  & Annulus           & \textbf{Annulus}       & - \\
10 & \url{2022-09-19_10-41-44.mp4}  & Annulus           & \textbf{Annulus}       & - \\
11 & \url{2022-09-19_12-14-47.mp4}  & \textbf{Disk}     & \textbf{Annulus}       & - \\
12 & \url{2022-09-19_13-43-33.mp4}  & \textbf{Disk}     & Disk                   & - \\
13 & \url{2022-09-19_15-24-21.mp4}  & \textbf{Disk}     & \textbf{Annulus}       & - \\
14 & \url{2022-09-19_16-58-24.mp4}  & Annulus           & \textbf{Annulus}       & - \\
\hline
\end{tabular}
}
\caption{List of Kilobots experimental runs and names of the associated video files. Each experiment involved two arenas: Disk (color code: cyan) and Annulus (color code: violet). Correctly detected arenas are in bold. Note that experiment 0 used the color code ``orange" rather than cyan, as we initially intended to use the former to indicate a disk arena. We changed the color code to cyan to ease visualization. However, the results of this experiment remain correct, as they used the same code, parameters and centroids as the other 14 experiments.
Raw video files can be found at~{\scriptsize\url{https://doi.org/10.6084/m9.figshare.31069147}}
}
\label{tab:expe_runs_list}
\end{center}
\end{table*}

\end{document}